 \documentclass[final,5p,times,twocolumn]{elsarticle}
\usepackage{cuted}
\usepackage{longtable}

\usepackage{multirow}
\usepackage[normalem]{ulem}

\usepackage{array}
\usepackage{booktabs}
\usepackage{multirow}
\renewcommand{\arraystretch}{1.2}  
\usepackage[edges]{forest}
\usepackage{hyperref} 
\usepackage{amssymb}
\usepackage{amsmath}

\usepackage{graphicx}
\usepackage{caption,subcaption}
\usepackage{amssymb}
\usepackage[utf8]{inputenc} % allow utf-8 input
\usepackage[T1]{fontenc}    % use 8-bit T1 fonts
\usepackage{hyperref}       % hyperlinks
\usepackage{url}            % simple URL typesetting
\usepackage{booktabs}       % professional-quality tables
\usepackage{amsfonts}       % blackboard math symbols
\usepackage{nicefrac}       % compact symbols for 1/2, etc.
\usepackage{microtype}      % microtypography
\usepackage{xcolor}         % colors
\usepackage{array} % 
\usepackage{multirow}
\usepackage{tabularx}
\usepackage{algorithm}  
\usepackage{algorithmicx}  
\usepackage{algpseudocode}
\usepackage{booktabs} %  
\usepackage{rotating} 
 \usepackage{amsthm}
 \newtheorem{theorem}{Theorem}
 \newtheorem{definition}{Definition}
 
\usepackage{array,makecell}

\usepackage{booktabs,makecell,multirow,tabularx}
\newcolumntype{C}{>{\centering\arraybackslash}X}

\usepackage{tikz}  
\usetikzlibrary{shapes, arrows, positioning, fit, backgrounds, calc}

\algrenewcommand\algorithmiccomment[1]{\hfill \textcolor{blue}{// #1}}

\usepackage{booktabs}

\usepackage{lineno}
\begin{document}

\begin{frontmatter}

%% Title, authors and addresses

%% use the tnoteref command within \title for footnotes;
%% use the tnotetext command for theassociated footnote;
%% use the fnref command within \author or \affiliation for footnotes;
%% use the fntext command for theassociated footnote;
%% use the corref command within \author for corresponding author footnotes;
%% use the cortext command for theassociated footnote;
%% use the ead command for the email address,
%% and the form \ead[url] for the home page:
%% \title{Title\tnoteref{label1}}
%% \tnotetext[label1]{}
% \author{Juan Zhang\corref{cor1}\fnref{label2}}
% \ead{zhang\_juan@buaa.edu.cn}
%% \fntext[label2]{}
%% \cortext[cor1]{}
%% \affiliation{organization={},
%%            addressline={}, 
%%            city={},
%%            postcode={}, 
%%            state={},
%%            country={}}
%% \fntext[label3]{}

\title{ViSymRe: Vision Multimodal Symbolic Regression} %% Article title

% use optional labels to link authors explicitly to addresses:
 \author[label1]{Da Li}
 \ead{dli@nenu.edu.cn }
 \author[label1,label3,label4,label6]{Junping Yin}
 \ead{yinjp829829@126.com}
 \author[label3]{Jin Xu}
\ead{xujin22@gscaep.ac.cn}
 \author[label5]{Xinxin Li}
 \ead{51265500102@stu.ecnu.edu.cn}
 \author{Juan Zhang\corref{cor1}\fnref{label2,label6}}
 \ead{zhang\_juan@buaa.edu.cn}
\cortext[cor1]{Corresponding author at: Institute of Artificial
	Intelligence, Beihang University}

% \ead{dli@nenu.edu.cn (Da Li), yinjp829829@126.com (Junping Yin), xujin22@gscaep.ac.cn (Jin Xu), 51265500102@stu.ecnu.edu.cn (Xinxin Li), zhang\_juan@buaa.edu.cn}
 \affiliation[label1]{organization={Academy for Advanced Interdisciplinary
 		Studies, Northeast Normal University},
%             addressline={},
             city={Changchun},
             postcode={130024},
             state={Jilin},
             country={China}}
 \affiliation[label2]{organization={Institute of Artificial
		Intelligence, Beihang University},
	%             addressline={},
	city={Beijing},
	postcode={100191},
	state={Beijing},
	country={China}}
 
 \affiliation[label3]{organization={Institute of Applied Physics and Computational Mathematics},
	%             addressline={},
	city={Beijing},
	postcode={100094},
	state={Beijing},
	country={China}}
\affiliation[label4]{organization={National Key Laboratory of Computational Physics},
	%             addressline={},
	city={Beijing},
	postcode={100088},
	state={Beijing},
	country={China}}
 \affiliation[label5]{organization={School of Mathematical Sciences, East China Normal University},
	%             addressline={},
	city={Shanghai},
		postcode={200241},
	state={Shanghai},
	country={China}}
\affiliation[label6]{organization={Shanghai Zhangjiang Institute of Mathematics},
	%             addressline={},
	city={Shanghai},
	postcode={201203},
	state={Shanghai},
	country={China}}
%\author{Da Li} %% Author name
%
%%% Author affiliation
%\affiliation{organization={},%Department and Organization
%            addressline={}, 
%            city={},
%            postcode={}, 
%            state={},
%            country={}}

%% Abstract
\begin{abstract}
Extracting interpretable equations from observational datasets to describe complex natural phenomena is one of the core goals of artificial intelligence. This field is known as symbolic regression (SR). In recent years, Transformer-based paradigms have become a new trend in SR, addressing the well-known problem of inefficient search. However, the modal heterogeneity between datasets and equations often hinders the convergence and generalization of these models. In this paper, we propose ViSymRe, a \textbf{Vi}sion \textbf{Sym}bolic \textbf{Re}gression framework, to explore the positive role of visual modality in enhancing the performance of Transformer-based SR paradigms. To overcome the challenge where the visual SR model is untrainable in high-dimensional scenarios, we present \textbf{M}ulti-\textbf{V}iew \textbf{R}andom \textbf{S}licing (MVRS). By projecting multivariate equations into 2-D space using random affine transformations, MVRS avoids common defects in high-dimensional visualization, such as variable degradation, non-linear interaction missing, and exponentially increasing sampling complexity, enabling ViSymRe to be trained with low computational costs. To support dataset-only deployment of ViSymRe, we design a dual-vision pipeline architecture based on generative techniques, which reconstructs visual features directly from the datasets via an auxiliary Visual Decoder and automatically suppresses the attention weights of reconstruction noise through a proposed Biased Cross-Attention feature fusion module, ensuring that subsequent processes are not affected by noisy modalities. Ablation studies demonstrate the positive contribution of visual modality to improving model convergence level and enhancing various SR metrics. Furthermore, evaluation results on mainstream benchmarks indicate that ViSymRe achieves competitive performance compared to baselines, particularly in low-complexity and rapid-inference scenarios.
\end{abstract}

%%graphical abstract
%\begin{graphicalabstract}
%\includegraphics[width=\textwidth]{g_abstract.png}
%\end{graphicalabstract}
%\begin{highlights}
%	\item We propose ViSymRe, a Visual Multimodal Symbolic Regression framework, achieving performance superior to traditional dataset-only paradigms.
%	
%	\item We introduce Multi-View Random Slicing (MVRS), a visualization strategy that projects multivariate equations into a 2-D space with limited sampling complexity, thereby facilitating large-scale pre-training. 
%	
%	\item We propose a Biased Cross-Attention feature fusion module. By introducing a bias term into the attention scores, this module suppresses the attention weights of noisy modalities, significantly enhancing multimodal learning performance.
%\end{highlights}
%%%Research highlights

%% Keywords

\end{frontmatter}

\section{Introduction}
\label{Introduction}

Revealing the underlying laws of natural phenomena from observational datasets has emerged as a core goal in artificial intelligence. In 2009, Schmidt and Lipson published an influential study in \emph{Science} demonstrating that symbolic regression (SR) can derive laws of classical mechanics from motion-tracking data~\cite{schmidt2009distilling}. This work highlights the potential of SR in promoting scientific discovery~\cite{wang2023scientific, krenn2022scientific}. As a data-driven method, SR aims to automatically discover equations that best describe the datasets. In fields such as physics, chemistry, and ecology, SR is widely used~\cite{quade2016prediction, neumann2020new, kim2020integration, burlacu2023symbolic}.

In the early 1990s, Koza introduced genetic programming (GP) for SR, where the individuals are regarded as equation trees with operators, variables, and constants as nodes~\cite{koza1994genetic}. GP simulates biological processes like gene replication, crossover, and mutation to evolve individuals~\cite{schmidt2010age, la2016epsilon, udrescu2020ai}. When the initial population is large, and the probabilities of crossover and mutation are well-calibrated, GP-based models exhibit resistance to local optima. Nevertheless, the extensive search space often necessitates substantial computational resources, which has inspired the development of deep learning (DL)-based SR models that guide equation search through gradient descent or reinforcement learning~\cite{derner2018data, petersen2019deep, mundhenk2021symbolic, scholl2025parfam}. However, both GP-based and DL-based models are optimized from scratch for each problem, which limits their ability to leverage prior experience to accelerate convergence on new problems.

Recent advances in SR have introduced Transformer architectures~\cite{biggio2021neural, valipour2021symbolicgpt, kamienny2022end, vastl2024symformer, chen2025bootstrapping}, reformulating the SR problem as a dataset-to-equation translation task, similar to machine translation models. Different from traditional GP and DL-based models, Transformer-based paradigms leverage the attention mechanism to capture dependencies within datasets, enabling direct equation generation without iteration. However, despite improving deployment efficiency, these models often fail to adequately account for the modal gap between datasets and equations, which increases convergence difficulty. To address this issue, MMSR~\cite {li2025mmsr} pioneered multimodal learning in SR. It considers the fusion representation of the datasets and equations as direct inputs during training, and introduces multimodal techniques such as Cross-Attention feature fusion and alignment loss, demonstrating excellent performance compared to dataset-only baselines across several scenarios. Our solution further advances multimodal learning by incorporating visual modality, a readily available and key resource, but has been overlooked in previous research.

Currently, visual multimodal models have proven effective for enhancing various machine learning tasks, especially in machine translation.~\cite{gupta2023cliptrans, shen2024survey}. The rationale is intuitive. For example, in machine translation, the English word ``Dog'' and the German word ``Hund'' share the same underlying visual perception. Analogously, in SR, although datasets and equations present different forms, they correspond to the same geometric topology. 

However, existing multimodal models typically rely on specialized equipment such as CT and radar, or precreated databases to ensure that the visual modality is readily available during training and inference~\cite{zhang2024mm}, which is challenging for SR. The main obstacle is that existing high-dimensional visualization approaches either incur prohibitive CPU overhead due to the exponential increase in the number of sampling points or suffer from variable degeneracy and non-linear interaction missing. Consequently, the scope of current discussions on visual SR is restricted to 2-D and below~\cite{chen2025bootstrapping}.

To address this issue, we propose \textbf{M}ulti-\textbf{V}iew \textbf{R}andom \textbf{S}licing (MVRS), which projects multivariate equations via 2-D affine transformations defined by a random orthogonal basis. With only controllable sampling points, the complex and abstract symbolic relationships can be transformed into visible patterns, such as curves and gradients, thereby allowing standard vision models to capture and understand them regardless of dimensionality. Theoretically, we prove that MVRS ensures the non-degeneracy of variables and does not lose crucial non-linear interactions.

Based on MVRS, we propose ViSymRe, a Transformer-based \textbf{Vi}sion \textbf{Sym}bolic \textbf{Re}gression framework. While MVRS enables low-cost visualization, its application is limited by the analytical equations, a condition that holds only during training.
To achieve dataset-only inference, ViSymRe employs an auxiliary Visual Decoder to predict visual features directly from dataset embeddings, thereby eliminating the need for real visual input. Diverging from traditional pixel-level reconstruction or direct feature regression, we incorporate a Vector Quantization mechanism to project visual features into a finite space~\cite{van2017neural}. Specifically, ViSymRe maintains a learnable Codebook that discretizes the continuous visual feature space into a finite set of codewords. The Visual Decoder performs multi-head classification conditioned on the dataset to predict the indices of codewords corresponding to the visual features. By transforming the visual reconstruction task from regression to classification, we avoid the common mean collapse issue in high-dimensional regression~\cite {mathieu2015deep}. In essence, ViSymRe establishes a dual-visual pipeline: a real pipeline and a virtual pipeline. The former supervises the convergence of the Visual Encoder to ensure the validity of visual features, while the latter enables dataset-only inference.

Furthermore, we propose a Biased Cross-Attention module inspired by~\cite{press2021train} to replace the standard Cross-Attention module in the virtual vision pipeline for feature fusion. By imposing a bias to suppress the attention weights of noisy components in the predicted visual features, this module effectively mitigates the issue of noise being overemphasized. Finally, we introduce a constraint algorithm during the decoding process to improve the interpretability and physical plausibility of the generated equations~\cite{petersen2019deep}. By enforcing arity constraints at each step and prohibiting physically illogical nested structures, this approach prunes the search space, thereby prioritizing the generation of structurally valid equations.

Overall, our main innovations are as follows:
\begin{itemize}
	\item We propose ViSymRe, a Transformer-based framework that integrates the visual modality as a bridge into SR to address the challenge of modal heterogeneity between datasets and equations faced by traditional pre-trained models. Extensive evaluations demonstrate that the proposed architecture is robust across various scenarios, particularly showing strong potential for low-complexity and rapid-inference tasks.

	\item We propose \textbf{M}ulti-\textbf{V}iew \textbf{R}andom \textbf{S}licing (MVRS), a visualization strategy to support large-scale pre-training.  MVRS projects multivariate equations into 2-D space without causing variable degeneracy or exponential sampling complexity, thereby generalizing ViSymRe to high-dimensional scenarios at a low computational cost. 
	
	\item We construct a dual-visual pipeline that introduces virtual vision to reconcile conflicts between multimodal training and dataset-only inference. By employing a Biased Cross-Attention feature fusion module to suppress attention weights assigned to noisy components in the virtual vision, the two vision pipelines are ensured to converge synchronously.
\end{itemize}

The remainder of this paper is organized as follows: Section~\ref{motivation} introduces the motivation underlying our work. Section~\ref{related work} reviews the related work. Section~\ref{method} presents the ViSymRe framework in detail, including its core modules and parameters. Section~\ref{data preprocessing} introduces the data preprocessing, including skeleton generation and dataset sampling. Section~\ref{experimental} provides experimental settings and results to validate the effectiveness of ViSymRe. Finally, Section~\ref{conclusions} discusses the limitations of ViSymRe and outlines potential directions for future research.

\section{Motivation}
\label{motivation}

Transformer-based paradigms have become a new trend for SR. They enable one-time inference, thus avoiding expensive iterative search. However, adapting the machine translation paradigms to SR remains challenging. A key difficulty lies in the modal heterogeneity between the datasets and equations, which limits the models' convergence. To address this problem, Multimodal learning applied to SR has been preliminarily explored in~\cite{li2025mmsr}, ensuring modal homogeneity by treating the equations as input. Our solution is to introduce the visual modality as a bridge to alleviate this limitation. The dataset-only paradigm requires models to generate equations from local, fragmented data points, whereas the multimodal paradigm provides a global perspective.

For a long time, a fundamental challenge in deploying visual multimodal models has been the unavailability of visual inputs during inference. Recently, this limitation has been addressed in the field of machine translation by generative techniques, such as ``visual hallucination'' proposed by VALHALLA~\cite{li2022valhalla}. This method trains a Visual Hallucination Transformer that enables the model to predict missing visual modality from available text sequences. Applying this idea to SR is even more promising than in machine translation. Unlike natural images, which require modeling complex textures, lighting, and fine-grained object details, equation images are structurally simpler. Clear, deterministic geometric lines characterize them and are free of cluttered backgrounds, thus being reconstructed more easily. Based on the above considerations, we propose ViSymRe, providing a reference solution for visual SR. Furthermore, compared to the multimodal models in other fields, the multimodal SR model exhibits significant advantages in terms of lightweight and cost, for the following reasons:
 \begin{itemize} 
	\item Unlike complex natural images, equation images are structurally sparse and informative even at lower resolutions, enabling the use of a lightweight Visual Encoder. 
	\item The pre-training benefits from automated data generation. The equations and their corresponding visual representations can be generated automatically, providing a cost-effective data source without manual labeling. 
\end{itemize}

\section{Related work}
\label{related work}

\textbf{GP-based SR models} GP is the foundational paradigm for SR~\cite{koza1994genetic}. By simulating biological evolution, GP-based models can autonomously search in a vast equation space. While suitable for most SR scenarios, GP-based models also face challenges in terms of computational burden and hyperparameter sensitivity. To address these issues, recent research has increasingly focused on improving search strategies and integrating gradient learning techniques to guide evolutionary processes~\cite{schmidt2009distilling, schmidt2010age, arnaldo2014multiple, la2016epsilon, virgolin2019linear, udrescu2020ai, cranmer2023interpretable, de2025improving}. Furthermore, some studies have considered using domain knowledge to purify the search space, further enhancing the interpretability of generative equations~\cite{kubalik2020symbolic, kubalik2021multi, radwan2024comparison}. 

\textbf{DL-based SR models} Neural networks, renowned for their superior numerical approximation capabilities, have become a powerful tool for SR. Equation Learner (EQL) innovates upon the standard fully connected architecture by replacing static activation functions with mathematical operators~\cite{martius2016extrapolation}, enabling the generation of interpretable equations. However, EQL faces optimization challenges. Despite inheriting the approximation capabilities of neural networks, singularities in division operations and high-frequency oscillations in periodic functions lead to gradient instability, prompting the development of $\textit{EQL}^{\div}$~\cite{sahoo2018learning}. Deep symbolic regression (DSR) employs recurrent neural networks to autoregressively construct the equation skeleton~\cite{petersen2019deep}. By imposing syntax constraints, DSR ensures the generation of physically valid equations. However, this typically comes at the cost of reduced numerical approximation capabilities. To address this problem, integration frameworks like uDSR have integrated GP, DL, and pre-trained strategies to refine the SR process, avoiding local optima caused by a single method~\cite{mundhenk2021symbolic, landajuela2021discovering, mundhenk2021symbolic, landajuela2022unified}. PARFAM further extends the application of DL in SR by representing equations as coupled rational neural networks, demonstrating robust performance in recovering complex polynomials~\cite{scholl2025parfam}.

\textbf{Transformer-based SR models} To address the computational bottlenecks of traditional search algorithms, recent research has introduced Transformer architectures. NeSymReS and SymbolicGPT provide inspiring paradigms~\cite{biggio2021neural, valipour2021symbolicgpt}, which treat the dataset as a high-dimensional point cloud and the equation skeleton as a tokenized sequence. They learn a robust mapping from the dataset to the equation skeleton through pre-training on tens of millions of pre-generated skeleton-dataset pairs. Inference requires only one forward pass, thus significantly improving efficiency. Subsequently, End-to-End (E2E)~\cite{kamienny2022end} and SymFormer have advanced this field by encoding constants directly as tokens or integrating dedicated regression heads, achieving simultaneous prediction of equation skeleton and constants~\cite{vastl2024symformer}.

Overall, Transformer-based models offer an advantage in inference efficiency, avoiding the iterative search of GP and DL-based models by leveraging priors learned during training. However, these models face challenges in out-of-distribution generalization. To bridge this gap, some rectification strategies, such as TPSR~\cite{shojaee2023transformer}, SNR~\cite{liu2023snr}, and SR-GPT~\cite{li2024discovering}, have been proposed to offer feedback about fitting losses and complexity to generation policies. These rectification strategies are crucial for enhancing the generalization of Transformer-based models, significantly broadening their application scenarios at low cost. 

\textbf{Multimodal models} For a long time, machine learning has primarily focused on unimodal tasks, processing individual modalities such as vision, text, or audio. The emergence of deep learning has promoted the development of multimodal learning. Early approaches typically integrate different modalities into a unified feature space directly through concatenation or feedforward networks~\cite{ngiam2011multimodal, srivastava2012multimodal, fukui2016multimodal}. The introduction of contrastive learning facilitates the modal alignment prior to fusion~\cite{radford2021learning, li2021align, jia2021scaling}. ViLBERT~\cite{lu2019vilbert} and LXMERT~\cite{tan2019lxmert} introduce the Cross-Attention, enabling more sophisticated interactions between vision and language. Generative models, such as DALL·E~\cite{ramesh2021zero}, demonstrate the capacity of large-scale multimodal architectures to synthesize semantically coherent images. Recently, multimodal learning has been introduced into SR. For instance, MMSR~\cite{li2025mmsr} considers the fusion representation of the dataset and equations while employing contrastive learning during training, achieving significant numerical approximation results.

\section{Method}
\label{method}

\begin{figure*}[t]
	\centering
	\includegraphics[width=1\textwidth]{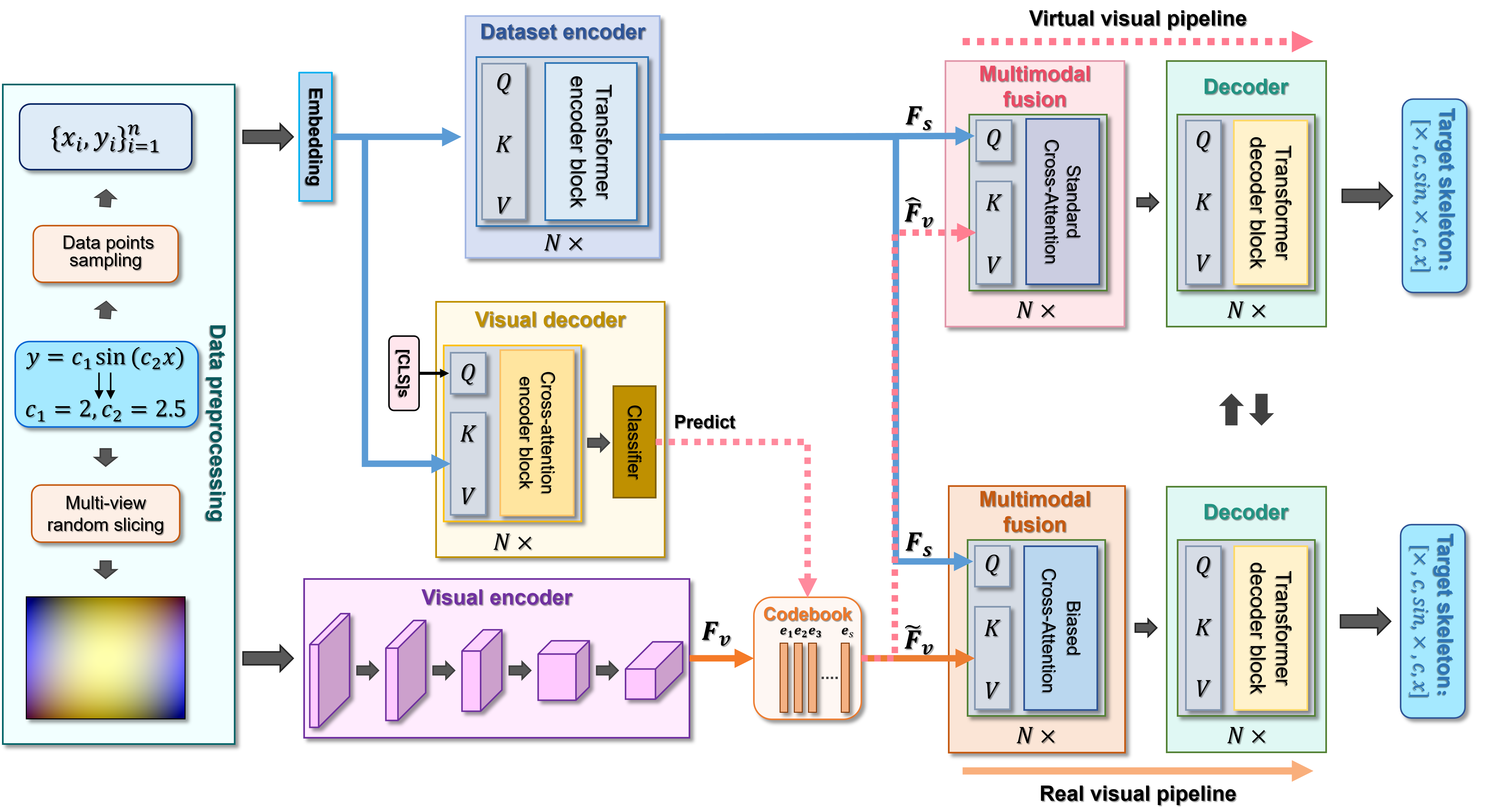}
	\caption{\textbf{Overview of the ViSymRe framework.} During data preprocessing, the inputs, consisting of MVRS slices and randomly sampled datasets, are generated. The Dataset Encoder and Visual Encoder extract visual features ($F_v$) and dataset features ($F_s$) from these inputs, respectively. $F_v$ are then quantized into discrete representations (denoted as $\widetilde{F}_v$) in a Codebook. $\widetilde{F}_v$ and $F_s$ are fused through a standard Cross-Attention module, which is decoded into the target equation skeleton. To address the issue of unavailable MVRS slices during inference, a Visual Decoder is integrated into  ViSymRe to learn to predict visual features (called virtual vision and denoted as $\hat{F}_v$) conditioned on dataset embeddings, thereby establishing an independent virtual visual pipeline that shares the same decoder with the real visual pipeline, but integrates a Biased Cross-Attention feature fusion module to suppress the attention weights of predicted noise.}
	\label{fig:architecture}
\end{figure*}

\begin{table*}[htbp]
	\centering
	\caption{Summary of key symbols used in this paper.}
	\label{tab:key_symbols}
	\renewcommand{\arraystretch}{1.3} 
	\begin{tabularx}{\textwidth}{l|X} 
		\toprule
		\textbf{Symbol} & \textbf{Description} \\
		\midrule
		\multicolumn{2}{l}{\textit{\textbf{MVRS Symbols}}} \\
		$K$ & Number of views generated by MVRS \\
		$\Psi$ & Affine function \\
		$s, t$ & Local coordinates of the slice, $s, t \in [-L, L]$ \\
		$\mathfrak{c}, u, v$ & Slice parameters: center vector $\mathfrak{c}$ and orthonormal basis vectors $\{u, v\}$ \\
		$L, \Omega$ & $L$: Field of slice; $\Omega$: Truncation threshold for visualization \\
		$I$ & Multi-view slice generated by MVRS, $I \in \mathbb{R}^{H \times W \times K}$ \\
		\midrule
		\multicolumn{2}{l}{\textit{\textbf{Architecture Symbols}}} \\
		$N$ & Number of sampled data points \\
		$d$ & Dimension of input variables \\
		$M$ & Sequence length of visual features \\
		$S$ & Size of the Codebook \\
		$x_i, y_i$ & Variable vector and target value for the $i$-th sample, where $x_i \in \mathbb{R}^d, y_i \in \mathbb{R}$ \\
		$F_s$ & Dataset features output by the Dataset Encoder \\
		$F_v$ & Continuous visual features output by the Visual Encoder \\
		$\tilde{F}_v, \hat{F}_v$ & $\tilde{F}_v$: Quantized real visual features; $\hat{F}_v$: Virtual visual features predicted by the Visual Decoder \\
		$E$ & Codebook containing $S$ learnable codewords \\
		$B$ & Bias matrix in the Biased Cross-Attention mechanism \\
		\bottomrule
	\end{tabularx}
\end{table*}

The overall architecture of ViSymRe is shown in Fig.~\ref{fig:architecture}. In the following sections, we detail all modules and parameters. The key symbols used in this paper are summarized in Table~\ref{tab:key_symbols}.

\subsection{Dataset Encoder}
\label{Dataset Encoder}
The input dataset is denoted as $\{(\mathbf{x}_i, y_i)\}_{i=1}^N$, where $\mathbf{x}_i \in \mathbb{R}^d$ represents the variables and $y_i \in \mathbb{R}$ represents the target. Inspired by NeSymReS\cite{biggio2021neural}, we adopt an IEEE-754–based normalization, which encodes each floating point number as a multi-bit binary representation composed of one sign bit, eight exponent bits, and eight mantissa bits (see~\ref{IEEE-754} for details). 

We use a Set Transformer~\cite{lee2019set} module as the Dataset Encoder, which has been proven effective in~\cite{biggio2021neural}. The output dataset features are denoted as ${F}_{s}=\{{f}_{s_i}\}_{i=1}^N$, aligned with each point in the dataset.

\subsection{Multi-view random slicing}
\label{MVRS}
In SR, the use of visual modality remains underexplored due to the lack of a unified visualization method. Traditional methods, such as Principal Component Analysis (PCA) and t-SNE, suffer from the curse of dimensionality, requiring an exponentially increasing sampling complexity to maintain fidelity, making large-scale pre-training difficult.
A feasible method to address this problem is marginal projection, which visualizes the curve of a single variable by fixing other variables. However, this approach leads to variable degradation when non-linear interaction terms are present. Consider a simple equation: $f(\mathbf{x}) = x_1 (x_2-x_3)$. If we project $x_1$ by fixing $x_2=x_3=1$, the equation collapses to $f(x_1, 1, 1) \equiv 0$, resulting in the interaction information missing.

To overcome the interaction information missing caused by static projection, we propose MVRS. This method does not use a fixed coordinate system. Instead, it leverages the statistical isotropy of the high-dimensional feature space to perform continuous slicing on a random affine subspace. Considering an equation with $d$ variables: $f:\mathbb{R}^d \to \mathbb{R}$. We define the slicing process as an isometric mapping from the 2-D visualization space to the $d$-D variable space. Let the local coordinates be $(s, t)$, where $s, t \in [-L, L]$. We define an affine function $\Psi: \mathbb{R}^2 \to \mathbb{R}^n$:
\begin{equation}
	\Psi(s, t) = \mathbf{c} + s \cdot \mathbf{u} + t \cdot \mathbf{v}
\end{equation}
Here, $\mathbf{c}$ denotes the slice center, and $\mathbf{U}= [\mathbf{u}, \mathbf{v}]$ form a basis matrix. The variable $x_i$ is parameterized as a linear function of the local coordinates: 
\begin{equation}
	x_i = c_i + s u_i + tv_i,
\end{equation}
yielding a bivariate function $f(\Psi(s,t))$. 

\begin{figure}[h]
	\centering
	\begin{subfigure}[b]{0.49\linewidth}
		\centering
		\includegraphics[width=\linewidth]{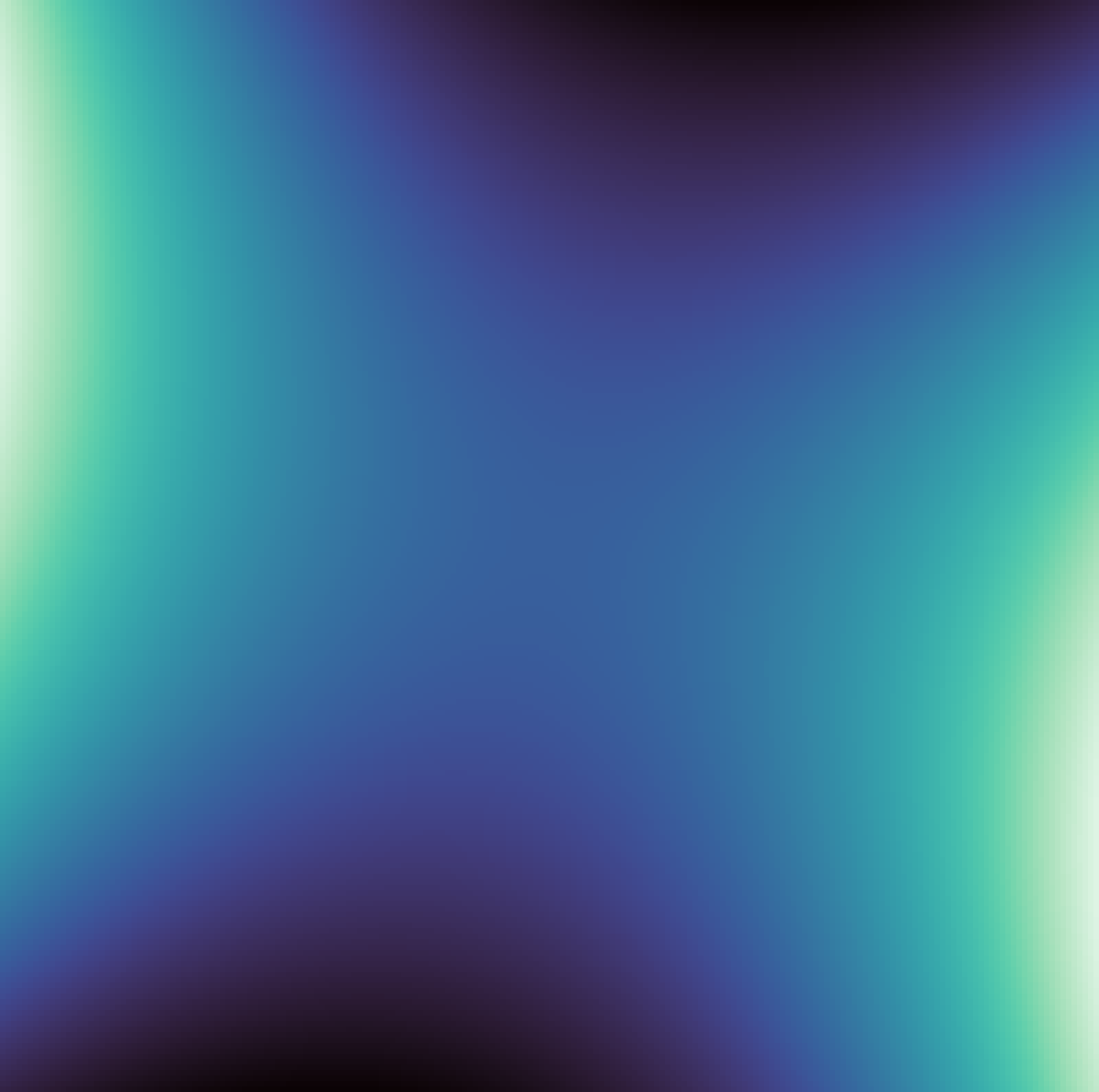}
		\caption{Single-view random slice}
		\label{fig:mvrs}
	\end{subfigure}
	\hfill
	\begin{subfigure}[b]{0.49\linewidth}
		\centering
		\includegraphics[width=\linewidth]{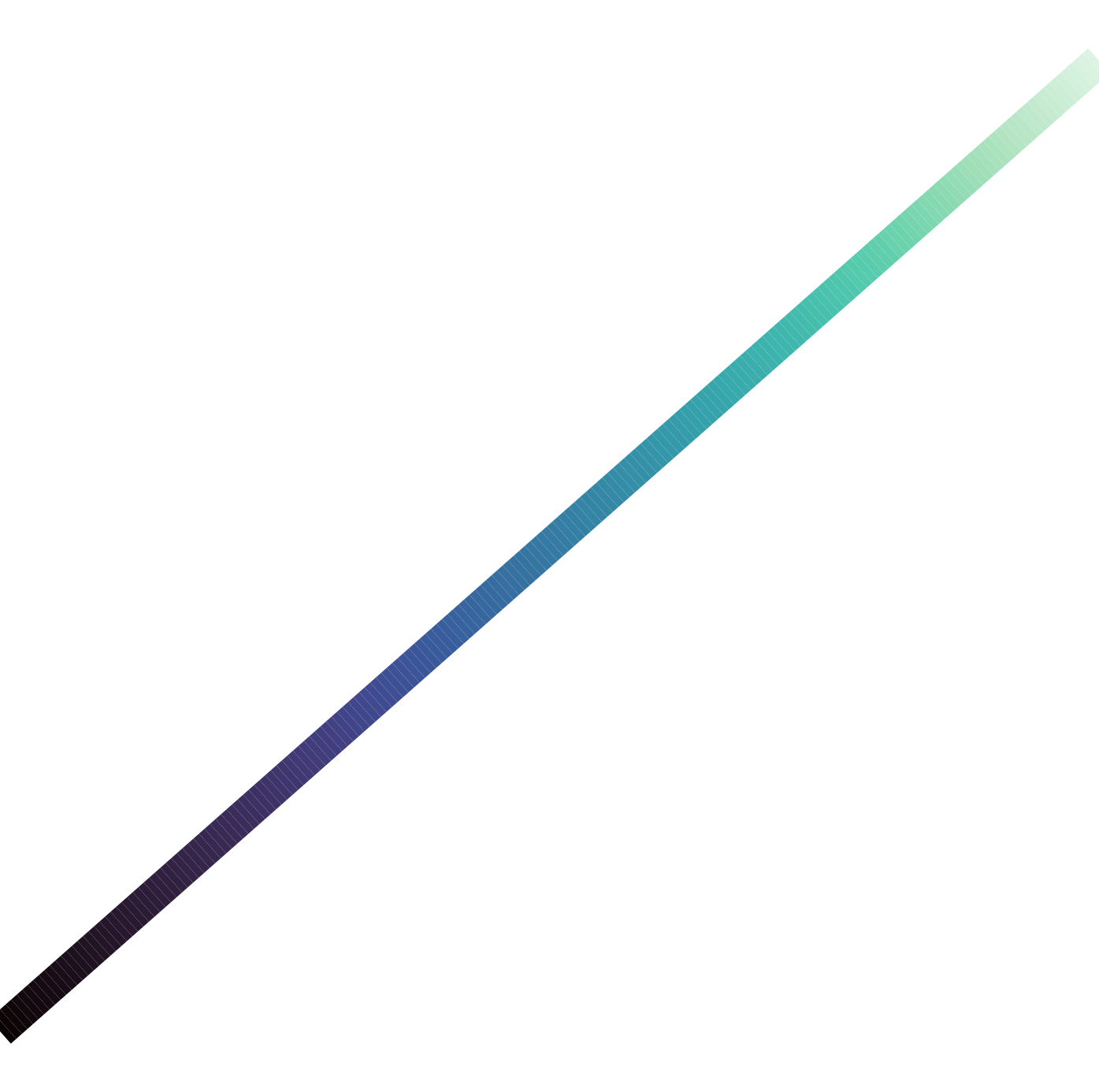}
		\caption{marginal projection by fixing $x_2=x_3=1$}
		\label{fig:marginal}
	\end{subfigure}
	\vspace{0.2cm} 
	\begin{subfigure}[b]{0.49\linewidth}
		\centering
		\includegraphics[width=\linewidth]{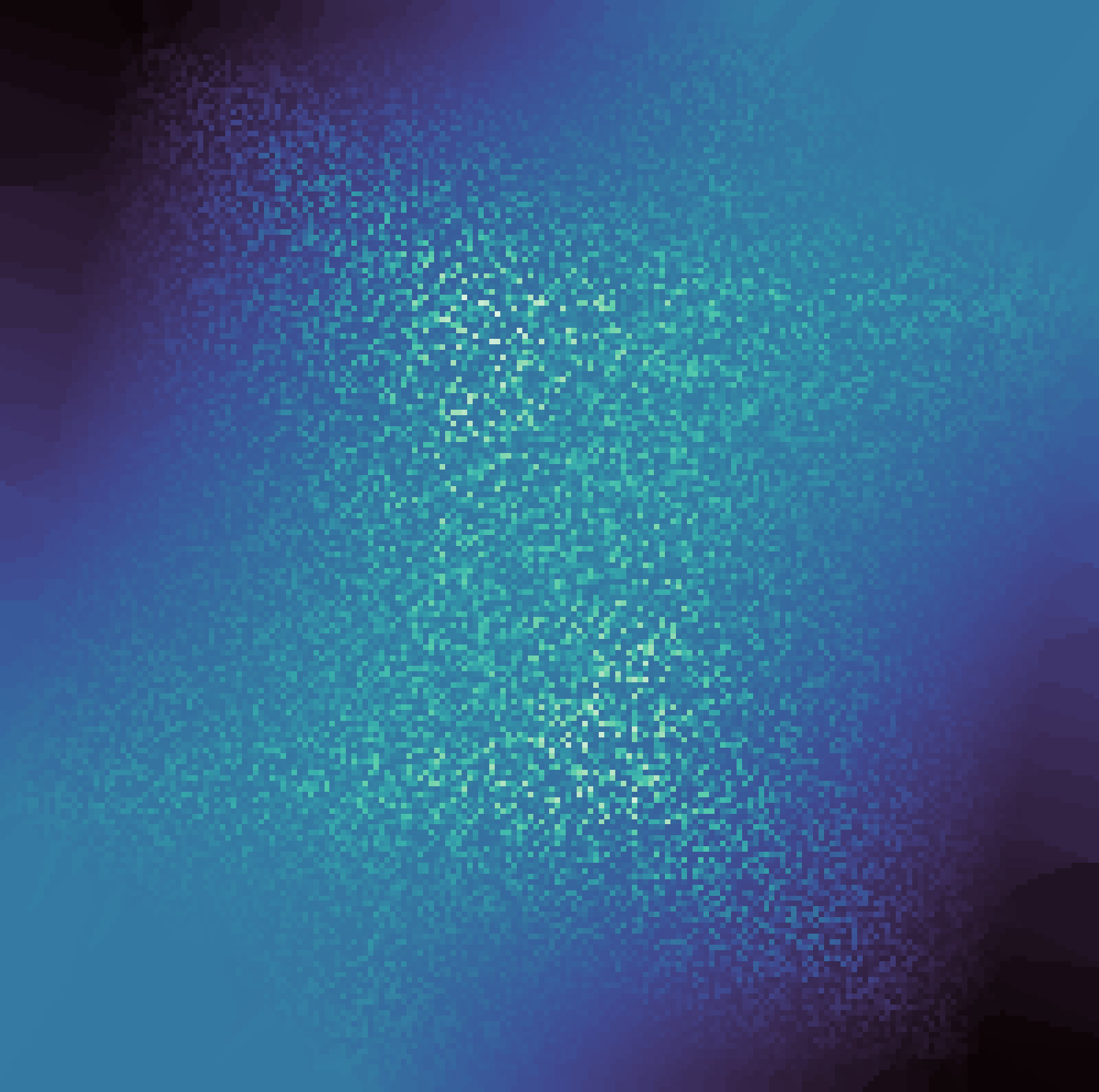}
		\caption{PCA}
		\label{fig:pca}
	\end{subfigure}
	\hfill
	\begin{subfigure}[b]{0.49\linewidth}
		\centering
		\includegraphics[width=\linewidth]{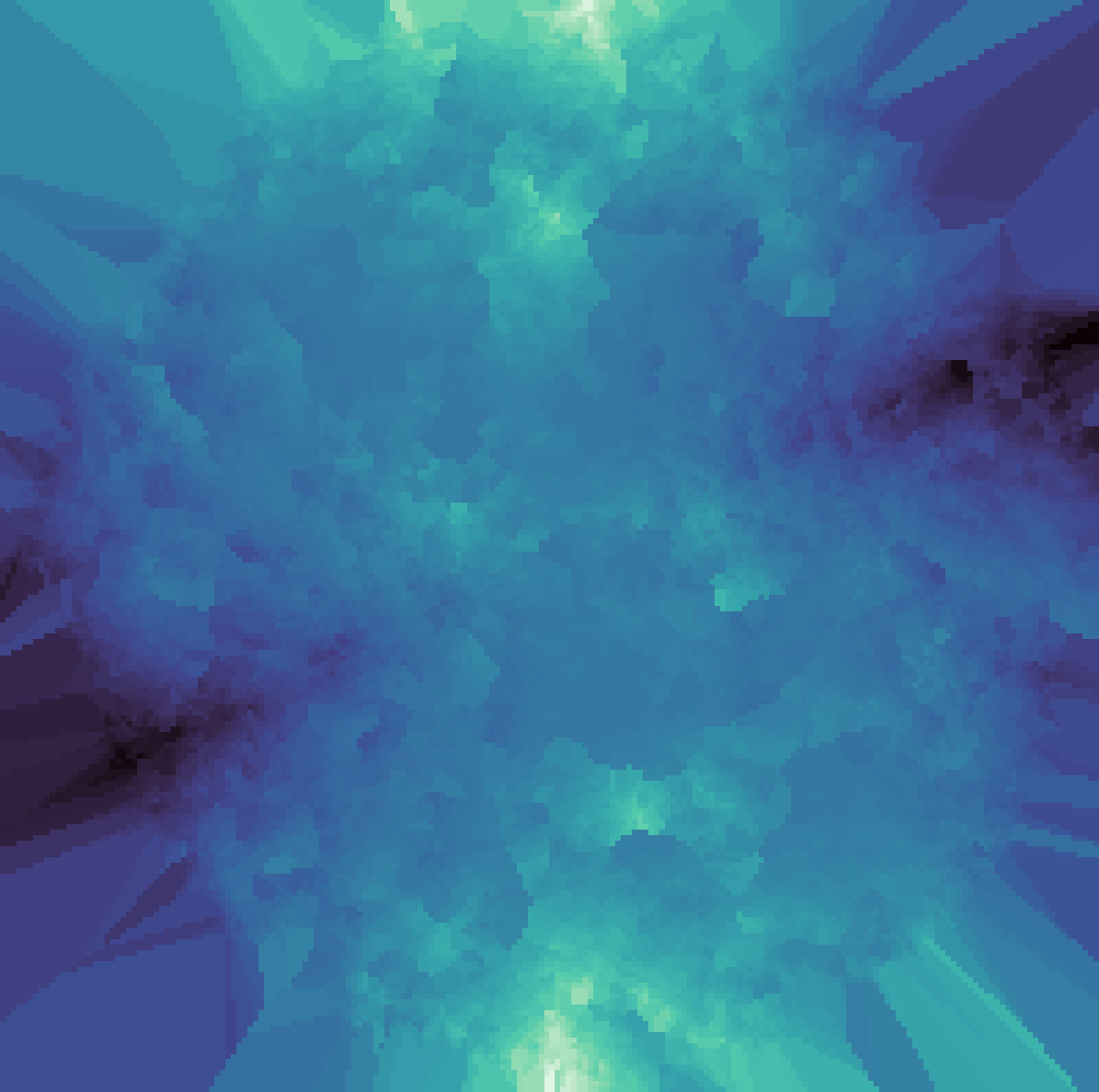}
		\caption{t-SNE}
		\label{fig:tsne}
	\end{subfigure}
	
	\caption{\textbf{Visualization results comparison of four visualization methods for the interaction term $x_1(x_2 - x_3)$}. 
 The proposed method \textbf{(a)} preserves the saddle shape via random isotropic slicing. Marginal projection \textbf{(b)} suffers from feature collapse, degenerating the non-linear interaction into a misleading linear response. Traditional methods \textbf{(c and d)}: PCA exhibits information loss due to projection overlap, while t-SNE fails to reconstruct the manifold continuity, resulting in topological tearing even with dense sampling. Notably, MVRS requires only $N \cdot d$ points, while PCA and t-SNE require $N^d$ points. The computational burden is a key reason why traditional methods cannot support large-scale pre-training.}
	\label{fig:visualization_comparison}
\end{figure}

To ensure unbiased projection, the basis matrix $\mathbf{U}$ is constructed as an orthonormal basis. To eliminate directional bias and ensure rotational invariance, we sample random vectors $\mathbf{Z} = [\mathbf{z_1}, \mathbf{z_2}]$ from a multivariate standard normal distribution $\mathcal{N}(\mathbf{0}, \mathbf{I}_d)$. Here, the probability density $p(\mathbf{z}) \propto \exp(-\frac{1}{2}\|\mathbf{z}\|^2)$ relies only on the radial distance, thus ensuring isotropy. The orthonormal basis vectors $\mathbf{u}$ and $\mathbf{v}$ are obtained by applying the Gram-Schmidt process to $\mathbf{z}_1$ and $\mathbf{z}_2$, respectively. To ameliorate the information loss associated with single-view, we employ multi-view projection. The output $I \in \mathbb{R}^{H \times W \times K}$ aggregates $K$ independent slices, which are generated based on different basis: $\{\mathbf{u}^{(k)}, \mathbf{v}^{(k)}\}_{i=1}^K$:
\begin{equation}
	I^{(k)}(s,t) = \mathcal{N} \left( f \left( \mathbf{c} + s \cdot \mathbf{u}^{(k)} + t \cdot \mathbf{v}^{(k)} \right) \right)
\end{equation}
where $\mathcal{N}$ denotes a normalization function. Fig.~\ref{fig:visualization_comparison} presents an example, illustrating how various visualization methods represent the equation $x_1(x_2 - x_3)$. Moreover, to prevent visual information degradation caused by MVRS in low-dimensional scenarios, we employ MVRS only for high-dimensional visualization. For 1-D and 2-D cases, we generate curves and surfaces respectively.

In the following section, we theoretically prove the non-degeneracy of MVRS and further demonstrate that it almost surely preserves high-dimensional topological structures.
\begin{theorem}\label{thm:non_degeneracy}(Non-degeneracy of affine transformation). 
$\forall i \in \{1, \dots, d\}$, the projection $x_i\circ \Psi$ constitutes a non-trivial function $\psi_i(s, t)$ almost surely.
\end{theorem}
\begin{proof}
	The local projection is defined as $\psi_i(s, t) \triangleq c_i + u_i s + v_i t$. The mapping degenerates to a constant if and only if its gradient vanishes:
	\begin{equation}
		\nabla \psi_i = \mathbf{0} \iff u_i = 0 \text{ and } v_i = 0.
	\end{equation}
	 Since the basis matrix $\mathbf{U}$ is obtained via the Gram-Schmidt orthogonalization of $\mathbf{Z} \sim \mathcal{N}(\mathbf{0}, \mathbf{I}_{d \times 2})$, the induced probability measure on the Stiefel manifold \footnote{The Stiefel manifold $V_{2}(\mathbb{R}^d)$ is defined as the set of $d \times 2$ matrices with orthonormal columns, i.e., $V_{2}(\mathbb{R}^d) = \{ \mathbf{M} \in \mathbb{R}^{d \times 2} \mid \mathbf{M}^\top \mathbf{M} = \mathbf{I}_2 \}$.} is absolutely continuous with respect to the Lebesgue measure. Consequently, the marginal distribution of any component $u_i$ (or $v_i$) exists a probability density function. 
	
	Since the singleton set $\{0\}$ has Lebesgue measure 0, it follows that:
	\begin{equation}
		\mathbb{P}(u_i = 0) = \mathbb{P}(v_i = 0) = 0.
	\end{equation}
	Degeneracy requires the intersection of these zero-probability events. Thus:
	\begin{equation}
		\mathbb{P}(\nabla \psi_i = \mathbf{0}) = \mathbb{P}(\{u_i = 0\} \cap \{v_i = 0\}) = 0.
	\end{equation}
	This implies that $\psi_i(s, t)$ is non-trivial almost surely.
\end{proof}

Having established the visibility of linear components in Theorem \ref{thm:non_degeneracy}, we focus on the challenge of non-linear interactions. We divide operators into two main categories:

\begin{itemize}
	\item\textbf{Unary operations and addition:} Operators such as $\{+, -, \sin(x), \cos(x), \sqrt{x}, |x|, \exp(x), x^p\}$ create geometric shapes, such as planes, bumps, or waves, which are inherently robust. Regardless of the viewing angle, these shapes do not completely degenerate into a line or a constant.
	
	\item \textbf{Multiplication:} The multiplication interaction term ($x_i x_j$) generates a saddle surface with opposite curvature, which may vanish under a specific projection. We regard this binary multiplication as the basic unit of non-linear interaction. Complex nested terms (such as $\sin(x_i x_j)$) or higher order couplings (such as $x_i x_j x_k$) structurally depend on this innermost multiplicative. Consequently, establishing the visibility of the fundamental saddle surface in Theorem \ref{thm:interaction_visibility} serves as a sufficient condition for resolving complex interactions.
\end{itemize}
 
\begin{theorem}\label{thm:interaction_visibility}
	(Non-degeneracy of non-linear interaction terms under affine transformation). Let the target equation contain a multiplicative interaction term $h(\mathbf{x}) = x_i x_j$ ($i \neq j$). The projection $h \circ \Psi$ constitutes a saddle surface almost surely.
\end{theorem}

\begin{proof}
	Substituting the affine transformation $\Psi$ (ignoring translation) into $h(\mathbf{x})$ yields the quadratic form $Q(s, t)$:
	\begin{equation}
		Q(s, t) = (u_i s + v_i t)(u_j s + v_j t) = A s^2 + B st + C t^2,
	\end{equation}
	where coefficients are defined as $A = u_i u_j$, $B = u_i v_j + u_j v_i$, and $C = v_i v_j$. The topological classification is determined by the discriminant $\Delta = B^2 - 4AC$. By algebraic reduction, $\Delta$ factorizes into a determinant square:
	\begin{equation}
		\begin{aligned}
			\Delta &= (u_i v_j + u_j v_i)^2 - 4(u_i u_j)(v_i v_j) \\
			&= (u_i v_j - u_j v_i)^2 = \left[ \det \begin{pmatrix} u_i & v_i \\ u_j & v_j \end{pmatrix} \right]^2.
		\end{aligned}
	\end{equation}
	Let $\mathbf{U}_{i,j} \in \mathbb{R}^{2 \times 2}$ be the submatrix formed by the $i$-th and $j$-th rows of $\mathbf{U}$. As established in the proof of Theorem \ref{thm:non_degeneracy}, $\mathbf{U}$ follows a continuous distribution on the Stiefel manifold. Since the condition $\det(\mathbf{U}_{i,j}) = 0$ defines a set of Lebesgue measure 0 within this manifold, the submatrix is full-rank almost surely:
	\begin{equation}
		P(\text{rank}(\mathbf{U}_{i,j}) < 2) = P(\det(\mathbf{U}_{i,j}) = 0) = 0.
	\end{equation}
	Consequently, $\Delta > 0$ holds with probability 1. In the classification of quadrics, $\Delta > 0$ is a sufficient and necessary condition for a saddle surface.
\end{proof}

Considering that equations often involve invalid domains or singularities, MVRS transforms these mathematical constraints into visual features through the following masking and truncation strategies:
\begin{equation}
	\hat{I}(s, t) = \begin{cases} 
		0 & \text{if} \; \Psi(s, t) \notin \mathcal{D}_{\text{valid}}  \\ 
		\text{sgn}[f(\Psi(s, t)] \cdot \Omega & \text{if} \; |f(\Psi(s, t))| > \Omega\\
		f(\Psi(s, t) & otherwise
	\end{cases}
\end{equation}
where $\Omega$ is a truncation threshold. 

To ensure that MVRS generalizes across datasets with different scales and distributions, the parameters $\mathbf{c}$ and $L$ must be adaptive. To this end, we propose a statistics-based adaptation strategy:

\begin{itemize}
\item \textbf{Center adaptation ($\mathbf{c}$):} To maximize information gain, we align $\mathbf{c}$ with the density center of the dataset:
\begin{equation}
	\mathbf{c}_i = \frac{1}{N} \sum_{n=1}^N \mathbf{x}_{i_n},(i = 1,\cdots,d).
\end{equation}

\item \textbf{Scale adaptation ($L$):} In MVRS, $L$ determines the field of vision. Instead of using a fixed $L$, we introduce randomness through the following strategy. Let $\sigma_{\mathcal{X}} = \max\limits_{d} \sigma^{(d)}$ be the maximum standard deviation across all variable values in the datasets. The $L_k$ for the $k$-th view is adapted as:
\begin{equation}
	L_k = 3.0 \cdot \sigma_{\mathcal{X}} \cdot \gamma_k,(k=1,\cdots,K).
\end{equation}
The constant factor $3.0$ follows the three-sigma rule~\cite{pukelsheim1994three}. $\gamma_k$ represents scale factors sampled from $\mathcal{U}[0.1, 10]$. 
\end{itemize}

\subsection{Visual Encoder}
\label{visual_encoder}

The Visual Encoder employs a standard ResNet~\cite{targ2016resnet} to map the input MVRS slices into continuous feature space. On the output, we remove the global average pooling layer to preserve the spatial structure of $M = 4 \times 4$ and then flatten them in a sequence (denoted as ${F}_{v}=\{{f}_{v_m}\}_{m=1}^M$). Quantization process is performed according to the following formula:
\begin{equation} \widetilde{f}_{v_m} := \arg\min_{e_s \in E} \left\| f_{v_m} - e_s \right\|_2,
\end{equation} 
where $\widetilde{F_{v}}=\{\widetilde{f}_{v_m}\}_{m=1}^M$ represent quantized visual features, $E=\{{e}_s\}_{i=1}^S$ is the Codebook containing $S$ learnable codewords, and ${\Vert \cdot \Vert}_2$ is the Euclidean distance. The quantization loss is described as:
\begin{equation}
	\label{loss_q}
	\begin{split}
		\mathcal{L}_{q} = {}& {\Vert \text{sg}[{F_{v}}]- \widetilde{F_{v}} \Vert}_2^2 + \beta_1{\Vert {F_{v}}-\text{sg}[\widetilde{F_{v}}] \Vert}_2^2 \\
		& + \beta_2 \sum_{s=1}^{S} \bar{p}_s \log (\bar{p}_s + \epsilon),
	\end{split}
\end{equation}
where $\text{sg}[\cdot]$ denotes the stop-gradient operator. The first term updates the Codebook to align with the Visual Encoder output, while the second term constrains the Visual Encoder output to commit to the Codebook. The third term is the diversity regularization term to prevent Codebook collapse. Here, $\bar{p}_s$ represents the average utilization probability of the $s$-th codeword in the batch. $\beta_1$ and $\beta_2$ are hyperparameters.

\subsection{Visual Decoder}
\label{visual_decoder}
The Visual Decoder establishes a mapping between the dataset and Codebook, enabling dataset-only inference. This module operates essentially as a parallel classifier. We initialize $M$ learnable classification tokens $[CLS]s$, corresponding to the $M$ visual features output by the Visual Encoder. $[CLS]s$ aggregates with the dataset features via a Cross-Attention module to generate $M$ classification vectors. Subsequently, these classification vectors are processed by a multi-head classifier to generate probability distributions of $S$ codewords. The training objective is the multi-class Cross-Entropy, called the visual prediction loss:
\begin{equation}
	\mathcal{L}_{vp} = -\frac{1}{M} \sum_{m=1}^{M} \log p_{m}(s_m^*),
	\label{loss_vp}
\end{equation}
where $s_m^*$ denotes the ground-truth codeword index corresponding to the $m$-th classification head. The codewords with the highest probability are selected as the predicted (virtual) visual features $\hat{F}_v$. 

To enable gradient backpropagation during training, we utilize a Gumbel-Softmax relaxation:
\begin{equation}
	\hat{f}_{v_m} = \sum_{s=1}^{S} \frac{\exp \Big( (\log p_{m,s} + g_s) / \tau \Big)}{\sum_{l=1}^{S} \exp \Big( (\log p_{m,l} + g_l) / \tau \Big)} e_s,
	\label{eq:gumbel}
\end{equation}
where $g$ is independent and identically distributed (i.i.d.) samples drawn from the Gumbel(0, 1) distribution, and $\tau$ is a temperature parameter. During inference, we set $\tau \to 0$ to approximate a hard categorical selection ($\arg\max$).

\subsection{Multimodal fusion and alignment}
\label{Multimodal SR module}

\begin{figure*}[t]
	\centering
	\includegraphics[width=1\textwidth]{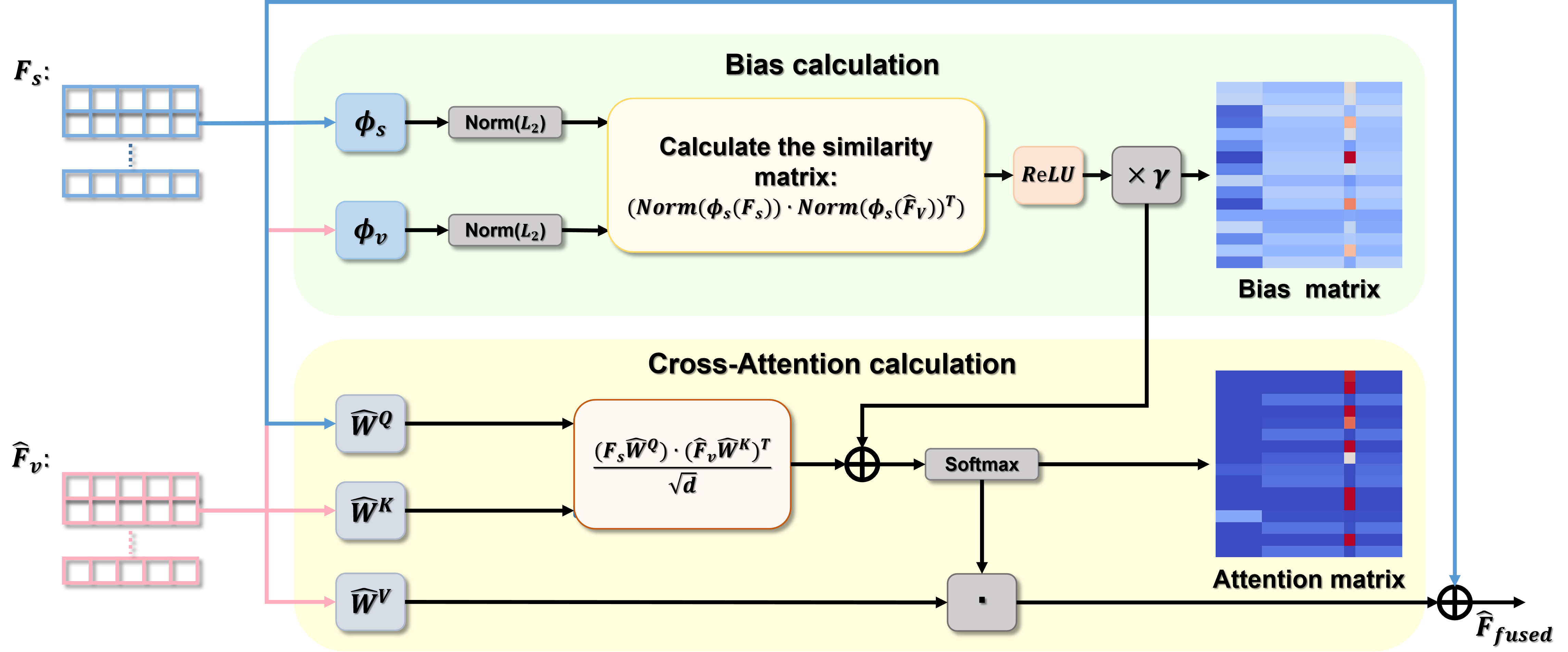}
	\caption{\textbf{Overview of the Biased Cross-Attention module.} $Norm$ denotes $L_2$ regularization. Only the top regions of the Bias and Attention matrices are shown for better visibility. Observed that the majority of the regions exhibit low attention scores. This phenomenon is intuitive, as it implies that each point attends to only a single or a limited number of visual features.}
	\label{biased_cross_attention}
\end{figure*}
Since the real visual pipeline has access to visual inputs during training, we adopt the standard Cross-Attention feature fusion module to fuse dataset features $F_{s}$ and real visual features $\widetilde{F_{v}}$, thereby fully utilizing the visual context. However, the virtual visual pipeline relies on virtual visual features $\hat{F}_{v}$ predicted by the Visual Decoder, which inherently contain noise. Directly feature fusion leads to modality contamination. Therefore, we introduce a Biased Cross-Attention module~\cite{press2021train}, as shown in Fig.~\ref{biased_cross_attention}. Departing from the standard paradigm, we incorporate a bias $B$ term into the attention scores to attenuate the attention paid to noise. The standard Cross-Attention and Biased Cross-Attention are respectively formulated as:
\begin{equation}
	\begin{split}\widetilde{F}_{\text{fused}} &= F_s + \text{Softmax}\left( \frac{(F_s \widetilde{W}^Q)(\widetilde{F}_v \widetilde{W}^K)^T}{\sqrt{d}} \right) (\widetilde{F}_v \widetilde{W}^V), \\ \hat{F}_{\text{fused}} &= F_s + \text{Softmax}\left( \frac{(F_s \hat{W}^Q)(\hat{F}_v \hat{W}^K)^T}{\sqrt{d}} + B \right) (\hat{F}_v \hat{W}^V)\end{split}.
\end{equation}

Specifically, we first project the dataset features $F_s$ and the virtual visual features $\hat{F}_v$ into a unified feature space. To balance feature preservation with noise suppression, we apply a gating function for the bias $B$:
\begin{equation}
	\label{eq:bias_matrix}
	\begin{split}
		B_{i,m} &= \gamma_{pos} \cdot \text{ReLU}\left( \text{sim}\left(\phi_s(f_{s_{i}}), \phi_v(\hat{f}_{v_{m}})\right) \right) \\
		&\quad - \gamma_{neg} \cdot \text{ReLU}\left( -\text{sim}\left(\phi_s(f_{s_{i}}), \phi_v(\hat{f}_{v_{m}})\right) \right),
	\end{split}
\end{equation}
where $\phi_s$ and $\phi_v$ are projection networks. $\gamma_{pos}$ and $\gamma_{neg}$ are learnable scaling factors initialized with $\lambda_{neg} \gg \lambda_{pos}$. $\text{sim}(\cdot,\cdot)$ denotes the cosine similarity. 

To make the bias $B$ effective, we introduce an alignment objective~\cite{liu2021variational}. For a pair of $(f_{s_i}, \hat{f}_{v_m})$, since the matching relationships are unavailable, we define the positive match score using the real visual features $\widetilde{F}_v$:
\begin{equation}
	S_{pos}(f_{s_i}, \widetilde{F}_v) = \max_{m} \left( \text{sim}(\phi_s(f_{s_{i}}),\phi_v(\widetilde{f}_{v_{m}})) \right),
\end{equation}
where $\phi_s$ and $\phi_v$ are weight-shared with Eq.~\ref{eq:bias_matrix}. ``Max" operation encourages each point feature $f_{s_i}$ to be closer to its most semantically relevant visual feature $\widetilde{f}_{v_{m}}$. We define the remaining codewords in the Codebook, excluding $\widetilde{F}_{v}$, as the negative set $\mathcal{V}_{neg}$. The alignment loss is defined as:

\begin{equation}
	\label{eq:align}
	\begin{aligned}
		\mathcal{L}_{align} = -\frac{1}{N} \sum_{i=1}^{N} \log & \left( \frac{\exp \left( S_{pos}(f_{s_i}, \widetilde{F}_v) / \tau \right)}
		{\begin{aligned} 
				&\exp \left( S_{pos}(f_{s_i}, \widetilde{F}_v) / \tau \right) \\ 
				&+ \sum_{e \in \mathcal{V}_{neg}} \exp \left( \text{sim}(f_{s_i}, e) / \tau \right) 
		\end{aligned}} \right).
	\end{aligned}
\end{equation}
where $\tau$ is the temperature coefficient. By minimizing $\mathcal{L}_{align}$, the similarity of matched pairs within the bias $B$ (Eq.~\ref{eq:bias_matrix}) is maximized, while that of mismatched pairs is minimized, ensuring that the Biased Cross-Attention module can distinguish between valid visual features and noise.
%\begin{equation}
%	\label{eq:align}
%	\mathcal{L}_{align} = - \frac{1}{N} \sum_{i=1}^{N} \log \frac{\exp \Big( S_{pos}(f_{s_i}, \widetilde{F}_v) / \tau \Big)}{Z_i},
%\end{equation}
%where  $\tau$ is the temperature coefficient, and $Z_i$ is the normalization factor covering both positive and negative pairs:
%\begin{equation}
%	Z_i = \exp \Big( S_{pos}(f_{s_i}, \widetilde{F}_v) / \tau \Big) + \sum_{e \in \mathcal{V}_{neg}} \exp \Big( \text{sim}(f_{s_i}, e) / \tau \Big).
%\end{equation}

\subsection{Decoder}
\label{decoder}

Following the design of \cite{biggio2021neural}, decoding is performed autoregressively using a Transformer decoder. The syntax follows the prefix of equation tree. 

However, we note that the standard autoregressive decoder often ignores syntax logic, leading to generate invalid equations. To address this problem, we integrate a syntax constraint algorithm (Algorithm~\ref{alg:syntax_constraint}) into the decoding process. Specifically, we maintain a stack to track the number of required child nodes and the inherited syntax constraints in current step. We enforce three constraints: Forcing the generation of variables or constants when the number of required child nodes is insufficient. Masking early stopping if the tree is incomplete. Prohibiting nested calls of transcendental or power functions by propagating a ban-set in the autoregressive process to ensure interpretability and numerical stability.

\begin{algorithm}[h]
	\caption{Syntax constraint algorithm.}
	\label{alg:syntax_constraint}
	\begin{algorithmic}[1]
		\Require Prefix sequence $S_{Prefix}$, Remaining budget $L_{rem}$.
		\Require Sets: $\mathcal{O}_{unary}:$ unary operators, $\mathcal{O}_{binary}:$ binary operators, $\mathcal{O}_{trans}:$ transcendental operators.
		
		\State \textbf{Definitions:} Stack item $(n, \Phi)$ denotes \textit{required child nodes} and \textit{inherited ban-set}.
		
		\State Initialize Stack $\mathcal{S} \gets [(1, \emptyset)]$
		
		\Statex \textcolor{blue}{\textbf{\textit{// Traversing prefix sequence to reconstruct the stack}}}
		\For{$w \in S_{Prefix}$} 
		\State $(n, \Phi) \gets \mathcal{S}.\text{Pop}()$ \Comment{Get parent's required child nodes}
		
		\If{$n > 1$} 
		 $\mathcal{S}.\text{Push}(n-1, \Phi)$ \Comment{Sibling node inherits parent's ban-set}
		\EndIf
		
		\State \textcolor{purple}{\textbf{\textit{// Update ban-set for child nodes}}}
		\State $\Phi' \gets \Phi$ 
		\If{$w \in \mathcal{O}_{trans}$} 
		$\Phi' \gets \Phi' \cup \mathcal{O}_{trans}$ \EndIf
		\If{$w = \text{'pow'}$} 
		$\Phi' \gets \Phi' \cup \{\text{'pow'}\}$ \EndIf
		
		\State \textcolor{purple}{\textbf{\textit{// Push required child nodes}}}
		\If{$w \in \mathcal{O}_{binary}$} 
		$\mathcal{S}.\text{Push}(2, \Phi')$ \Comment{Binary Operators require 2 nodes}
		\ElsIf{$w \in \mathcal{O}_{unary}$} 
		$\mathcal{S}.\text{Push}(1, \Phi')$ \Comment{Unary operators require 1 node}
		\EndIf
		\EndFor
		
		\Statex \textcolor{red}{\textbf{\textit{// Apply syntax constraints based on current step}}}
		\State Initialize mask $\mathbf{M} \gets \mathbf{0}$
		
		\State $N_{req} \gets \sum_{(n, \Phi) \in \mathcal{S}} n$ \Comment{Calculate total nodes required to complete tree}
		
		\State $\mathbf{M}[\mathcal{S}.\text{Peek}().\Phi]$ \Comment{Masking structurally forbidden operators}
		
		\If{$N_{req} \ge L_{rem}$} 
		\State $\mathbf{M}[\mathcal{O}_{unary} \cup \mathcal{O}_{binary}]$\Comment{Forcing leaf nodes} 
		\EndIf
		
		\If{$N_{req} > 0$} 
		\State $\mathbf{M}[\{\text{STOP}\}]$ \Comment{Prevent stopping}
		\EndIf
		
		\State \Return $\mathbf{M}$
	\end{algorithmic}
\end{algorithm}

\subsection{Optimization}

ViSymRe is optimized end-to-end through a multi-objective loss function. Since it contains two visual pipelines, the skeleton prediction loss consists of two components:
\begin{equation}
	\label{eq:sp}
	\mathcal{L}_{sp} =\mathcal{L}_{rsp} + \mathcal{L}_{vsp},
\end{equation}
where $\mathcal{L}_{rsp}$ and $\mathcal{L}_{vsp}$ optimize the real and virtual visual pipelines, respectively. Both are defined by Cross-Entropy:
\begin{equation}
	\mathcal{L}_{rsp} = - \sum_{t=1}^{T} \log P(\widetilde{y}_t | \widetilde{y}_{<t}) , \quad \mathcal{L}_{vsp} = - \sum_{t=1}^{T} \log P(\hat{y}_t | \hat{y}_{<t}).
\end{equation}
Here, $\widetilde{y}_t$ and $\hat{y}_t$ denote the ground-truth tokens at step $t$ given the preceding contexts $\widetilde{y}_{<t}$ and $\hat{y}_{<t}$. In essence, $\mathcal{L}_{rsp}$ serves as supervision for the Visual Encoder, encouraging it to output useful visual features.

Furthermore, considering the lag in optimizing the virtual visual pipeline relative to the real visual pipeline, which is mainly reflected in the inconsistency of the probability distributions output by their decoders. The causes for this phenomenon include noise in the predicted virtual visual features and the slower convergence of the Biased Cross-Attention feature fusion module compared to the standard paradigm. Therefore, we employ a KL divergence to define the consistency loss $\mathcal{L}_{cons}$, forcing the probability distributions output by the decoders of the two visual pipelines to be consistent: 
\begin{equation}
	\label{eq:cons}
	\mathcal{L}_{cons} = \sum_{t=1}^{T} D_{KL} \left( P(\widetilde{y}_t | \widetilde{y}_{<t}) \parallel P(\hat{y}_t | \hat{y}_{<t}) \right).
\end{equation}

The overall loss function is a weighted sum of the quantization loss $\mathcal{L}_{q}$ (Eq.~\ref{loss_q}), the visual prediction loss $\mathcal{L}_{vp}$ (Eq.~\ref{loss_vp}), the alignment loss $\mathcal{L}_{align}$ (Eq.~\ref{eq:align}), the skeleton prediction loss $\mathcal{L}_{sp}$ (Eq.~\ref{eq:sp}), and the consistency loss $\mathcal{L}_{cons}$ (Eq.~\ref{eq:cons}):
\begin{equation}
	\label{eq:loss_total}
	\mathcal{L}_{total} = \mathcal{L}_{sp} + \lambda (\mathcal{L}_{q} + \mathcal{L}_{vp} + \mathcal{L}_{align} + \mathcal{L}_{cons}).
\end{equation}
where $\lambda$ is a hyperparameter balancing the main and auxiliary losses.

\section{Data preprocessing}
\label{data preprocessing}
 
In this section, we detail the data preprocessing pipeline used for generating the training dataset for ViSymRe. This process is divided into three stages: First, generating diverse equation skeletons based on a binary tree; Second, sampling the constant in the skeleton to generate analytical equations; Finally, sampling points to generate the input datasets.
 
 \subsection{Skeleton generation}
 
 Our method for generating equation skeletons improves upon the public procedure~\cite{biggio2021neural}, which conceptualize the equation skeleton as a binary tree. The non-leaf nodes represent operators, and the leaf nodes represent variables or constants. Table~\ref {tab:equation_generator} lists all parameters used by our equation skeleton generator. Furthermore, we apply the same syntax constraints as the Algorithm ~\ref{alg:syntax_constraint} and add fractions to the exponent to enhance the interpretability and compatibility of the generated skeletons.

 \subsection{Dataset sampling}
 
 To improve the robustness of ViSymRe, we introduce variability by randomly sampling constants and points for each equation skeleton, ensuring that the sampled datasets cover a wide range of numerical scales. For each equation skeleton, we sample the constants according to the distribution in Table~\ref{tab:equation_generator} to generate the analytical equation. Then, we iteratively sample $N$ points for each variable and compute the target values to construct the datasets. To enhance the diversity of the numerical scales of the datasets, we employ the following hybrid sampling strategy: First, we randomly sample two constants from $\mathcal{U}(-10, 10)$ and sort them to determine $a$ and $b$. Then, we sample points from $\mathcal{U}(a, b)$ or $\text{Log-}\mathcal{U}(a, b)$ with equal probability to generate datasets. If any extreme value (NaN or Inf) is detected, it is set to 0.

\section{Experiments}
\label{experimental}

In this section, we present the training and inference details of ViSymRe.
Subsequently, we conduct extensive experiments on mainstream benchmarks to validate ViSymRe's advantages over existing competing baselines and its potential limitations.

\subsection{Training details}
During training, all resources are available, making ViSymRe a fully multimodal model. In a batch, the input consists of MVRS slices and a sampled dataset, with the target being the corresponding equation skeleton. Table~\ref{tb:parameters} reports the detailed hyperparameters used during training. As a joint optimization framework, the forward propagation process of ViSymRe is detailed in Algorithm~\ref{alg:training}. After training, the real visual pipeline is automatically disabled to ensure inference efficiency.
\begin{algorithm}[h]
	\caption{Multimodal training algorithm of ViSymRe.}
	\label{alg:training}
	\begin{algorithmic}[1]
		\Require Batch $\mathcal{B} = \{(X, y, I, E_{skeleton})\}$.
		\Require \textbf{Modules:}
		\Statex \quad $\bullet$ \textbf{Real visual pipeline:} Visual Encoder: $E_{v}$, Vector Quantizer: $VQ$, standard Cross-Attention: $\text{Attn}_{std}$.
		\Statex \quad $\bullet$ \textbf{Virtual visual pipeline:} Visual Decoder: $D_{v}$, Biased Cross-Attention: $\text{Attn}_{bias}$.
		\Statex \quad $\bullet$ \textbf{Shared module:} Dataset Encoder: $E_{s}$, Codebook: $E$, Transformer-decoder: $D_{d}$, IEEE-754 Encoder: $E_{ieee}$.
		
		\For{epoch $= 1,\cdots,50$}
		\For{$(X, y, I, E_{skeleton}) \in \mathcal{B}$}
		\State $F_{emb} \leftarrow E_{ieee}(X, y)$ \Comment{IEEE-754 encoding}
		\State $F_{s} \leftarrow E_{s}(F_{\text{emb}})$\Comment{Dataset encoding}
		\State \textcolor{blue}{\textbf{\textit{// Real visual pipeline}}}
		\State $F_{v} \leftarrow E_{v}(I)$
		\State $\widetilde{F}_v \leftarrow VQ(F_{v}, E)$ \Comment{Quantizing in Codebook}
		\State $\widetilde{F}_{fused} \leftarrow \text{Attn}_{std}(F_s, \widetilde{F}_v)$ \Comment{Standard Cross-Attention feature fusion}
		
		\State \textcolor{red}{\textbf{\textit{// Virtual visual pipeline}}}
		
		\State $P_{logits} \leftarrow D_{v}(F_{emb})$ \Comment{Predicting codeword indices}
		\State $s^* \leftarrow \text{argmax}(P_{logits})$; \quad $\hat{F}_v \leftarrow E(s^*)$ \Comment{Lookup in Codebook}
		\State $\hat{F}_{fused} \leftarrow \text{Attn}_{bias}(F_s, \hat{F}_v)$ \Comment{Biased Cross-Attention feature fusion}
		
		\State  \textcolor{green}{\textbf{\textit{// Symbolic decoding \& Loss calculation}}}
		\State $\widetilde{Y} \leftarrow D_{d}(\widetilde{F}_{fused})$; \quad $\hat{Y} \leftarrow D_{d}(\hat{F}_{fused})$
		
		\State 	$\mathcal{L}_{total} = \mathcal{L}_{sp} + \lambda (\mathcal{L}_{q} + \mathcal{L}_{vp} + \mathcal{L}_{align} + \mathcal{L}_{cons})$ \Comment{Calculate total loss by Eq.~\ref{eq:loss_total}}
		\State Update parameters w.r.t $\nabla \mathcal{L}_{total}$
		\EndFor
		\EndFor
	\end{algorithmic}
\end{algorithm}

\subsection{Testing details}

ViSymRe faces the challenge of out-of-distribution robustness due to the sparsity of the training datasets. To ensure that ViSymRe can adapt to extremely large or small numerical scales that may appear in the test datasets (such as in the field of astronomy and nuclear physics), we propose a median-based \textbf{A}daptive \textbf{M}agnitude \textbf{S}caling (AMS) algorithm, aiming to scale the test datasets to a scale familiar to ViSymRe. Specifically, consider a $d$-D test dataset. For $i$-th feature $x_i$, we determine a magnitude $\rho_{x_i}$ based on the median of its absolute values:
\begin{equation}
	\rho_{x_i} = \text{median}(|x_i|).
\end{equation}
The scaling factor $s_i$ is then derived logarithmically to capture the order of magnitude:
\begin{equation}
	s_{x_i} = \begin{cases}
		10^{\lfloor \log_{10} \rho_{x_i} \rfloor} & \text{if } \left| \lfloor \log_{10} {\rho}_{x_i} \rfloor \right| \ge 1 \\
		1 & \text{otherwise}.
	\end{cases}
\end{equation}
Finally, $x_i$ is transformed via $\tilde{x_i} = {x_i} / s_{x_i}$. 

Through ablation studies, we demonstrate that the proposed AMS is effective in enhancing the out-of-domain extrapolation capability of ViSymRe compared to traditional Z-Score and Min-Max methods. The details are discussed further in~\ref{analysis_AMS}.

After scaling, we apply a bagging strategy to the test datasets, inspired by~\cite{kamienny2022end}, to avoid performance degradation of the Dataset Encoder when the test datasets are significantly larger than the training datasets. For each bag, we generate multiple candidate equation skeletons via beam search, and then optimize the constants in these skeletons using the BFGS algorithm. Once the analytical equations are obtained, we further process the optimized constants using a rounding strategy to ensure sparsity: negligible coefficients are pruned to 0, and the remaining coefficients are rounded to the nearest integer or a simple rational number (such as 2.4999 to 2.5).
Algorithm~\ref{alg:inference} demonstrates the detailed inference process of ViSymRe. All relevant parameters can be found in Table~\ref{tb:parameters}.  
\begin{algorithm}[h]
	\caption{Inference algorithm of ViSymRe.}
	\label{alg:inference}
	\begin{algorithmic}[1]
		\Require Test dataset $(X, y)$, Number of bags $N_{bag}$.
		\Require ViSymRe $\mathcal{M}$, Error threshold $\epsilon$.
		
		\State Initialize best solution $E^* \gets \emptyset$, min error $\mathcal{L}_{min} \gets \infty$.
		
		\State $\tilde{X}, \tilde{y} \leftarrow \text{AMS.Transform}(X, y)$ \Comment{Mapping test dataset to numerical stable domain}
		
		\For{$i = 1,\cdots,N_{bag}$}
		
		\State $\tilde{X}_{sub}, \tilde{y}_{sub} \leftarrow \text{RandomSample}((\tilde{X},\tilde{y}))$ \Comment{Bagging on scaled test dataset}
		
		\State $E_{candidates} \leftarrow \text{BeamSearch}_{\mathcal{M}}((\tilde{X}_{sub}, \tilde{y}_{sub}), \text{BeamSize}=\min(30, 3*i))$ \Comment{Beam search with dynamic beam size }
		
		\For{$E_{skeleton} \in E_{candidates}$}
		\If{$E_{skeleton}$ has constants}
		
		\State Optimize constants $\tilde{\theta}$ in $E_{skeleton}$ via BFGS. 
		\State Update equation $\tilde{E} \leftarrow E_{skeleton}(\tilde{\theta}^*)$ \Comment{Constant pruning and rounding}
		\EndIf
		
		\State $E \leftarrow \text{AMS.Restore}(\tilde{E}, \mathcal{S})$ \Comment{Symbolic recovery}
		
		\State Calculate Error $\mathcal{L}_{curr} \leftarrow \text{MSE}(E, X, y)$ \Comment{Evaluate fitnes}
		
		\If{$\mathcal{L}_{curr} < \mathcal{L}_{min}$}
		\State $E^* \leftarrow E$; \quad $\mathcal{L}_{min} \leftarrow \mathcal{L}_{curr}$
		\EndIf
		\EndFor
		
		\If{$\mathcal{L}_{min} < \epsilon$} \State \textbf{break} \EndIf 
		\EndFor
		
		\State \Return $E^*$
	\end{algorithmic}
\end{algorithm}
\subsection{Benchmarks}
To comprehensively evaluate the performance of ViSymRe in various SR scenarios, we select the following widely used benchmarks:
\begin{itemize}
	\item \textbf{Low-dimensional benchmarks:} The low-dimensional benchmarks include "Nguyen"~\cite{uy2011semantically}, "Keijzer"\cite{keijzer2003improving}, "Korns"~\cite{korns2011accuracy},  "Livermore"~\cite{petersen2019deep}, "Jin"~\cite{jin2019bayesian} and "Neat". These benchmarks contain equations with up to three variables. We randomly sample 200 points from the sampling range reported in~\ref{tab:Details on low-dimensional benchmarks} to generate the test datasets.
	
	\item \textbf{SRBench 1.0~\cite{la2021contemporary}:} We select 133 ground-truth problems from the following two benchmarks in SRBench 1.0:
	\begin{itemize}
		\item \textbf{Feynman benchmark~\cite{udrescu2020ai}:} The Feynman benchmark is derived from first-principle models of physical systems. It consists of 119 equations from the Feynman Lectures on Physics~\cite{feynman2015feynman}, each paired with a corresponding sampled dataset.
		
		\item \textbf{ODE-Strogatz benchmark~\cite{strogatz2018nonlinear}:} The ODE-Strogatz benchmark includes a variety of nonlinear dynamical systems, some of which exhibit chaotic behavior. It comprises 14 datasets, each of which samples the time evolution of a two-dimensional system governed by coupled first-order ordinary differential equations.
	\end{itemize}
	
	\item \textbf{SRSD-Feynman benchmark~\cite{matsubara2022srsd}}: SRSD-Feynman is an improved version of the Feynman benchmark that redefines the sampling ranges of coefficients and variables to reflect typical experimental settings and physical scales more accurately, including more challenging extreme values.
	
	\item \textbf{SRBench 2.0~\cite{aldeia2025call}:} The SRBench 2.0 contains 12 black-box problems and 12 phenomenological \& first-principles problems:
	\begin{itemize}
		\item \textbf{Black-box benchmark:} The black-box benchmark comprises 12 diverse regression datasets rigorously curated from PMLB~\cite{romano2022pmlb}. To ensure a challenging benchmark, the authors excluded the simple problems that can be solved by linear regression.
		
		\item \textbf{Phenomenological \& first-principles benchmark:} This benchmark evaluates scientific discovery tasks using datasets published by Russeil et al.~\cite{russeil2024multiview} and Cranmer~\cite{cranmer2023interpretable}. It comprises tasks derived from empirical measurements and fundamental physical equations with realistic noise.
	\end{itemize}

\end{itemize}

\subsection{Baselines}
We compare ViSymRe with the following baselines, including recently proposed dataset-only Transformer-based models and several representative GP-based and DL-based models:

\begin{itemize}
	
	\item \textbf{NeSymReS~\cite{biggio2021neural}:} A Transformer-based architecture that represents the classic paradigm of pre-trained SR models. It treats the components of equations, including operators, constants, and variables, as tokens, enabling it to learn a mapping from the dataset to the equation skeleton.
	
	\item \textbf{TPSR~\cite{shojaee2023transformer}:} It combines Monte Carlo Tree Search (MCTS) as a rectification strategy to the classic pre-trained paradigms, to enhance their generalization ability.
	
	\item \textbf{PYSR~\cite{cranmer2023interpretable}:} The PYSR represents a new trend in GP-based SR models. It employs a multi-objective, highly parallel evolutionary algorithm and integrates a noise filter, aiming to achieve robust, efficient, and interpretable SR.
	
	\item  \textbf{uDSR~\cite{landajuela2022unified}:} uDSR proposes a modular, unified framework that hybridizes five distinct strategies, including recursive simplification, DL, large-scale pre-trained, GP, and linear models, thereby integrating these existing mainstream approaches into a unified system.
	
	\item \textbf{RILS-ROLS~\cite{kartelj2023rils}:} RILS-ROLS is a metaheuristic-based approach that combines iterated local search and least squares to solve the symbolic regression problem, offering robust performance under varying SR scenarios.
	
	\item \textbf{SRBench Baselines~\cite{la2021contemporary}:} The 14 open-source SR models reported in SRBench, which represent the mainstream performance.
	
\end{itemize}
All parameters used by NeSymReS, TPSR and SRBench baselines are set to the default values provided by the authors. For PYSR, uDSR, and RILS-ROLS, refer to~\ref{parameters for baselines} for parameter settings. 

\subsection{Metrics}
In our experiments, there are three metrics to evaluate the performance of ViSymRe:

\begin{itemize}
	\item \textbf{$R^2$ score:} $R^2$ score measures how well the generated equation fits the dataset. Its formula is described as:
	\begin{equation}
		R^2 = 1 - \frac{\sum_{i=1}^n (y_i - \hat{y}_i)^2}{\sum_{i=1}^n (y_i - \overline{y})^2},
	\end{equation}
	If $R^2 \geq 0.999$, the generated equation is considered to fit the dataset well.
	
	\item \textbf{Complexity:} Complexity is the number of nodes in equations. If an equation fits the dataset well and is simple, it is considered to be a good equation.
	
	\item \textbf{Symbolic Solution:} Symbolic Solution aims to identify the generated equation that differs from the ground-truth one by a constant or scalar~\cite{la2021contemporary}. It is defined as follow:
	
	\begin{definition}
		A generated equation $\hat{f}$ is considered as a Symbolic Solution to a problem with the ground-truth equation $f+\epsilon$, if $\hat{f}$ does not simplify to a constant, and at least one of the following conditions holds:
		\begin{itemize}
			\item $f = \hat{f}+\alpha$, where $\alpha$ is constant;
			\item  $f = \hat{f}/\beta$, where $\beta$ is constant and $\beta \neq 0$.
		\end{itemize}
	\end{definition}
	Compared to the $R^2$ score, which focuses on numerical approximation, the Symbolic Solution provides a more comprehensive assessment of the models' ability to capture the ground-truth equation structure. 
	
\end{itemize}

\subsection{Results on low-dimensional benchmarks}

Considering that scientific discovery often aims to distill complex natural phenomena into simple equations that involve only a few key variables, low-dimensional scenarios serve as a cornerstone for validating the ability of SR models to recover fundamental physical laws. Therefore, we first evaluate ViSymRe on low-dimensional benchmarks. Each experiment is repeated 10 times and averaged to ensure statistical significance.

In Table~\ref{tab:r2_Complexity_results}, we report the $R^2$ score and Complexity of the models, which are considered the fundamental metrics for evaluating SR performance. It can be seen that ViSymRe achieves $R^2 > 0.99$ across all benchmarks, showing robust numerical approximation capability. Furthermore, ViSymRe achieves lower Complexity than the baselines while maintaining such a high $R^2$ score, demonstrating its tendency to describe datasets with simple equations, an important aspect for interpretable scientific discovery. In comparison, NeSymReS also generates simple equations, ranking second in Complexity, but its $R^2$ score drops on several benchmarks. PYSR, uDSR, and TPSR are competitive in terms of $R^2$ score, but fail to simplify the equations effectively. 
\begin{table*}[h]
	\centering
	\caption{The $R^2$ score and Complexity results on low-dimensional benchmarks. The best and second-best results are marked in \textbf{bold} and \underline{underlined}, respectively.}
	\label{tab:r2_Complexity_results}
	\renewcommand{\arraystretch}{1.3}
	\setlength{\tabcolsep}{4pt}
	\resizebox{\textwidth}{!}{
		\begin{tabular}{l | c | c c c c c c}
			\toprule
			\textbf{Benchmarks} & \textbf{Metric} & \textbf{ViSymRe} & \textbf{uDSR} & \textbf{NeSymReS} & \textbf{PYSR} & \textbf{RILS-ROLS} & \textbf{TPSR} \\
			\midrule
			\multirow{2}{*}{Jin} 
			& $R^2$ & \underline{0.9955} $\pm$ 3.2e-3 & 0.9946 $\pm$ 9.8e-4 & 0.9018 $\pm$ 1.3e-2 & \textbf{0.9999} $\pm$ 1.2e-6 & 0.9614 $\pm$ 2.5e-4 & \textbf{0.9999} $\pm$ 5.2e-5 \\
			& Compl. & \textbf{13.37} $\pm$ 0.63 & 28.83 $\pm$ 2.83 & \underline{13.50} $\pm$ 0.63 & 14.72 $\pm$ 0.19 & 22.27 $\pm$ 4.27 & 22.50 $\pm$ 2.05 \\
			\midrule
			
			\multirow{2}{*}{Keijzer} 
			& $R^2$ & 0.9909 $\pm$ 2.3e-4 & 0.9948 $\pm$ 5.0e-5 & 0.8745 $\pm$ 1.1e-2 & 0.9990 $\pm$ 1.4e-4 & \underline{0.9996} $\pm$ 1.0e-5 & \textbf{0.9997} $\pm$ 1.5e-4 \\
			& Compl. & \underline{11.45} $\pm$ 0.93 & 19.73 $\pm$ 1.73 & \textbf{10.76} $\pm$ 0.47 & 12.92 $\pm$ 2.06 & 15.46 $\pm$ 1.76 & 20.35 $\pm$ 0.49 \\
			\midrule
			
			\multirow{2}{*}{Korns} 
			& $R^2$ & \textbf{0.9999} $\pm$ 4.6e-5 & 0.9916 $\pm$ 1.9e-4 & 0.8200 $\pm$ 2.6e-2 & \underline{0.9993} $\pm$ 1.4e-6 & 0.9779 $\pm$ 4.0e-6 & 0.9982 $\pm$ 4.1e-4 \\
			& Compl. & \textbf{8.55} $\pm$ 0.99 & 25.38 $\pm$ 2.38 & 11.76 $\pm$ 0.43 & \underline{11.71} $\pm$ 3.21 & 15.52 $\pm$ 1.78 & 16.77 $\pm$ 1.64 \\
			\midrule
			
			\multirow{2}{*}{Livermore} 
			& $R^2$ & \underline{0.9962} $\pm$ 3.7e-3 & 0.9890 $\pm$ 4.5e-6 & 0.9238 $\pm$ 1.8e-2 & \textbf{0.9999} $\pm$ 6.6e-5 & 0.9529 $\pm$ 2.7e-3 & 0.9716 $\pm$ 5.7e-3 \\
			& Compl. & \textbf{9.12} $\pm$ 0.39 & 15.09 $\pm$ 2.51 & \underline{9.48} $\pm$ 0.64 & 11.93 $\pm$ 1.30 & 13.26 $\pm$ 0.86 & 23.72 $\pm$ 0.56 \\
			\midrule
			
			\multirow{2}{*}{Neat} 
			& $R^2$ & 0.9901 $\pm$ 5.5e-3 & \underline{0.9974} $\pm$ 1.1e-3 & 0.7638 $\pm$ 2.3e-1 & \textbf{0.9998} $\pm$ 2.2e-4 & 0.9626 $\pm$ 4.6e-4 & \underline{0.9974} $\pm$ 7.1e-4 \\
			& Compl. & \textbf{10.50} $\pm$ 0.99 & 18.37 $\pm$ 2.37 & \textbf{10.50} $\pm$ 1.01 & 17.66 $\pm$ 2.22 & \underline{17.54} $\pm$ 2.54 & 21.83 $\pm$ 1.66 \\
			\midrule
			
			\multirow{2}{*}{Nguyen} 
			& $R^2$ & \underline{0.9999} $\pm$ 1.6e-4 & \textbf{1.0000} $\pm$ 0.0 & 0.9678 $\pm$ 5.2e-3 & \underline{0.9999} $\pm$ 3.1e-5 & \underline{0.9999} $\pm$ 4.7e-6 & 0.9998 $\pm$ 9.8e-5 \\
			& Compl. & \textbf{9.45} $\pm$ 0.6 & 12.47 $\pm$ 0.79 & \underline{10.19} $\pm$ 1.04 & 12.68 $\pm$ 0.05 & 13.16 $\pm$ 0.56 & 21.08 $\pm$ 0.52 \\
			\midrule
			
			\multirow{2}{*}{Nguyen'} 
			& $R^2$ & \textbf{1.0000} $\pm$ 0.0 & \textbf{1.0000} $\pm$ 0.0 & 0.9957 $\pm$ 2.7e-3 & \underline{0.9999} $\pm$ 9.6e-6 & \textbf{1.0000} $\pm$ 0.0 & 0.9998 $\pm$ 8.8e-5 \\
			& Compl. & \underline{8.56} $\pm$ 0.14 & 8.91 $\pm$ 0.29 & \textbf{7.91} $\pm$ 0.63 & 8.66 $\pm$ 0.76 & 11.50 $\pm$ 0.00 & 14.75 $\pm$ 3.90 \\
			\midrule
			
			\multirow{2}{*}{Nguyen$^c$} 
			& $R^2$ & \textbf{0.9999} $\pm$ 1.3e-6 & \textbf{0.9999} $\pm$ 1.7e-6 & 0.9990 $\pm$ 5.7e-4 & \textbf{0.9999} $\pm$ 4.5e-6 & \textbf{0.9999} $\pm$ 5.3e-9 & \textbf{0.9999} $\pm$ 4.1e-5 \\
			& Compl. & \textbf{8.26} $\pm$ 0.42 & 10.80 $\pm$ 0.20 & 10.53 $\pm$ 1.15 & \underline{9.60} $\pm$ 0.46 & 9.86 $\pm$ 0.16 & 12.73 $\pm$ 1.10 \\
			
			\bottomrule
		\end{tabular}
	}
\end{table*}

\begin{figure}[h]
	\centering
	\includegraphics[width=1\linewidth]{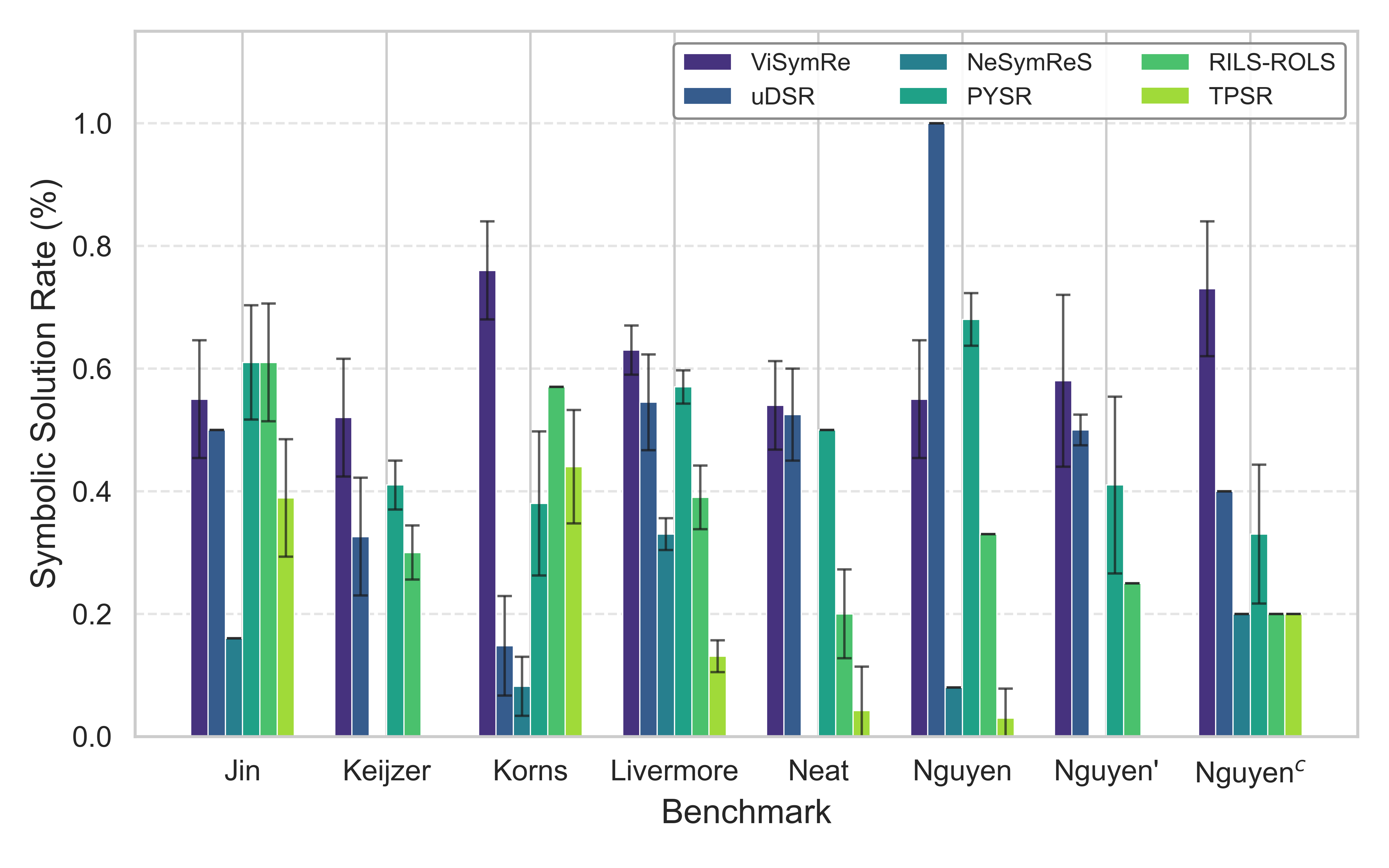}
	\caption{\textbf{The Symbolic Solution Rate results on low-dimensional benchmarks.} Missing results denote a Symbolic Solution Rate of 0. }
	\label{fig:SSR}	
\end{figure}

In scientific discovery, the objective of SR is not limited to numerical approximation, but to accurately reconstruct the ground-truth equation structure. Fig.~\ref{fig:SSR} presents the Symbolic Solution Rate, a metric quantifying this capability. As we can see, ViSymRe obtains the highest Symbolic Solution Rates on most benchmarks, including Korns, Livermore, Neat, and Nguyen$^c$. While baselines like uDSR, PYSR, and RILS-ROLS achieve high Symbolic Solution Rates on Nguyen and Jin benchmarks, they exhibit volatility, failing to generalize across different scenarios. 

Table~\ref{tab:test_time} compares the testing time of models. Observed that ViSymRe achieves the shortest testing time, outperforming the standard pre-trained baseline NeSymRes by a factor of four. Besides the convergence level, a possible explanation is that the syntax constraint algorithm promotes ViSymRe to generate simpler equation skeletons, enabling faster constant optimization, thereby triggering early stopping more effectively than NeSymReS. Furthermore, RILS-ROLS requires less testing time than PYSR and uDSR, reflecting that the evolution of equations via perturbation may be more controllable when dealing with low-complexity scenarios than population-based methods.

\begin{table}[h]
	\centering
	\renewcommand{\arraystretch}{1.3}
	\caption{The test time results on low-dimensional benchmarks.}
	\label{tab:test_time}

	\begin{tabularx}{\columnwidth}{@{} >{\centering\arraybackslash}X >{\centering\arraybackslash}X @{}}
		\toprule
		\textbf{Model} & \textbf{Test Time (s)} \\
		\midrule
		ViSymRe   & 7.03 \\
		PYSR      & 74.92 \\
		NeSymReS  & 28.77 \\
		uDSR      & 82.76 \\
		TPSR      & 136.46 \\
		RILS-ROLS & 10.43 \\
		\bottomrule
	\end{tabularx}
\end{table}

\begin{figure}[h]
	\centering
\includegraphics[width=1\linewidth]{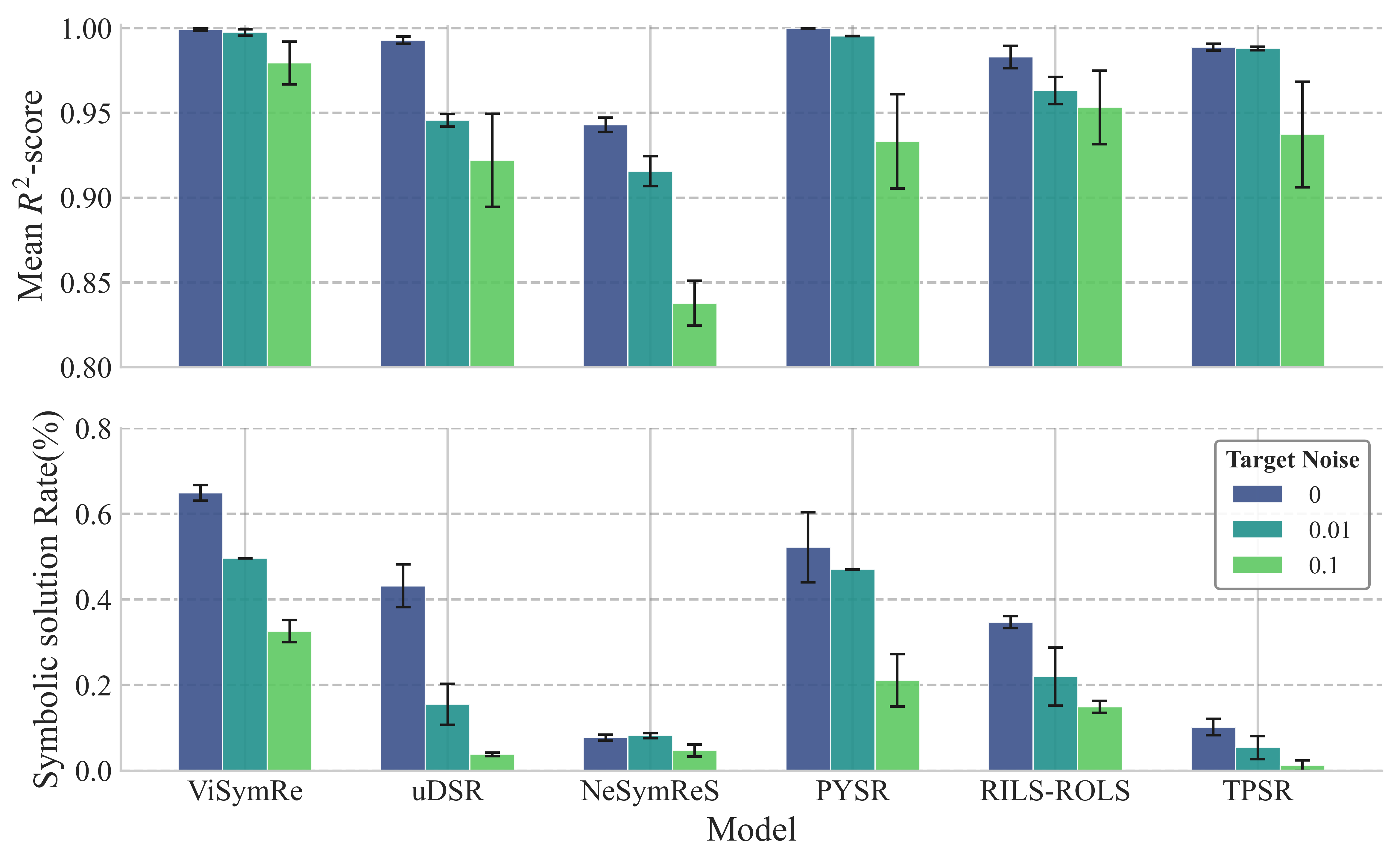}
	\caption{\textbf{Robustness of models under varying noise levels.} Results are averaged over 8 low-dimensional benchmarks at three noise intensities (0, 0.01, and 0.1).}
	\label{fig:noise_robustness}
\end{figure}

Real-world experimental datasets often contain noise, which may mislead models into overfitting complex equations rather than recovering the physical laws. To evaluate the noise robustness, we tracked the mean $R^2$ score and Symbolic Solution Rate across all low-dimensional benchmarks under varying noise levels, as visualized in Fig.~\ref{fig:noise_robustness}. Overall, higher noise degrades all models, but ViSymRe exhibits robustness. It maintains the $R^2>0.98$ even at the highest noise level, whereas baselines like NeSymReS show a marked deterioration. In terms of Symbolic Solution Rate, ViSymRe also outperforms all baselines. While the Symbolic Solution Rates of uDSR and TPSR collapse to near 0\% under intense noise, ViSymRe shows only a smooth and gradual decline. 

\begin{figure}[h]
	\centering
	\includegraphics[width=\linewidth]{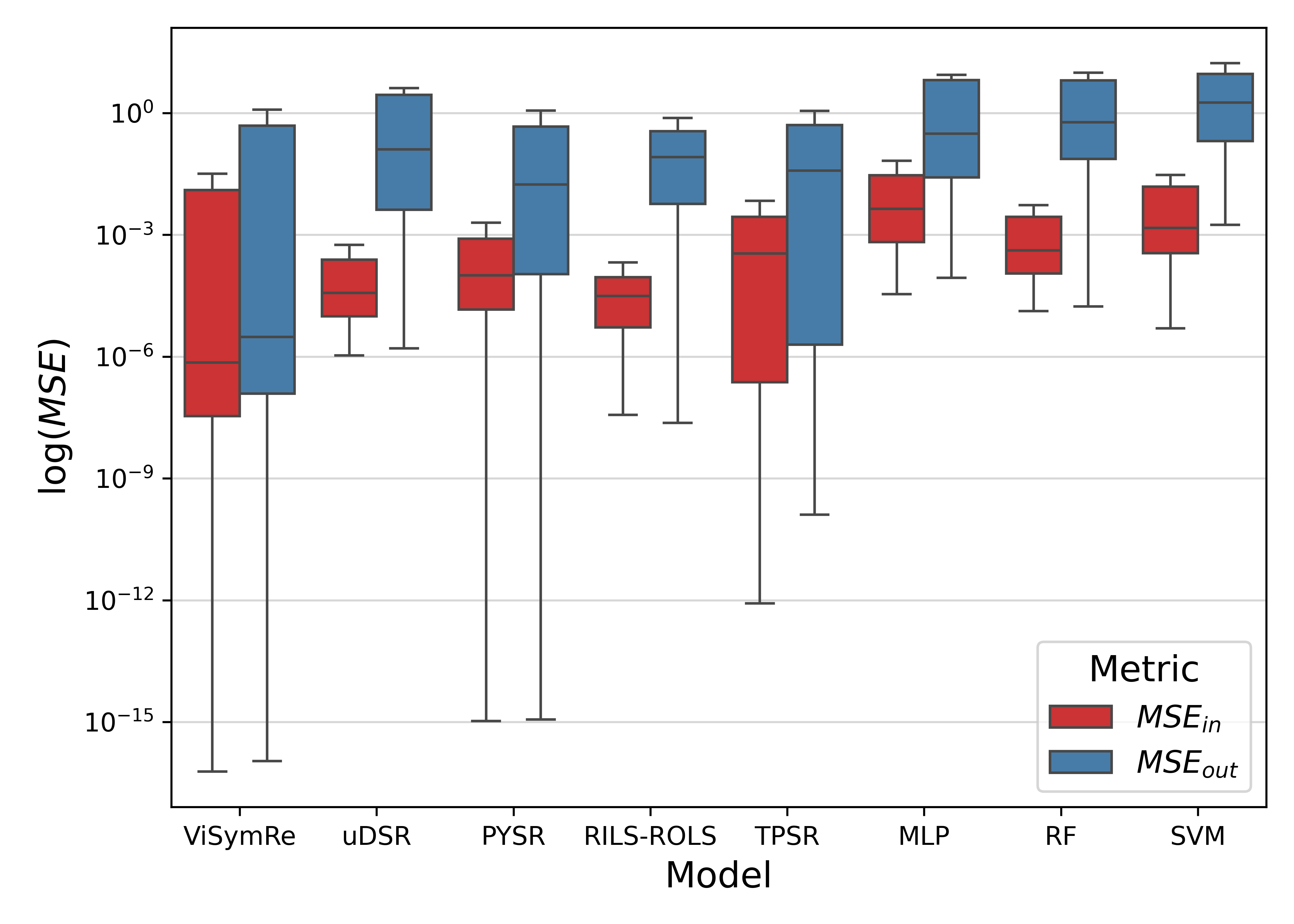}
	\caption{\textbf{The extrapolation evaluation results of models.} Noise is introduced into the training datasets, thereby implicitly evaluating how overfitting to noise compromises the models' extrapolation ability. Traditional ML models exhibit significantly higher extrapolation errors compared to SR models.}
	\label{fig:extrapolation_results}
\end{figure}

To evaluate the generalization ability of the equations generated by the models, we conducted extrapolation tests. Specifically, we construct the out-of-domain datasets by doubling the sampling range of the in-domain datasets. All models generate equations based on in-domain datasets and are tested for extrapolation on out-of-domain datasets. To demonstrate the superiority of the SR model in extrapolation scenarios, three classic machine learning models, including Multilayer Perceptron (MLP), Random Forest (RF), and Support Vector Machine (SVM), are introduced as supplementary baselines (detailed parameter settings are provided in Table~\ref{tab:PYSR_UDSR_RILS_ROLS_ML_parameters}). To enhance experimental realism, we introduce noise at a level of 0.01 to the in-domain datasets. Fig.~\ref{fig:extrapolation_results} presents the in-domain error distributions and out-of-domain error distributions of models. It can be observed that while the classic machine learning baselines achieve outstanding numerical approximation results on the in-domain datasets, their out-of-domain errors increased significantly. This phenomenon highlights the well-known tendency of them to overfit the local distribution of in-domain datasets rather than learning global laws. In contrast, SR models demonstrate superior extrapolation robustness. ViSymRe exhibits low median and peak errors across both in-domain and out-of-domain datasets, a significant achievement given the noisy condition. While other SR models also outperform classic machine learning models, they tend to minimize error by overfitting noise, thereby compromising their extrapolation potential, as evidenced by median out-of-domain errors that are significantly higher than that of in-domain errors. 

\subsection{Results on SRBench 1.0}
In this section, we evaluate ViSymRe on Feynman and ODE-Strogatz benchmarks. For the SRBench baselines, we utilize the results reported by the authors~\cite{la2021contemporary}.

\begin{figure}[h]
	\centering
	\includegraphics[width=1\linewidth]{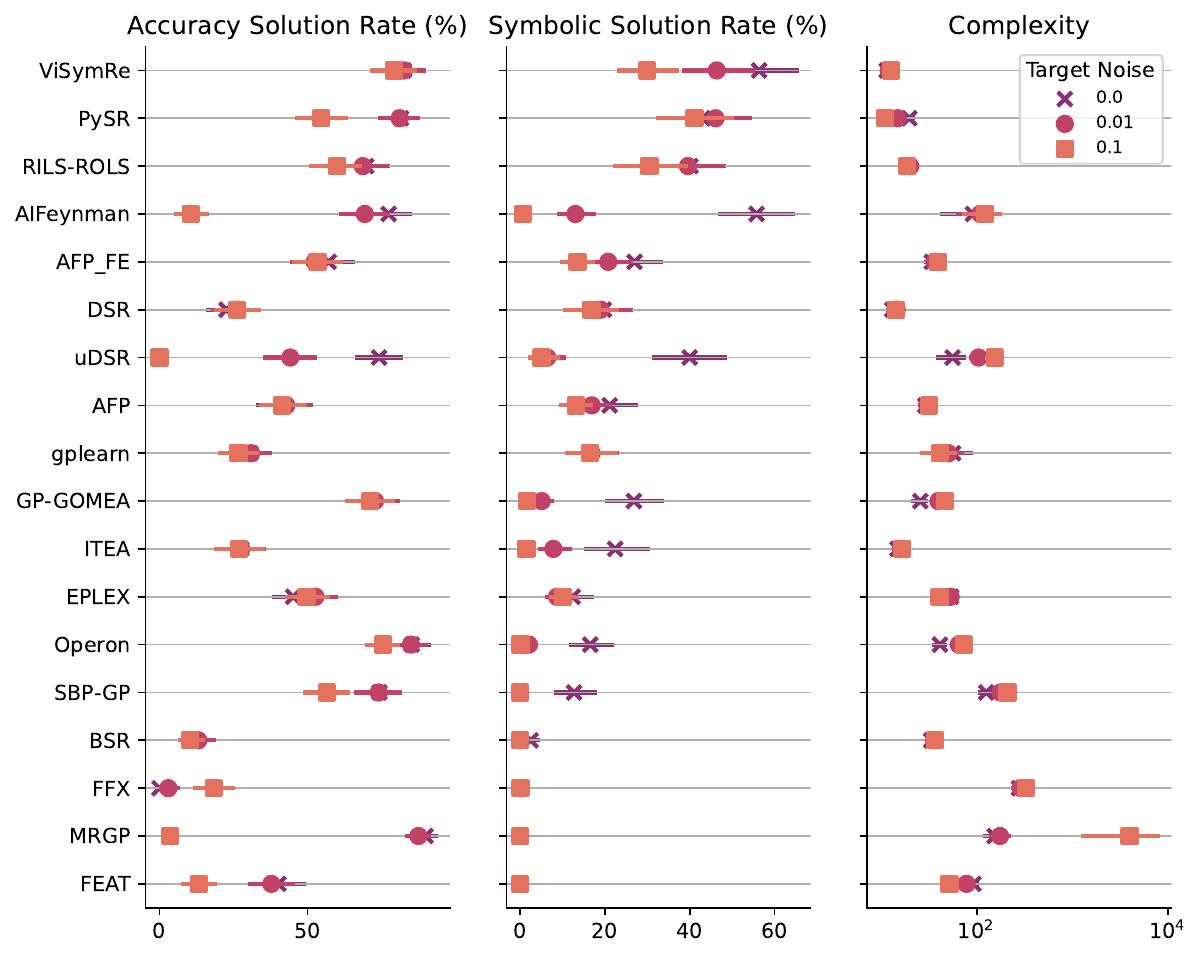}
	\caption{\textbf{The results of models on Feynman benchmark under three noise levels.} Each model is ranked based on three metrics. Overall, ViSymRe demonstrates a competitive advantage in balancing Accuracy Solution Rate, Symbolic Solution Rate, and Complexity, regardless of noise levels.
	}\label{fig:feynman_results}
\end{figure}

Fig.~\ref{fig:feynman_results} presents the evaluation results on the Feynman benchmark. Under low noise levels, ViSymRe, PYSR, uDSR, and RILS-ROLS achieve comparable Accuracy Solution Rate, which is on par with the SOTA baselines reported in SRBench 1.0, including GP-GOMEA, Operon, and MRGP. However, as noise increases, ViSymRe maintains a leading Accuracy Solution Rate, demonstrating superior stability. PYSR, uDSR, RILS-ROLS, and MRGP all experience varying degrees of performance degradation. In terms of Symbolic Solution Rate, ViSymRe, PYSR, and RILS-ROLS are in the top tier, even at a noise level of 0.1, their Symbolic Solution Rates remain above 25\%. uDSR and AI Feynman rank highly in Symbolic Solution Rate only under low noise levels but cannot maintain this position. Concerning Complexity, ViSymRe, PYSR, and RILS-ROLS consistently generate simple equations regardless of noise levels, while other models struggle to strike a balance between complexity and numerical approximation. 

\begin{figure}[h]
	\centering
	\includegraphics[width=1\linewidth]{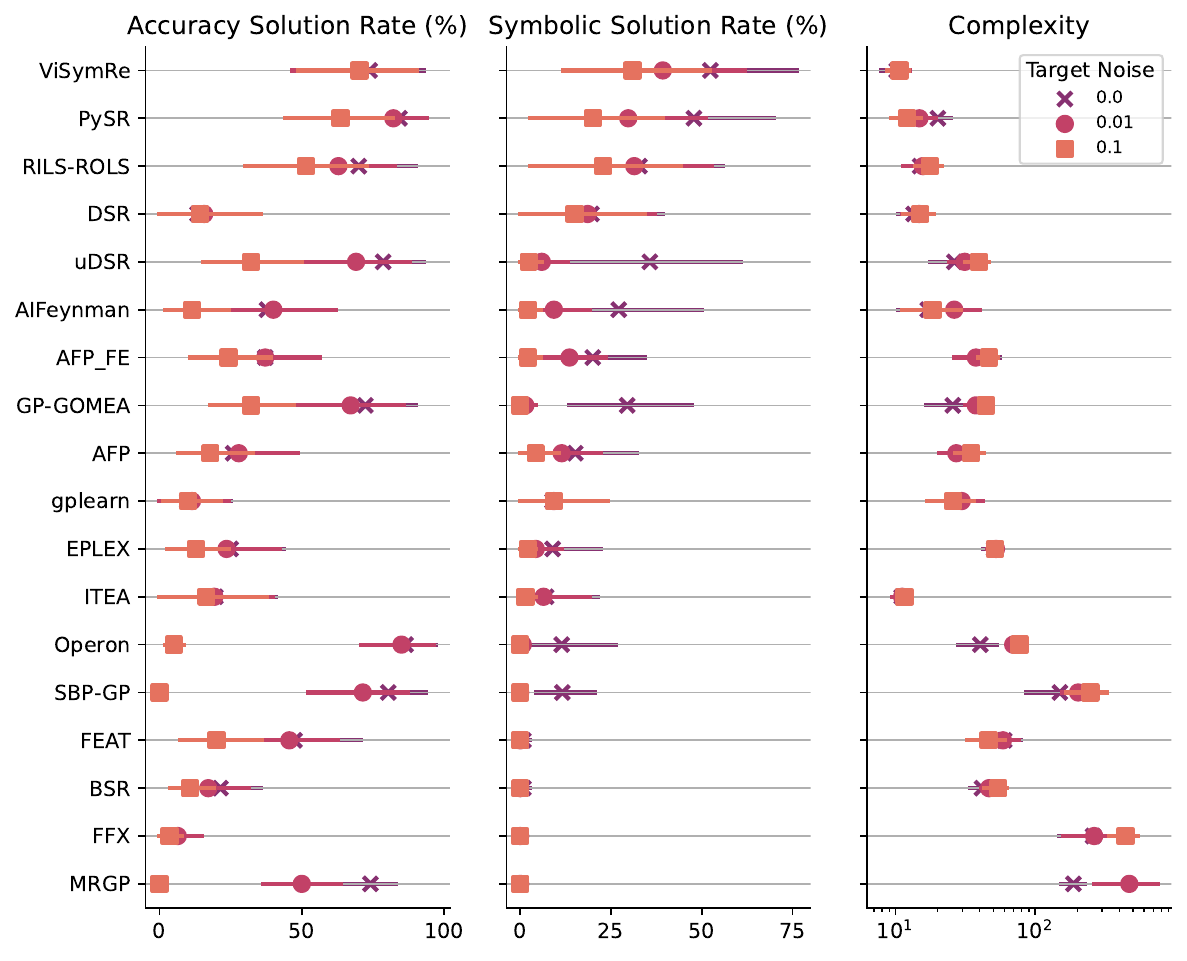}
	\caption{\textbf{The results of models on ODE-Strogatz benchmark under three noise levels} Despite the distributional gap between the ODE-Strogatz benchmark and the training datasets, ViSymRe exhibits consistent robustness across the three metrics, particularly under high noise levels.
	}\label{fig:ode-strogaz_results}
\end{figure}

The datasets in the ODE-Strogatz benchmark exhibit temporal correlation, which contradicts the uniform distribution sampled by the training datasets used by ViSymRe. Therefore, the ODE-Strogatz benchmark challenges ViSymRe's generalization ability. As shown in Fig.~\ref {fig:ode-strogaz_results}, ViSymRe maintained its overall leading ranking, but its performance fluctuated significantly, which validates our hypothesis that out-of-distribution testing leads to unstable performance of pre-trained models. Although PYSR demonstrates advantages in Accuracy Solution Rate, its Symbolic Solution Rate and Complexity are inferior to ViSymRe. Furthermore, compared to other models, ViSymRe exhibits superior noise robustness, showing no significant decrease in its metrics as noise increases.

\begin{figure}[h]
	\centering
	\begin{subfigure}{1\columnwidth}
		\centering
		\includegraphics[width=1\textwidth]{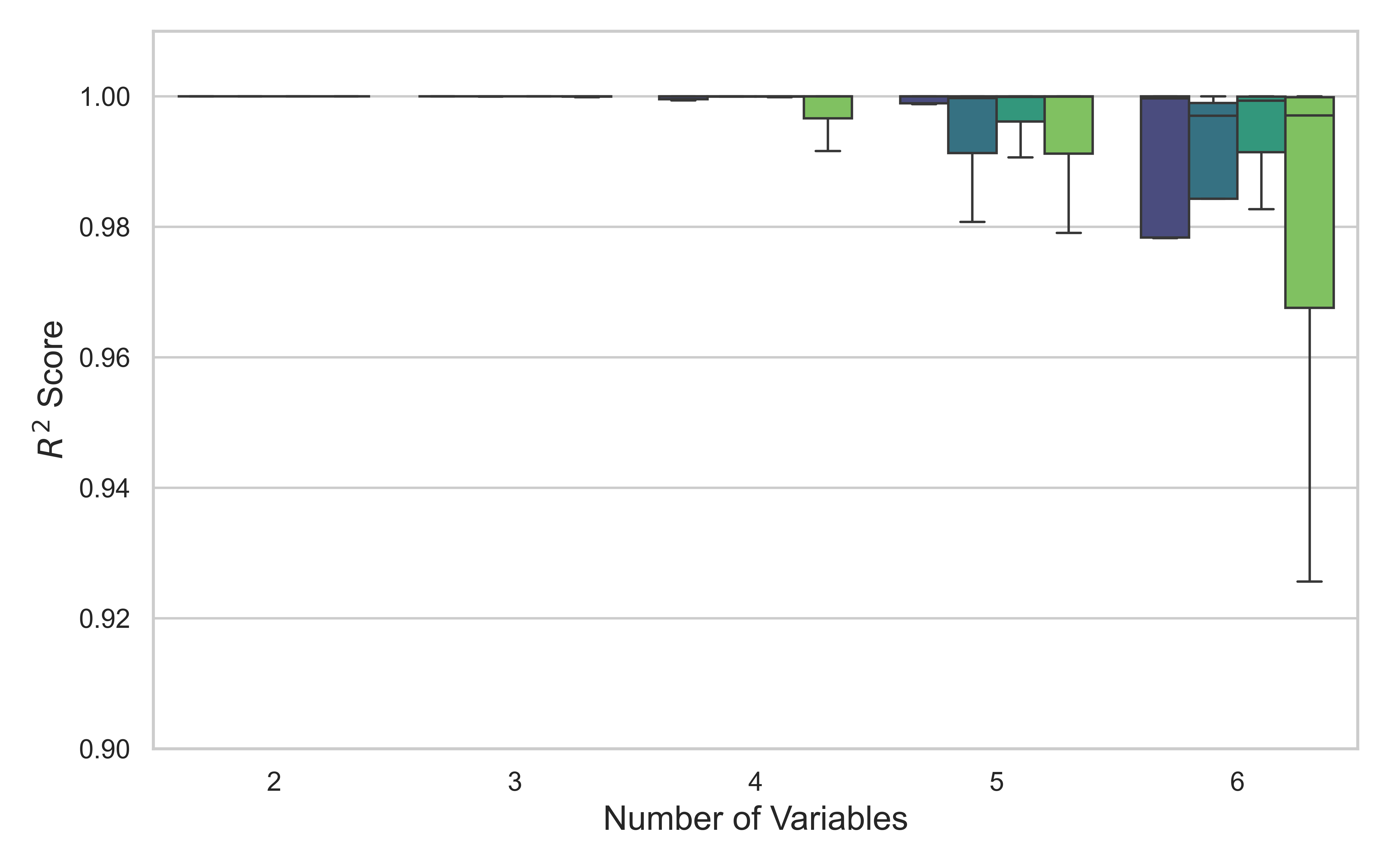}
		\caption{$R^2$ Score with number of variables}
		\label{fig:r2_vars}
	\end{subfigure}
	\hfill
	\begin{subfigure}{1\columnwidth}
		\centering
		\includegraphics[width=1\textwidth]{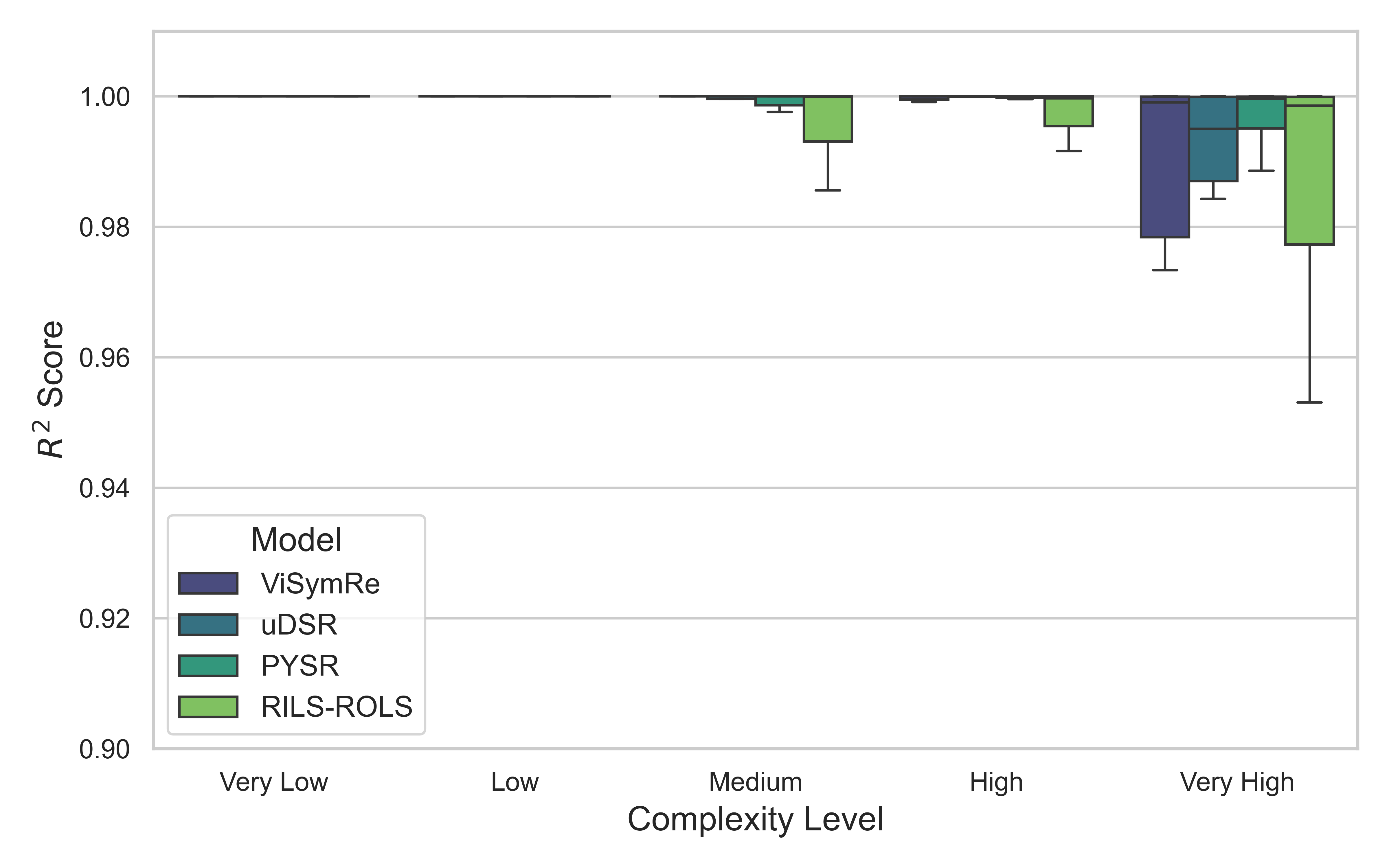}
		\caption{$R^2$ Score with Complexity level}
		\label{fig:r2_Complexity}
	\end{subfigure}
	
	\caption{\textbf{Stratified analysis results of models on the Feynman benchmark.}  Complexity levels are defined by quantiles of the ground-truth equations. In low-difficulty scenarios, the median $R^2$ scores of all models are stable, but as the difficulty increased, RILS-ROLS fluctuated first. ViSymRe consistently maintained a stable median $R^2$ score, particularly in low-difficulty scenarios where its standard deviation is near 0.
	}
	\label{fig:feynman_analysis}
\end{figure}

To evaluate how ViSymRe's performance changes with problem difficulty, we conduct a stratified analysis on the Feynman benchmark. The difficulty is distinguished by two dimensions: the number of variables and the complexity of the ground-truth equations. To ensure statistical significance, we exclude subsets containing fewer than 5 problems. Fig~\ref{fig:feynman_analysis} illustrates the distribution of $R^2$ score. As we can see, ViSymRe exhibits a significant advantage in low-complexity scenarios. The median $R^2$ score of ViSymRe is close to 1.0 with negligible variance. As the problems become more difficult, performance degradation is observed across all models. RILS-ROLS exhibits markedly increased variance. uDSR suffers a significant drop in median $R^2$ score. Notably, even in extremely difficult scenarios, ViSymRe maintains a high median $R^2$ score, with variance only marginally exceeding that of PYSR. Overall, the results indicate that ViSymRe is robust overall when handling complex problems, but is more susceptible to outliers. This phenomenon is intuitive, as processing complex, long-sequence poses challenges for pre-trained models from both training and decoding perspectives.

\subsection{Results on SRSD-Feynman benchmark}

The SRSD-Feynman benchmark is divided into three difficulty levels: Easy, Medium, and Hard, comprising 30, 40, and 50 problems, respectively. In Table~\ref{tb:SRSD-Feynman}, we detail the Accuracy Solution Rate results of the models. This benchmark presents a significant challenge to ViSymRe because the datasets it uses frequently contain extremely small or large values. However, by leveraging the proposed AMS algorithm, ViSymRe effectively mitigates this issue. As evidenced by the results, ViSymRe achieves a perfect 100\% Accuracy Solution Rate on the Easy subset and exceeds 82.5\% on the Medium subset, outperforming PYSR, uDSR, and RILS-ROLS. On the Hard subset, where the  Accuracy Solution Rate of PYSR and uDSR deteriorates to 38\% and 20.0\%, respectively, ViSymRe remains highly competitive. Overall, these results validate the effectiveness of the AMS algorithm, enabling ViSymRe to achieve capabilities comparable to GP and DL-based models in handling datasets with extreme scale.

\begin{table}[h] 
	\centering
	\caption{Accuracy Solution Rate results on SRSD-Feynman benchmark. The results for uDSR and PYSR are referenced from~\cite{matsubara2022srsd}.}
	\label{tb:SRSD-Feynman}
	\renewcommand{\arraystretch}{1.3}
	\footnotesize 
	\newcolumntype{Y}{>{\centering\arraybackslash}X} 
	\begin{tabularx}{\columnwidth}{@{} l YYYY @{}}
		\toprule
		\textbf{Datasets} & \textbf{ViSymRe} & \textbf{uDSR} & \textbf{PYSR} & \textbf{RILS-ROLS} \\
		\midrule
		Easy   & \textbf{100.0\%} & \textbf{100.0\%} & 66.7\% & 83.3\% \\
		Medium & \textbf{82.5\%}  & 75.0\%           & 45.0\% & \underline{76.9\%} \\
		Hard   & \underline{48.0\%} & 20.0\%           & 38.0\% & \textbf{50.0\%} \\
		\bottomrule
	\end{tabularx}
\end{table}

\subsection{Results on SRBench 2.0}
In this section, we evaluate ViSymRe on the black-box and phenomenological \& first-principles benchmarks published by SRBench 2.0. For datasets with dimensions exceeding 10, we employ the Top-k feature selection method recommended by SRBench 2.0 to reduce the feature space to ensure compatibility with ViSymRe.

\begin{figure}[h]
	\centering
	\begin{subfigure}{1\columnwidth}
		\centering
		\includegraphics[width=\linewidth]{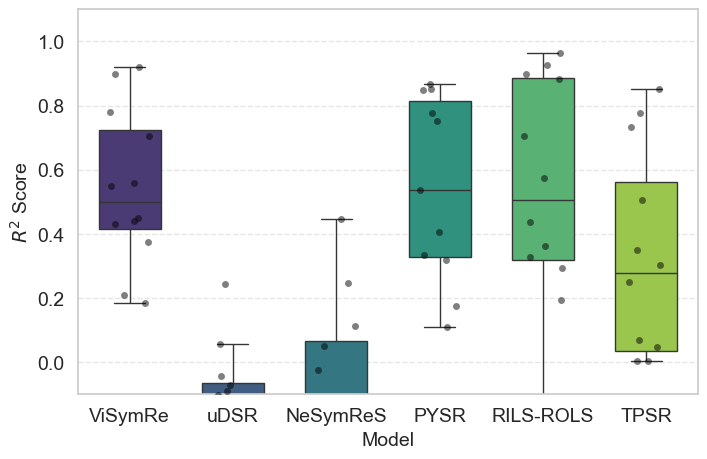}
		\caption{$R^2$ Score distribution}
		\label{fig:blackbox_r2}
	\end{subfigure}
	\hfill 
	\begin{subfigure}{1\columnwidth}
		\centering
		\includegraphics[width=\linewidth]{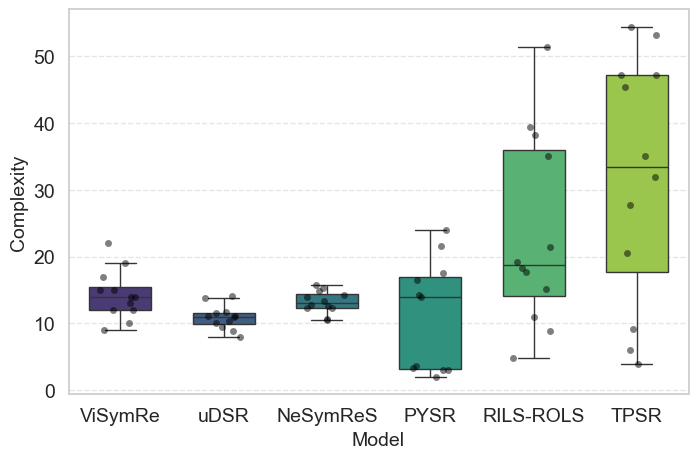}
		\caption{Complexity distribution}
		\label{fig:blackbox_Complexity}
	\end{subfigure}
	
	\caption{\textbf{The results of models on Black-box problems.}ViSymRe demonstrates a robust trade-off, achieving a competitive $R^2$ score with lower Complexity compared to baselines.}
	\label{fig:blackbox_results}
\end{figure}

Black-box benchmark represents real-world phenomena where the underlying mathematical structure is unknown. Testing on it is crucial for determining a model's availability in empirical data mining tasks where no prior physical assumptions exist. Fig.~\ref{fig:blackbox_results} presents the distribution of $R^2$ score and Complexity for all models. uDSR and NeSymRes struggle to capture the underlying laws of these datasets, resulting in the median $R^2$ score close to 0. RILS-ROLS and TPSR achieve a high $R^2$ score but at the cost of generating highly complex equations that are difficult to interpret. ViSymRe obtains a high median $R^2$ score comparable to PYSR and RILS-ROLS, while maintaining a low and stable Complexity. Although ViSymRe does not consistently achieve the highest $R^2$ score (possibly because black-box datasets lack physical priors and feature cleaning, resulting in inherently complex underlying equations), its performance exceeds that of TPSR, making it the best-performing pre-trained SR model on this benchmark. Overall, the generally low $R^2$ score across all models highlights the intrinsic difficulty of the black-box benchmark, which remains a significant challenge for the field of SR.

\begin{figure}[h]
	\centering
	
	\begin{subfigure}[b]{1\columnwidth}
		\centering
		
		\includegraphics[width=\linewidth]{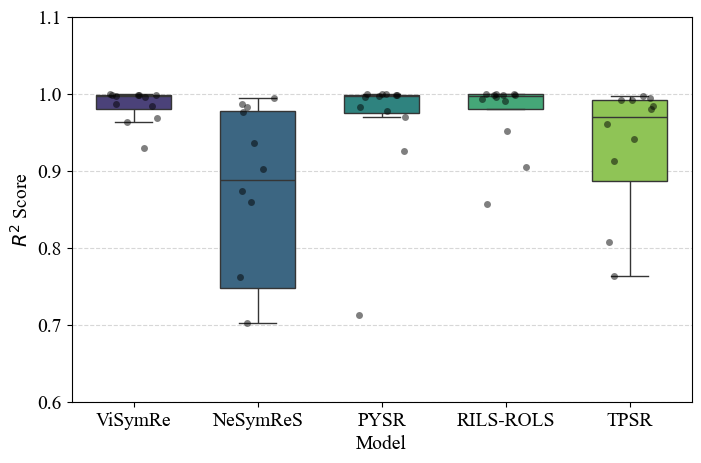}
		\caption{$R^2$ Score distribution}
		\label{fig:first_principles_r2}
	\end{subfigure}
	\hfill
	\begin{subfigure}[b]{1\columnwidth}
		\centering
		\includegraphics[width=\linewidth]{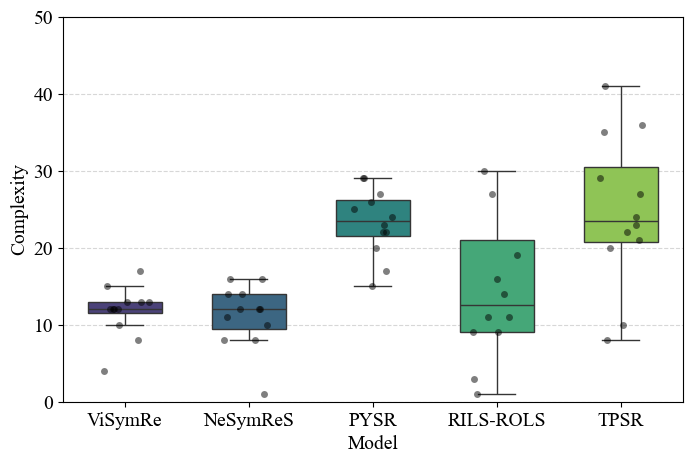}
		\caption{Complexity distribution}
		\label{fig:first_principles_Complexity}
	\end{subfigure}
	
	\caption{\textbf{The results of models on Phenomenological \& First-principles benchmark.} ViSymRe achieves $R^2$ score comparable to leading baselines, while outperforming them in terms of Complexity. }
	\label{fig:first_principles_result}
\end{figure}

To assess the ViSymRe's performance in realistic scientific discovery scenarios, we evaluate it on the Phenomenological \& First-principles problems. As illustrated in Fig.~\ref{fig:first_principles_result}, ViSymRe demonstrates a superior trade-off between $R^2$ score and Complexity. ViSymRe achieves a median $R^2$ score close to $1.0$ with a remarkably tight distribution. Importantly, the advantage of ViSymRe lies in its interpretability, as reflected in the Complexity. While PYSR and RILS-ROLS achieve a high $R^2$ score, they exhibit high variance in Complexity, often yielding complex equations. Furthermore, ViSymRe produces fewer outliers in both $R^2$ score and Complexity compared to baselines, further corroborating its robustness when handling real-world noisy datasets. 

%\begin{table}[htbp]
%	\centering
%			\caption{The details of the equations generated by ViSymRe for Phenomenological \& First-principles problems.}
%	\label{tb:First-principles_equations}
%	\renewcommand{\arraystretch}{1.3}
%	\begin{tabularx}{\linewidth}{l| c c}
%		\toprule
%		\textbf{Dataset} & \textbf{equation} & \textbf{$R^2$} \\
%		\midrule
%		bode & $0.15 e^{0.68 x_{1}} + 0.39$ & 0.997 \\
%		hubble & $480 x_{1} + 169 \sin{\left(10 x_{1} + 9.11 \right)} + 25$ & 0.925 \\
%		kepler & $362 x_{1}^{1.5}$ & 0.999 \\
%		tully\_fisher & $\begin{aligned} & \left(\sin{\left(6.23 \cdot 10^{-8} x_{1}^{3.6} \right)} + 3.57\right) \\ & \cdot \cos{\left(0.1 \sqrt{x_{1}} \right)} - 18.4 \end{aligned}$ & 0.986 \\
%		planck & $\frac{- 1.06 \cdot 10^{-12} x_{1} - 47.9}{\left(0.011 x_{2} + 1\right)^{1.3}} - 23$ & 0.996 \\
%		ideal\_gas & $\log{\left(\frac{10.5 x_{1} x_{2}}{x_{3}} - 6.93 \right)}$ & 0.998 \\
%		leavitt & $- 0.43 x_{1} + 11.8 + 3.14 e^{- 0.32 x_{1}^{3}}$ & 0.980 \\
%		newton & $\frac{0.89 x_{3}^{0.016} \log{\left(x_{2} \right)}}{x_{1}^{0.031}} - 17$ & 0.990 \\
%		rydberg & $\frac{3.55 x_{1} - 3.73}{x_{2}^{0.69}} - 15.9$ & 0.993 \\
%		schechter & $\frac{831}{x_{1}^{0.0015}} - 947 + 114 e^{- 3.75 \cdot 10^{-11} x_{1}}$ & 0.996 \\
%		absorption & $- 0.0045 x_{1} + 3.57 - 3.43 e^{- 0.032 x_{1}^{1.41}}$ & 0.995 \\
%		supernovae\_zr & $0.0064 \pi^{4.44} e^{- 0.084 \left|{x_{1}}\right|^{0.82}}$ & 0.967 \\
%		supernovae\_zg & $0.94 - 0.93 e^{- \frac{89.4}{x_{1}^{2}}}$ & 0.981 \\
%		\bottomrule
%	\end{tabularx}
%\end{table}

\subsection{Ablation analysis}
\begin{figure}[h]
	\centering
	\begin{subfigure}[b]{0.48\columnwidth}
		\centering
		\includegraphics[width=\linewidth]{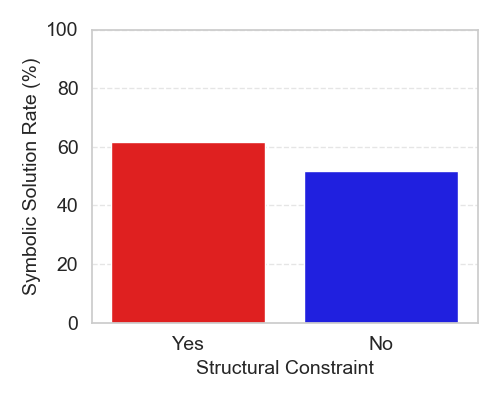}
		\caption{Symbolic Solution Rate}
		\label{fig:constraint algorithm_ssr}
	\end{subfigure}
	\hfill
	\begin{subfigure}[b]{0.48\columnwidth}
		\centering
		\includegraphics[width=\linewidth]{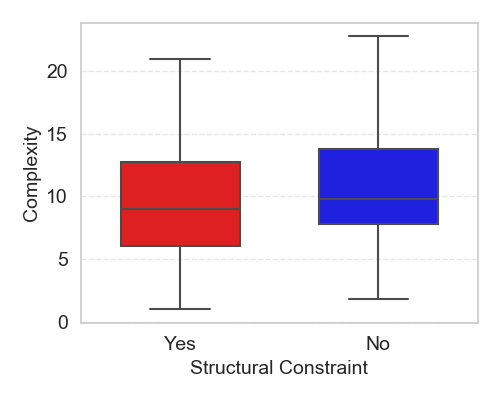}
		\caption{Complexity distribution}
		\label{fig:constraint algorithm_complexity}
	\end{subfigure}
		\vspace{0.5em} 
	\vspace{1em}
	\begin{subfigure}[b]{0.48\columnwidth}
		\centering
		\includegraphics[width=\linewidth]{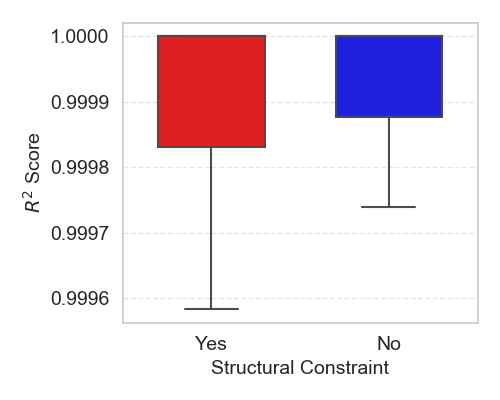}
		\caption{$R^2$ Score distribution}
		\label{fig:constraint algorithm_r2}
	\end{subfigure}
	\hfill
	\begin{subfigure}[b]{0.48\columnwidth}
		\centering
		\includegraphics[width=\linewidth]{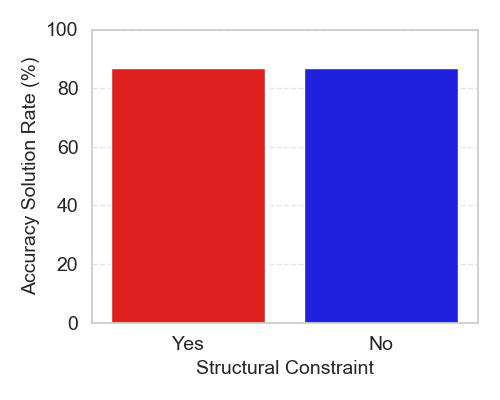}
		\caption{Accuracy Solution Rate}
		\label{fig:constraint algorithm_asr}
	\end{subfigure}
	
	\caption{\textbf{Ablation study results of the syntax constraint algorithm on low-dimensional benchmarks. } The syntax constraint algorithm is beneficial for improving the Symbolic Solution Rate and reducing Complexity. However, this comes at the cost of decreased numerical precision.}
	\label{fig:constraint algorithm ablation}
\end{figure}

As illustrated in Fig.~\ref{fig:constraint algorithm ablation}, we conduct an ablation study of the proposed syntax constraint algorithm, aiming to verify its contribution to promoting ViSymRe to generate simple and valid equations. Figs.~\ref{fig:constraint algorithm_ssr} and~\ref{fig:constraint algorithm_complexity} show that the constrained version achieves a superior Symbolic Solution Rate and lower Complexity compared to the unconstrained version, indicating that the syntax constraint algorithm effectively prunes the search space, guiding the model to converge toward ground-truth equations rather than merely numerical approximations. However, reducing Complexity requires a trade-off at the numerical level. As shown in Figs.~\ref{fig:constraint algorithm_r2} and \ref{fig:constraint algorithm_asr}, while the mean $R^2$ score distributions are similar, the Accuracy Solution Rate of the constrained version is slightly lower than that of the unconstrained version. This marginal numerical difference validates our hypothesis: the unconstrained version tends to exploit its higher degree of freedom to generate complex equations to optimize numerical approximation. In contrast, the constrained version, by restricting structurally invalid combinations, sacrifices negligible numerical precision to avoid overfitting traps.

\begin{figure}[h]
	\centering
	
	\begin{subfigure}[b]{0.48\linewidth}
		\centering
		\includegraphics[width=\textwidth]{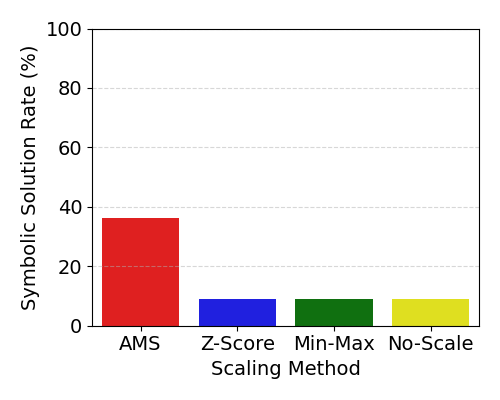}
		\caption{Symbolic Solution Rate}
		\label{fig:scale_ssr}
	\end{subfigure}
	\hfill
	\begin{subfigure}[b]{0.48\linewidth}
		\centering
		\includegraphics[width=\textwidth]{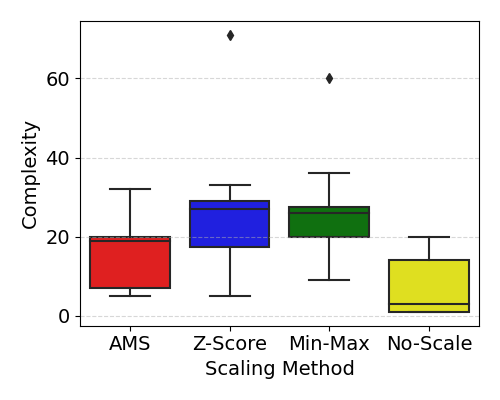}
		\caption{Complexity distribution}
		\label{fig:scale_complexity}
	\end{subfigure}
	\vspace{0.5em} 
	\begin{subfigure}[b]{0.48\linewidth}
		\centering
		\includegraphics[width=\textwidth]{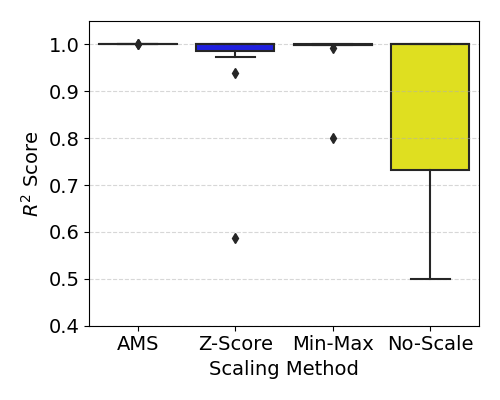}
		\caption{$R^2$ Score distribution}
		\label{fig:scale_r2}
	\end{subfigure}
	\hfill
	\begin{subfigure}[b]{0.48\linewidth}
		\centering
		\includegraphics[width=\textwidth]{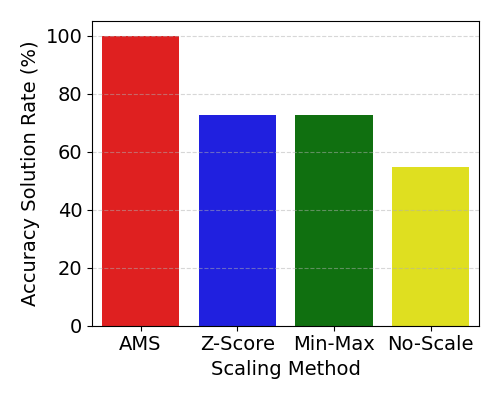}
		\caption{Accuracy Solution Rate}
		\label{fig:scale_asr}
	\end{subfigure}
	
	\caption{\textbf{Ablation study results of scaling strategies on low-dimensional benchmarks.} Traditional normalization methods fail to facilitate ViSymRe to recover the ground-truth symbolic structures, despite their contribution to improving $R^2$ scores. }
	\label{fig:scaling_comparison}
\end{figure}

To evaluate the effectiveness of the proposed AMS algorithm in adapting ViSymRe to datasets with varying scales, we randomly select several equations from low-dimensional benchmarks and construct multi-scale test datasets by resampling coefficients and variable values (see Table~\ref{tab:Supplementary_scale_ablation} for details). We compare AMS with standard normalization techniques (Z-Score and Min-Max) and a baseline without scaling. As illustrated in Fig.~\ref{fig:scaling_comparison}, the AMS version demonstrates superior performance across all metrics. AMS is the only version capable of reliably recovering the ground-truth symbolic structure. In terms of Complexity, we can see that the AMS version generates equations with significantly lower Complexity compared to Z-Score and Min-Max versions, avoiding the overcomplexity problem caused by traditional scale scaling methods when recovering the symbolic structure of equations. Furthermore, the $R^2$ distribution and the Accuracy Solution Rate confirm that AMS version achieves robust numerical approximation results, achieving a 100\% Accuracy Solution Rate. 

Subsequently, we perform ablation analysis of the key factors that potentially affect the performance of ViSymRe, including the visual modality, Codebook size, and feature fusion module. We train all variants for 20 epochs on a training set comprising 100,000 equation skeletons, each with a maximum of three variables. To facilitate model convergence, the sampling range for all variables is restricted to $\mathcal{U}(-5, 5)$. We employ a controlled variable approach to evaluate the impact of each ablation factor. The configuration of the Base model is provided in Table~\ref{tb:ablation_parameters}.

\begin{table}[h]
	\centering
	\caption{Default parameters are used for ablation analysis.}
	\label{tb:ablation_parameters}
	\renewcommand{\arraystretch}{1.3}
	\begin{tabularx}{\linewidth}{l| >{\centering\arraybackslash}X}
		\hline
		\textbf{Name} & \textbf{Default value} \\ \hline
		Paradigm & Multimodal \\
		Codebook size & 512 \\
		Feature fusion module & Biased Cross-Attention \\ \hline
	\end{tabularx}
\end{table}

\begin{figure}[h]
	\centering
	\includegraphics[width=1\linewidth]{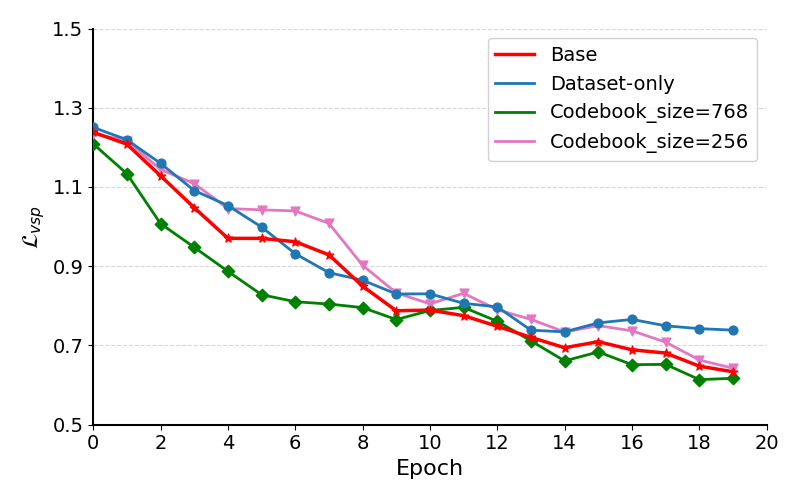}
	\caption{\textbf{Skeleton prediction loss of various ablation variant.} The dataset-only variant exhibits limited convergence and plateaus prematurely due to the absence of the visual modality. Furthermore, the codebook size exerts minimal influence on the ultimate convergence level.
	}\label{fig:skeleton_prediction_loss}
\end{figure}

Fig.~\ref{fig:skeleton_prediction_loss} illustrates the convergence results of the skeleton prediction loss ($\mathcal{L}_{vsp}$) for variants on the validation set. It can be observed that the loss of the dataset-only variant is significantly higher than that of the others, and it stops converging as early as the 14th epoch. Furthermore, although the large Codebook exhibits a faster loss decrease in the early stages of training, its final convergence level is on par with the Base model, whereas the small Codebook, due to limited capacity, results in insufficient visual representation capability, leading to higher loss. However, overall, the Codebook size has a limited impact on the final convergence level of the model.

In addition, it is observed that, except for the dataset-only variant, the loss curves for all variants and the Base model exhibit a plateau lasting 3–4 epochs. For example, this phenomenon occurs during epochs 4–6 for the Base model and the variant with $Codebook\_size=256$, and during epochs 9–12 for the variant with $Codebook\_size=768$. The possible reason for this stagnation is attributable to the instability of the Codebook during the early stages of training.

\begin{figure}[h]
	\centering
	\includegraphics[width=1\linewidth]{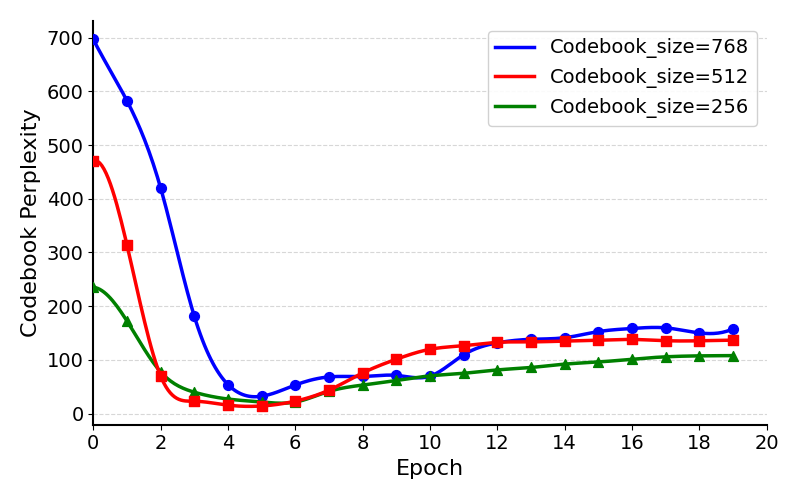}
	\caption{\textbf{Comparison of perplexity under different Codebook sizes.} Regardless of the initial Codebook size, the perplexity eventually converges to a similar level, demonstrating that the model is insensitive to the Codebook size.
	}\label{fig:Codebook_perplexity}
\end{figure}

To quantify the effective capacity of the Codebook, we investigate its utilization using perplexity, which is defined as the exponent of the entropy of the Codebook distribution:
\begin{equation}
	perplexity = \exp ( - \sum_{s=1}^{S} p_s \log p_s ),
\end{equation}
where $p_s$ denotes the empirical usage frequency of the $s$-th codeword. Fig.~\ref{fig:Codebook_perplexity} records the perplexity results for different Codebook sizes. It is observed that all curves exhibit a U-shaped pattern. In the initial training stage, the model shows a trend of concentrated codeword usage, leading to a brief decrease in perplexity. As training progresses, the model learns to utilize a wider range of codewords, causing the perplexity to rebound and stabilize. Notably, a larger Codebook ultimately yields higher perplexity, suggesting richer feature representations. However, comparing this conclusion with the validation loss in Fig.~\ref{fig:skeleton_prediction_loss} reveals a discrepancy: such high Codebook utilization does not lead to higher SR performance. This suggests the presence of redundant codewords, which may introduce ambiguity during downstream multimodal fusion and decoding. Furthermore, the results show that the model eventually converges to a reasonable perplexity regardless of the initial Codebook size. Therefore, Codebook size is not a sensitive parameter. In large-scale pre-training, a large Codebook is still necessary to avoid insufficient representation capabilities.

\begin{figure}[h]
	\centering
	\includegraphics[width=1\linewidth]{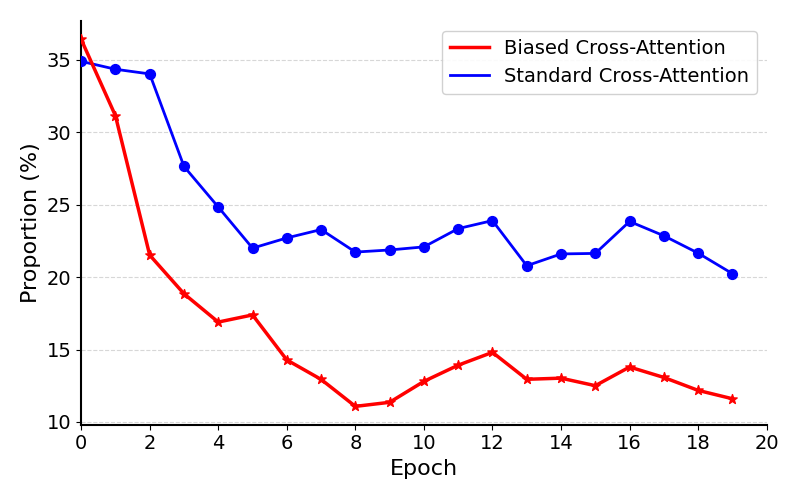}
	\caption{\textbf{The proportion of noise attention weights in standard Cross-Attention and Biased Cross-Attention feature fusion modules. The Biased Cross-Attention module significantly suppresses the proportion of noise attention weights through the bias term, achieving a reduction of over 10\% compared to the standard paradigm.}
	}\label{fig:noise_proportion}
\end{figure}

Fig.~\ref{fig:noise_proportion} further evaluates the contribution of the bias term within the Biased Cross-Attention feature fusion module to the suppression of noise. The results demonstrate that in the standard Cross-Attention module, the proportion of noise attention weights fluctuates between 20\% and 30\% throughout the training process, exhibiting only a marginal downward trend. This implies that non-negligible noise interference persists. In the Biased Cross-Attention module, the initially proportion of noise attention weights is high, which is likely attributable to bias perturbation resulting from unconverged alignment loss $\mathcal{L}_{align}$(Eq.~\ref{eq:align}).  As the number of epochs increases,   the proportion of noise attention weights demonstrates a significant decline. These findings corroborate our hypothesis that the bias term effectively guides the attention mechanism to progressively discard irrelevant noise features, which is critical for enhancing the effectiveness of multimodal learning.

\begin{table}[h]
	\centering
	\caption{The mean median $R^2$ score of major ablation variants on low-dimensional benchmarks. Std-CA is short for standard Cross-Attention.}
	\label{tb:ablation_r2}

	\footnotesize 
	\renewcommand{\arraystretch}{1.3} 
	\setlength{\tabcolsep}{1pt} 
	
	\newcolumntype{Y}{>{\centering\arraybackslash}X}
	
	\begin{tabularx}{\columnwidth}{l *{3}{Y}}
		\toprule
		\textbf{Benchmarks} & \textbf{Base} & \textbf{Std-CA} & \textbf{Dataset-only} \\
		\midrule
		Constant & \textbf{0.9872} $\pm$ {\scriptsize 0.0010} & \underline{0.9772} $\pm$ {\scriptsize 0.0109} & 0.9707 $\pm$ {\scriptsize 0.0097} \\
		\addlinespace[0.5ex]
		Jin & \textbf{0.9543} $\pm$ {\scriptsize 0.0174} & \underline{0.8827} $\pm$ {\scriptsize 0.0334} & 0.7861 $\pm$ {\scriptsize 0.0652} \\
		\addlinespace[0.5ex]
		Keijzer & \textbf{0.9907} $\pm$ {\scriptsize 0.0007} & \underline{0.9415} $\pm$ {\scriptsize 0.0063} & 0.8561 $\pm$ {\scriptsize 0.0942} \\
		\addlinespace[0.5ex]
		Korns & \textbf{1.0000} $\pm$ {\scriptsize 0.0000} & \underline{0.9930} $\pm$ {\scriptsize 0.0088} & 0.9743 $\pm$ {\scriptsize 0.0104} \\
		\addlinespace[0.5ex]
		Livermore & \textbf{0.9978} $\pm$ {\scriptsize 0.0009} & \underline{0.9832} $\pm$ {\scriptsize 0.0059} & 0.9693 $\pm$ {\scriptsize 0.0064} \\
		\addlinespace[0.5ex]
		Neat & \textbf{0.9740} $\pm$ {\scriptsize 0.0064} & \underline{0.9529} $\pm$ {\scriptsize 0.0099} & 0.9436 $\pm$ {\scriptsize 0.0341} \\
		\addlinespace[0.5ex]
		Nguyen & \textbf{0.9974} $\pm$ {\scriptsize 0.0018} & \underline{0.9875} $\pm$ {\scriptsize 0.0017} & 0.9706 $\pm$ {\scriptsize 0.0058} \\
		\addlinespace[0.5ex]
		Nguyen$'$ & \textbf{0.9988} $\pm$ {\scriptsize 0.0005} & \underline{0.9907} $\pm$ {\scriptsize 0.0006} & 0.9752 $\pm$ {\scriptsize 0.0224} \\
		\addlinespace[0.5ex]
		Nguyen$^c$ & \textbf{0.9900} $\pm$ {\scriptsize 0.0066} & \underline{0.9859} $\pm$ {\scriptsize 0.0038} & 0.9764 $\pm$ {\scriptsize 0.0192} \\
		\midrule
		\textbf{Average} & \textbf{0.9878} $\pm$ {\scriptsize 0.0040} & \underline{0.9661} $\pm$ {\scriptsize 0.0091} & 0.9358 $\pm$ {\scriptsize 0.0308} \\
		\bottomrule
	\end{tabularx}
\end{table}

Table~\ref{tb:ablation_r2} presents the SR performance evaluation results of the Base model and main ablation variants on low-dimensional benchmarks. The results are the average of the median $R^2$ scores from 10 independent runs. It can be observed that the performance ranking of the variants is generally consistent with the convergence levels reported in Fig.~\ref{fig:skeleton_prediction_loss}. The Base model achieves the highest $R^2$ score and exhibits the most stable performance. In contrast, the dataset-only variant exhibits the most pronounced degradation, with an average $R^2$ score dropping by more than 0.05 compared to the Base model, alongside increased fluctuations. Furthermore, the variant utilizing standard Cross-Attention yields a suboptimal solution, suggesting that the inadequate suppression of noise in the virtual vision, thus compromising the final SR performance.

\section{Conclusions}
\label{conclusions}

In this paper, we propose ViSymRe, a pre-trained Transformer-based framework, exploring how the visual modality can enhance the SR performance. We propose MVRS to achieve non-degenerate visualization of multivariate equations with minimal sampling complexity, thereby enabling ViSymRe to be large-scale pre-trained. By employing the dual-visual  pipeline and the Biased Cross-Attention feature fusion module, we incorporate visual priors into the inference process without requiring explicit visual input, ensuring that ViSymRe works in the dataset-only paradigm during deployment. Experiments demonstrate that ViSymRe is competitive in balancing various SR metrics, including $R^2$ score, Symbolic Solution Rate, efficiency, and Complexity. In particular, ViSymRe excels at handling low-complexity and rapid-inference scenarios, such as low-dimensional benchmarks, Feynman problems with dimensions less than 6, and phenomenological \& first-principles problems.

\textbf{Limitations:}  As with some pre-trained models, the generalization capability of ViSymRe on out-of-distribution domain remains constrained. Although the AMS algorithm has been proven to be able to mitigate this problem, ViSymRe may still encounter difficulties when test datasets differ significantly from the training distribution. Moreover, while MVRS enhances the richness of visual features through multiple views, the need for analytical equations forces ViSymRe to rely on predicting visual features to ensure dataset-only inference. Essentially, MVRS is a trade-off under resource constraints, ensuring the feasibility of large-scale pre-training through limited sampling complexity.

\textbf{Potential contributions to broader fields:} In addition to the field of SR, the proposed architecture also has potential in a wider range of multimodal learning scenarios. Firstly, the Biased Cross-Attention mechanism offers a feasible solution for robust multimodal fusion in scenarios affected by noisy modalities (such as audio-visual recognition under noise~\cite{zhu2021deep}). By dynamically suppressing task-irrelevant features via the bias term, it ensures stability even when one modality is unreliable. Secondly, the dual-visual pipeline architecture demonstrates how to effectively transform knowledge from data-rich training environments into lightweight, resource-constrained inference environments, such as multi-view sensors in robotics or multimodal medical records~\cite{xu2023multimodal,huang2020fusion,lin2024ai}. This module effectively bridges the gap in modality availability between training and deployment, providing a reference for efficient multimodal inference.

\section*{Acknowledgments}
This work was partially supported by Major Program of National Key R\&D Program of China (2023YFA1009002), National Natural Science Foundation of China (Nos.12292980, 12292984, 12031016, 12201024), National Key R\&D Program of China (Nos.2023YFA1009000, 2023YFA1009004, 2020YFA0712203, 2020YFA0712201) and Beijing Natural Science Foundation (BNSF-Z210003).

%\section*{Author contributions}
%\textbf{Da Li:} Conceptualization, Methodology, Software, Investigation, Writing - Original Draft \textbf{Junping Yin:} Supervision, Project administration, Resources \textbf{Jin Xu:} Investigation, Data Curation, Resources \textbf{Xinxin Li:} Validation, Visualization \textbf{Juan Zhang:} Conceptualization, Methodology, Writing - Review \& Editing, Supervision.
\bibliographystyle{elsarticle-num}
\bibliography{rel}

\clearpage
\onecolumn 
\appendix

\section{IEEE-754}
\label{IEEE-754}

\begin{figure}[htbp]
	\centering
	\includegraphics[width=0.6\linewidth]{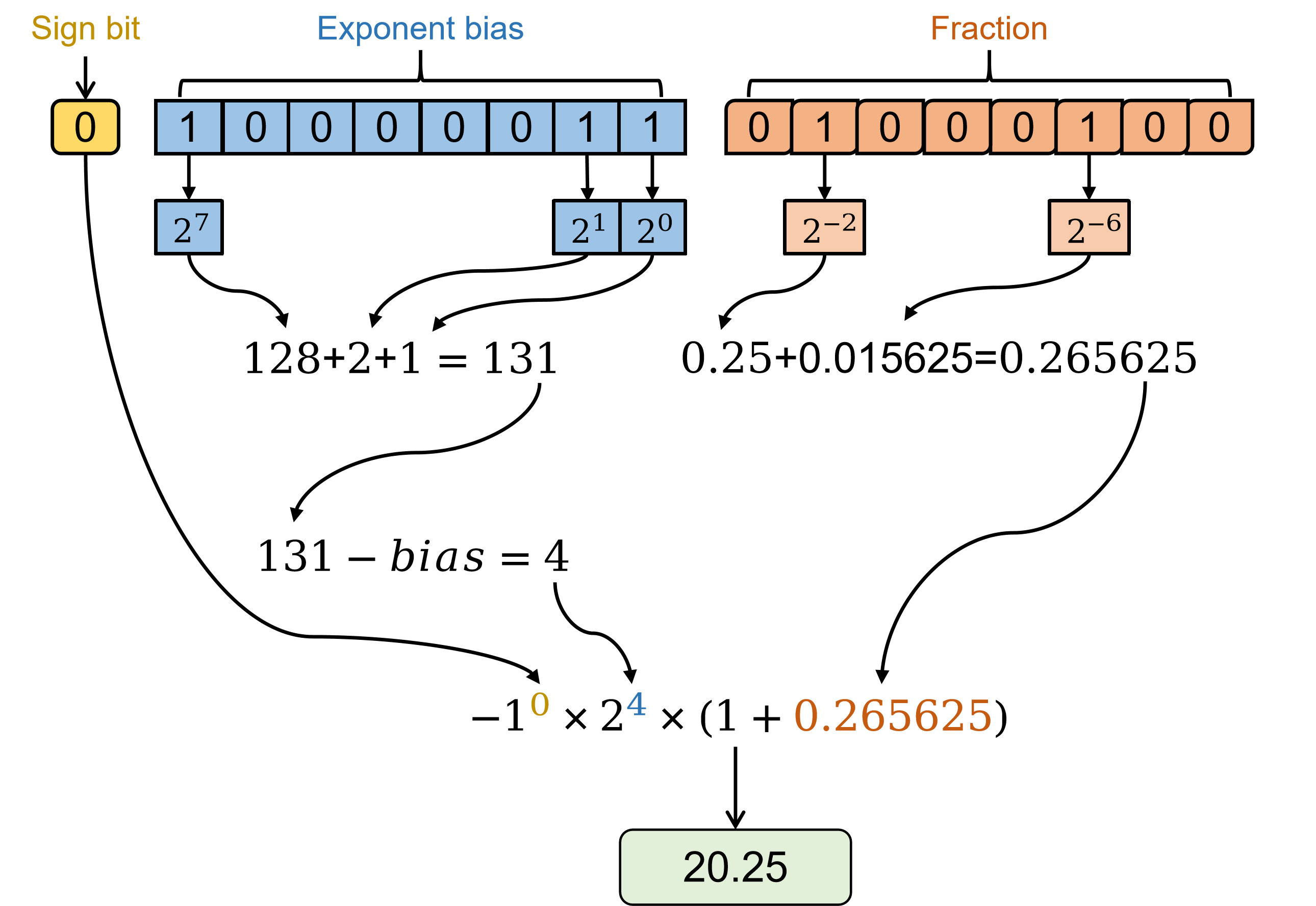}
	\caption{\textbf{Illustration of the mapping between the floating-point number 20.25 and its corresponding IEEE-754 format.} }
	\label{ieee754}
\end{figure}

The IEEE-754 standard is a binary representation method used in computer science. It represents floating-point numbers in the following form:
\[
(-1)^S \times 1.F \times 2^{(E - \text{bias})}
\]

where $S$ is the sign bit (0 for positive, 1 for negative), $E$ is the biased exponent, $F$ is the fractional part, and 
$\text{bias}$ is a fixed offset added to the actual exponent to represent both positive and negative exponents ($\text{bias}=127$ for single precision). For example, in the 32-bit single-precision format, the structure consists of 1 bit for the sign, 8 bits for the exponent, and 23 bits for the fraction. In certain applications, reduced-precision formats (e.g., with an 8-bit fraction) are used for better performance.
Fig.~\ref{ieee754} presents an example showing the relationship between 20.25 and its IEEE-754 representation. We use the decoding process to illustrate this correspondence because it is clearer.  

\section{Main parameters for ViSymRe}
The main parameters of ViSymRe are shown in Table~\ref{tb:parameters}. Except for the ablation experiments, all experimental results are obtained under this configuration.

\begin{table}[h]
	\centering
	\renewcommand{\arraystretch}{1.3} 
	\caption{The default parameters employed of ViSymRe.}
	\label{tb:parameters}
	\begin{tabularx}{\textwidth}{Xcc} 
		\hline
		\textbf{Description} & \textbf{Symbol} & \textbf{Value} \\ \hline
		\textbf{Architecture} & & \\ \hline
		Number of Dataset Encoder layers & $\sim$ & 4 \\
		Number of Visual Decoder layers & $\sim$ & 4 \\
		Number of (standard and Biased) Cross-Attention layers & $\sim$ & 1 \\
		Number of Decoder layers & $\sim$ & 8 \\
		Model dimension & $\sim$ & 512 \\
		Number of views generated by MVRS & $K$ & 3 \\
		Sequence length visual features & $M$ & 16 \\
		Size of the Codebook & $S$ & 4096 \\
		
		\textbf{Pre-training} & & \\ \hline
		Training set size &$\sim$ &50,000,000\\
		Number of sampled data points & $N$ & 200 \\
		Batch size & $B_n$ & 500 \\
		Epochs & $\sim$ & 50 \\
		$\beta_1$ and $\beta_2$ in Eq.~\ref{loss_q} & $\beta_1, \beta_2$ & 0.25, 0.001 \\
		$\lambda$ in Eq.~\ref{eq:loss_total} & $\lambda$ & 0.1 \\
		
		 Truncation threshold for visualization & $\Omega$ & 1e6\\ \hline
		
		\textbf{Testing} & & \\ \hline
		Beam size & $\sim$ & 3--30 \\
		Number of bags & $N_{bag}$ & 10 \\
		Number of test points per bag & $\sim$ & 200 \\
		Initial sampling range of constant for BFGS & $\sim$ & $\mathcal{U}(0,10)$ \\
		Number of BFGS restarts & $\sim$ & 10 \\ 
		Error threshold &$\epsilon$ &1e-5\\
		\hline
	\end{tabularx}
\end{table}

\section{Main parameters for baselines}
\label{parameters for baselines}
The parameters for the baselines compared in the experiments, including PYSR, uDSR, RILS-ROLS, MLP, Random Fores and SVM, are presented in Table~\ref{tab:PYSR_UDSR_RILS_ROLS_ML_parameters}.

\begin{table}[h]
	\centering
	\caption{The parameters of PYSR, uDSR, RILS-ROLS, MLP, Random Fores and SVM. }
	
	\renewcommand{\arraystretch}{1.3}
	\label{tab:PYSR_UDSR_RILS_ROLS_ML_parameters}
	\begin{tabular}{l|p{14cm}}
		\hline
		\textbf{Baselines} & \textbf{Hyperparameter sets} \\
		\hline
		uDSR & 
		$\{ \textit{function\_set}: [\text{'add'}, \text{'sub'}, \text{'mul'}, \text{'div'}, \text{'sin'}, \text{'cos'}, \text{'exp'}, \text{'log'}, \text{'poly'}] $ \newline
		$\textit{batch\_size}: 500, \textit{n\_samples}: 200000 \}$  \\ \hline
		
		PYSR & 
		$\textit{procs}: 5, \textit{populations}: 10, \textit{population\_size}: 40, \textit{ncyclesperiteration}: 500, $ \newline
		$\textit{niterations}: 10000, \textit{timeout\_in\_seconds}: 5000, \textit{maxsize}: 50, $ \newline
		$\textit{binary\_operators}: [\text{'*'}, \text{'+'}, \text{'-'}, \text{'/'}], \textit{unary\_operators}: [\text{'sin'}, \text{'cos'}, \text{'exp'}, \text{'log'}] $ \newline
		$\textit{nested\_constraints}: \{\text{sin}: \{\text{sin}: 0, \text{cos}: 0\}, \text{cos}: \{\text{sin}: 0, \text{cos}: 0\}, $ \newline
		$\text{exp}: \{\text{exp}: 0\}, \text{log}: \{\text{log}: 0\}\}, \textit{progress}: \text{False}, \textit{weight\_randomize}: 0.1, $ \newline
		$\textit{precision}: 32, \textit{warm\_start}: \text{True}, \textit{turbo}: \text{True}, \textit{update}: \text{False}$  \\ \hline
		
		RILS-ROLS & $\textit{max\_fit\_calls}: 100000, \textit{max\_time}: 5000, \textit{Complexity\_penalty}: 0.001, \textit{verbose}: \text{False} $ \\ \hline
		
		MLP & $\textit{hidden\_layer\_sizes}: (100, 100), \textit{activation}: \text{'relu'}, \textit{solver}: \text{'adam'}, $ \newline
		$\textit{max\_iter}: 2000, \textit{scaling}: \text{Z-Score}$ \\ \hline
		
		Random Forest & $\textit{n\_estimators}: 100, \textit{scaling}: \text{None}$ \\ \hline
		
		SVM & $\textit{kernel}: \text{'rbf'}, \textit{C}: 10.0, \textit{epsilon}: 0.1, \textit{scaling}: \text{Z-Score}$ \\ \hline
		
	\end{tabular}
\end{table}

\section{Main parameters for skeleton generator}
\label{equation_generator}
Table~\ref{tab:equation_generator} details the parameters employed by the equation skeleton generator. To maintain the complexity of the equation skeletons, we impose strict upper bounds on the number of nodes, constants, and operators within each skeleton. Furthermore, different sampling probabilities are assigned to different unary and binary operators. Note that although the distribution of exponents and constants is pre-defined, they produce a wider distribution after internal operations, such as $({x^3})^3$ $\to$ $x^9$. 
\begin{table}[h]
	\centering
	\renewcommand{\arraystretch}{1.3}
	\caption{The parameters employed by equation generator.}
	\label{tab:equation_generator}
	\begin{tabularx}{\linewidth}{c|>{\centering\arraybackslash}X}
		\hline
		\textbf{Description} & \textbf{Value} \\ \hline
		Max dimension of input variables       & 10         \\
		Max number of unary operators       & $5$            \\ 
		Max number of binary operators &$Number(variables)+5$               \\ 
		Max number of constants       & 3               \\ 
		Max number of nodes           & 50               \\ 
		Distribute of constants    & $\mathcal{U}(-10,10)$ \\ 
		Sampling frequency of binary operators  & add:1, sub:0.5, mul:1, div:0.5  \\ 
		Sampling frequency of unary operators  & abs:0.1, pow2:1, pow3:1, pow5:0.1, sqrt:1,  sin:0.5, cos:0.5, tan:0.1, arcsin:0.1, log:0.5, exp:0.5 \\ \hline
	\end{tabularx}
	
\end{table}

\section{Supplementary analysis of AMS}
\label{analysis_AMS}
In SR, normalization plays a different role compared to traditional machine learning tasks. While standard normalization methods (such as Z-score, Min-Max) aim to accelerate gradient descent, the primary objective in SR is to recover the ground-truth symbolic structure of equations. Furthermore, pre-trained SR models are typically trained on bounded data distributions. To enable these models to extrapolate to datasets with distributions of varying magnitudes, normalization must bridge the scale gap without corrupting the underlying symbolic form.

Z-score normalization applies an affine transformation to both inputs and targets:
$$x' = \frac{x - \mu_x}{\sigma_x}, \quad y' = \frac{y - \mu_y}{\sigma_y}.$$
This implies that the physical variables are represented as $x = \sigma_x x' + \mu_x$ and $y = \sigma_y y' + \mu_y$. Consider a equation $y = x^2$. In the normalized space, the relationship becomes:
$$\sigma_y y' + \mu_y = (\sigma_x x' + \mu_x)^2.$$
Solving for the target $y'$, the model is forced to fit:
$$y' = \frac{\sigma_x^2}{\sigma_y}(x')^2 + \frac{2\sigma_x\mu_x}{\sigma_y}x' + \frac{\mu_x^2 - \mu_y}{\sigma_y}.$$
Consequently, a simple structure $x^2$ is distorted into a quadratic polynomial involving additive bias terms and lower-order interaction terms. The SR model is compelled to predict additional subtraction operators and constant to cancel out the artificial shift $\mu_x$, significantly increasing the complexity of the search space. 

In contrast, AMS applies a linear transformation defined by dynamic magnitude estimation:
$$x' = \frac{x}{s_x}, \quad y' = \frac{y}{s_y},$$
where $s$ represents the characteristic scale. Substituting $x = s_x x'$ and $y = s_y y'$ into the same law $y = x^2$ yields:
$$s_y y' = (s_x x')^2 \implies y' = \frac{s_x^2}{s_y}(x')^2.$$
Under AMS, the target equation retains the identical symbolic structure $y' = C (x')^2$. The scaling factors are absorbed into a coefficient $C$. Since pre-trained SR models typically decouple skeleton generation from constant optimization, this coefficient can be effortlessly resolved by numerical algorithms.

\section{Supplementary results on Feynman benchmark}
\label{Supplementary results on Feynman benchmark}

{
	\renewcommand{\arraystretch}{2.0}
	\begin{longtable}{| >{\centering\arraybackslash}m{0.08\linewidth} | >{\centering\arraybackslash}m{0.31\linewidth} | >{\centering\arraybackslash}m{0.31\linewidth} | >{\centering\arraybackslash}m{0.09\linewidth} | >{\centering\arraybackslash}m{0.06\linewidth} |}
		\caption{Details results on Feynman benchmark. All physical variables are mapped into a format that the ViSymRe can recognize. The SSR stands for Symbolic Solution Rate.}
		 \label{tab:Feynman} \\
		\hline
		\textbf{Name} & \textbf{True} & \textbf{Predicted} & \textbf{$R^2$} & \textbf{SSR} \\ \hline
		\endfirsthead
		
		\multicolumn{5}{c}%
		{{\bfseries \tablename\ \thetable{} -- continued}} \\
		\hline
		\textbf{Name} & \textbf{True} & \textbf{Predicted} & \textbf{$R^2$} & \textbf{SSR} \\ \hline
		\endhead
		
		\hline \multicolumn{5}{|r|}{{Continued}} \\ \hline
		\endfoot
		
		\hline
		\endlastfoot
		II.6.15a & $\frac{3 x_{2} x_{6} \sqrt{x_{4}^{2} + x_{5}^{2}}}{4 \pi x_{1} x_{3}^{5}}$ & $\frac{x_{2} x_{6} (0.244 \log{(x_{4} + x_{5} )}^{2} + 0.223)}{x_{1} x_{3}^{5}}$ & 0.9995 & 0 \\ \hline
		test\_12 & $\frac{x_{1} (- \frac{x_{1} x_{2}^{3} x_{4}}{(x_{2}^{2} - x_{4}^{2})^{2}} + 4 \pi x_{3} x_{4} x_{5})}{4 \pi x_{2}^{2} x_{5}}$ & $-11x_1x_3\log((0.000038x_5 + 1.034)/(0.035x_4 + 1)^{3.05})/{x_2}^2$ & 1.0000 & 0 \\ \hline
		II.24.17 & $\sqrt{\frac{x_{1}^{2}}{x_{2}^{2}} - \frac{\pi^{2}}{x_{3}^{2}}}$ & $\sqrt{\frac{x_{1}^{2}}{x_{2}^{2}} - \frac{\pi^{2}}{x_{3}^{2}}}$ & 1.0000 & 1 \\ \hline
		III.15.14 & $\frac{x_{1}^{2}}{8 \pi^{2} x_{2} x_{3}^{2}}$ & $\frac{x_{1}^{2}}{8 \pi^{2} x_{2} x_{3}^{2}}$ & 1.0000 & 1 \\ \hline
		II.11.28 & $\frac{x_{1} x_{2}}{- \frac{x_{1} x_{2}}{3} + 1} + 1$ & $1.244 e^{0.787 x_{1} x_{2}} - 0.243$ & 1.0000 & 0 \\ \hline
		test\_11 & $\frac{4 x_{1} \sin^{2}{(\frac{x_{2}}{2} )} \sin^{2}{(\frac{x_{3} x_{4}}{2} )}}{x_{2}^{2} \sin^{2}{(\frac{x_{3}}{2} )}}$ & $(x_{1} \log{(x_{4}^{3} )} \cos{(0.7 x_{3} )} + x_{1}) \cos{(\log{(x_{2} )} )}$ & 0.9736 & 0 \\ \hline
		II.36.38 & $\frac{x_{1} x_{2}}{x_{3} x_{4}} + \frac{x_{1} x_{5} x_{8}}{x_{3} x_{4} x_{6} x_{7}^{2}}$ & $\frac{x_{1} (x_{2} + \frac{x_{5} x_{8}}{x_{6} x_{7}^{2}})}{x_{3} x_{4}}$ & 1.0000 & 1 \\ \hline
		I.44.4 & $x_{1} x_{2} x_{3} \log{(\frac{x_{5}}{x_{4}} )}$ & $x_{1} x_{2} x_{3} \log{(\frac{x_{5}}{x_{4}} )}$ & 1.0000 & 1 \\ \hline
		I.24.6 & $\frac{x_{1} x_{4}^{2} (x_{2}^{2} + x_{3}^{2})}{4}$ & $0.25 x_{1} x_{4}^{2} (x_{2}^{2} + x_{3}^{2})$ & 1.0000 & 1 \\ \hline
		I.15.3x & $\frac{x_{1} - x_{2} x_{4}}{\sqrt{- \frac{x_{2}^{2}}{x_{3}^{2}} + 1}}$ & $\frac{0.375 x_{1} x_{2}^{2}}{x_{3}^{2}} + x_{1} - x_{2} x_{4}$ & 0.9995 & 0 \\ \hline
		II.11.3 & $\frac{x_{1} x_{2}}{x_{3} (x_{4}^{2} - x_{5}^{2})}$ & $\frac{x_{1} x_{2}}{x_{3} (x_{4} - x_{5}) (x_{4} + x_{5})}$ & 1.0000 & 1 \\ \hline
		III.9.52 & $\frac{8 \pi x_{1} x_{2} \sin^{2}{(\frac{x_{3} (x_{5} - x_{6})}{2} )}}{x_{3} x_{4} (x_{5} - x_{6})^{2}}$ & $\frac{6.124 x_{1} x_{2} x_{3} \cos^{3}{(0.222 x_{3} (- x_{5} + x_{6}) )}}{x_{4}}$ & 0.9913 & 0 \\ \hline
		III.21.20 & $\frac{x_{1} x_{2} x_{3}}{x_{4}}$ & $\frac{x_{1} x_{2} x_{3}}{x_{4}}$ & 1.0000 & 1 \\ \hline
		test\_2 & $\frac{x_{1} x_{2} (\sqrt{1 + \frac{2 x_{3}^{2} x_{4}}{x_{1} x_{2}^{2}}} \cos{(x_{5} - x_{6} )} + 1)}{x_{3}^{2}}$ & $0.034x_1(0.17{x_2}^{2.932}x_6(x_1 + cos(x_4))(x_5 + 3.881)/{x_3}^{5.356} + 11.667)$ & 0.2590 & 1 \\ \hline
		I.50.26 & $x_{1} (x_{4} \cos^{2}{(x_{2} x_{3} )} + \cos{(x_{2} x_{3} )})$ & $x_{1} (x_{4} \cos^{2}{(x_{2} x_{3} )} + \cos{(x_{2} x_{3} )})$ & 1.0000 & 1 \\ \hline
		I.41.16 & $\frac{x_{1}^{3} x_{3}}{2 \pi^{3} x_{5}^{2} (e^{\frac{x_{1} x_{3}}{2 \pi x_{2} x_{4}}} - 1)}$ & $00.0063{x_1}^2(17.08x_2 - 1.46x_3)(x_4 - 0.25)/{x_5}^2.008$ & 0.9993 & 0 \\ \hline
		I.10.7 & $\frac{x_{1}}{\sqrt{- \frac{x_{2}^{2}}{x_{3}^{2}} + 1}}$ & $\frac{x_{1} (0.153 x_{2} e^{\frac{2.352x_{2}}{x_{3}}} + x_{3})}{x_{3}}$ & 0.9996 & 0 \\ \hline
		II.6.11 & $\frac{x_{2} \cos{(x_{3} )}}{4 \pi x_{1} x_{4}^{2}}$ & $\frac{0.8 x_{2} \cos{(x_{3} )}}{\pi^{2} x_{1} x_{4}^{2}}$ & 1.0000 & 1 \\ \hline
		I.25.13 & $\frac{x_{1}}{x_{2}}$ & $\frac{x_{1}}{x_{2}}$ & 1.0000 & 1 \\ \hline
		test\_3 & $\frac{x_{1} (1 - x_{2}^{2})}{x_{2} \cos{(x_{3} - x_{4} )} + 1}$ & $- x_{1} x_{2} - 0.736 x_{1} (- x_{3} + x_{4})^{2} \log{(x_{2} )}^{2} + x_{1}$ & 0.9995 & 0 \\ \hline
		III.8.54 & $\sin^{2}{(\frac{2 \pi x_{1} x_{2}}{x_{3}} )}$ & $0.5 - 0.5 \cos{(\frac{12.566 x_{1} x_{2}}{x_{3}} + 18.85 )}$ & 1.0000 & 0 \\ \hline
		test\_18 & $\frac{\frac{3 x_{2} x_{5}^{2}}{x_{3}^{2}} + 3 x_{4}^{2}}{8 \pi x_{1}}$ & $\frac{\frac{0.12 x_{2} x_{5}^{2}}{x_{3}^{2}} + 0.12 x_{4}^{2}}{x_{1}}$ & 1.0000 & 1 \\ \hline
		II.34.2 & $\frac{x_{1} x_{2} x_{3}}{2}$ & $\frac{x_{1} x_{2} x_{3}}{2}$ & 1.0000 & 1 \\ \hline
		II.38.3 & $\frac{x_{1} x_{2} x_{4}}{x_{3}}$ & $\frac{x_{1} x_{2} x_{4}}{x_{3}}$ & 1.0000 & 1 \\ \hline
		I.6.2b & $\frac{\sqrt{2} e^{- \frac{(x_{2} - x_{3})^{2}}{2 x_{1}^{2}}}}{2 \sqrt{\pi} x_{1}}$ & $\frac{0.4 e^{- \frac{0.5 (x_{2} - x_{3})^{2}}{x_{1}^{2}}}}{x_{1}}$ & 1.0000 & 1 \\ \hline
		II.11.17 & $x_{1} (1 + \frac{x_{5} x_{6} \cos{(x_{4} )}}{x_{2} x_{3}})$ & $x_{1} + \frac{x_{1} x_{5} x_{6} \cos{(x_{4} )}}{x_{2} x_{3}}$ & 1.0000 & 1 \\ \hline
		test\_7 & $\sqrt{\frac{8 \pi x_{1} x_{2}}{3} - \frac{x_{3} x_{4}^{2}}{x_{5}^{2}}}$ & $- \frac{0.545 x_{3} \log{(x_{4} )}}{x_{5}^{2}} + \log{(19.577 (0.254 x_{1} x_{2} + 1)^{6} )} - 1.4$ & 0.9991 & 0 \\ \hline
		III.4.32 & $\frac{1}{e^{\frac{x_{1} x_{2}}{2 \pi x_{3} x_{4}}} - 1}$ & $\frac{6.4 x_{3} x_{4}}{x_{1} x_{2}}$ & 0.9996 & 0 \\ \hline
		I.34.8 & $\frac{x_{1} x_{2} x_{3}}{x_{4}}$ & $\frac{x_{1} x_{2} x_{3}}{x_{4}}$ & 1.0000 & 1 \\ \hline
		III.13.18 & $\frac{4 \pi x_{1} x_{2}^{2} x_{3}}{x_{4}}$ & $\frac{12.566 x_{1} x_{2}^{2} x_{3}}{x_{4}}$ & 1.0000 & 1 \\ \hline
		I.11.19 & $x_{1} x_{4} + x_{2} x_{5} + x_{3} x_{6}$ & $x_{1} x_{4} + x_{2} x_{5} + x_{3} x_{6}$ & 1.0000 & 1 \\ \hline
		II.15.5 & $x_{1} x_{2} \cos{(x_{3} )}$ &  $x_{1} x_{2} \cos{(x_{3} )}$ & 1.0000 & 1 \\ \hline
		II.11.20 & $\frac{x_{1} x_{2}^{2} x_{3}}{3 x_{4} x_{5}}$ & $\frac{x_{1} x_{2}^{2} x_{3}}{3 x_{4} x_{5}}$ & 1.0000 & 1 \\ \hline
		I.38.12 & $\frac{x_{3}^{2} x_{4}}{\pi x_{1} x_{2}^{2}}$ & $\frac{0.318 x_{3}^{2} x_{4}}{x_{1} x_{2}^{2}}$ & 1.0000 & 1 \\ \hline
		II.27.16 & $x_{1} x_{2} x_{3}^{2}$ & $x_{1} x_{2} x_{3}^{2}$ & 1.0000 & 1 \\ \hline
		I.13.12 & $x_{1} x_{2} x_{5} (\frac{1}{x_{4}} - \frac{1}{x_{3}})$ & $\frac{x_{1} x_{2} x_{5} (x_{3} - x_{4})}{x_{3} x_{4}}$ & 1.0000 & 1 
		\\ \hline
		II.13.23 & $\frac{x_{1}}{\sqrt{- \frac{x_{2}^{2}}{x_{3}^{2}} + 1}}$ & $x_{1} (0.237 e^{\frac{2 x_{2}^{2}}{x_{3}^{2}}} + 0.764)$ & $ 0.9997$ & 0 \\ \hline
		I.29.4 & $\frac{x_{1}}{x_{2}}$ & $\frac{x_{1}}{x_{2}}$ & 1.0000 & 1 \\ \hline
		I.29.16 & $\sqrt{x_{1}^{2} - 2 x_{1} x_{2} \cos{(x_{3} - x_{4} )} + x_{2}^{2}}$ & $1.537 \log{((x_{1} - x_{2})^{2} + (x_{3} - x_{4})^{2} + 0.5 )} + 1.482$ & 0.4948 & 0 \\ \hline
		I.43.31 & $x_{1} x_{2} x_{3}$ & $x_{1} x_{2} x_{3}$ & 1.0000 & 1 \\ \hline
		II.34.29a & $\frac{x_{1} x_{2}}{4 \pi x_{3}}$ & $\frac{x_{1} x_{2}}{4 \pi x_{3}}$ & 1.0000 & 1 \\ \hline
		I.14.3 & $x_{1} x_{2} x_{3}$ & $x_{1} x_{2} x_{3}$ & 1.0000 & 1 \\ \hline
		II.27.18 & $x_{1} x_{2}^{2}$ & $x_{1} x_{2}^{2}$ & 1.0000 & 1 \\ \hline
		I.15.3t & $\frac{- \frac{x_{1} x_{3}}{x_{2}^{2}} + x_{4}}{\sqrt{1 - \frac{x_{3}^{2}}{x_{2}^{2}}}}$ & $\frac{x_{2}^{2} x_{4} - 0.6 x_{3} (1.879 x_{1} - x_{3} x_{4})}{x_{2}^{2}}$ & 0.9999 & 0 \\ \hline
		II.2.42 & $\frac{x_{1} x_{4} (- x_{2} + x_{3})}{x_{5}}$ & $\frac{x_{1} x_{4} (- x_{2} + x_{3})}{x_{5}}$ & 1.0000 & 1 \\ \hline
		I.34.27 & $\frac{x_{1} x_{2}}{2 \pi}$ & $0.16 x_{2} \log{(e^{x_{1}} )}$ & 1.0000 & 0 \\ \hline
		II.34.11 & $\frac{x_{1} x_{2} x_{3}}{2 x_{4}}$ & $\frac{x_{1} x_{2} x_{3}}{2 x_{4}}$ & 1.0000 & 1 \\ \hline
		III.15.27 & $\frac{2 \pi x_{1}}{x_{2} x_{3}}$ & $\frac{6.283 x_{1}}{x_{2} x_{3}}$ & 1.0000 & 1 \\ \hline
		III.14.14 & $x_{1} (e^{\frac{x_{2} x_{3}}{x_{4} x_{5}}} - 1)$ & $x_{1} (e^{\frac{x_{2} x_{3}}{x_{4} x_{5}}} - 1)$ & 1.0000 & 1 \\ \hline
		I.12.4 & $\frac{x_{1}}{4 \pi x_{2} x_{3}^{2}}$ & $\frac{x_{1}}{4 \pi x_{2} x_{3}^{2}}$ & 1.0000 & 1 \\ \hline
		I.48.2 & $\frac{x_{1} x_{3}^{2}}{\sqrt{- \frac{x_{2}^{2}}{x_{3}^{2}} + 1}}$ & $x_{1} (x_{3}^{2} + \log{(x_{2}^{3} )})$ & 0.9993 & 0 \\ \hline
		test\_5 & $\frac{2 \pi x_{1}^{\frac{3}{2}}}{\sqrt{x_{2} (x_{3} + x_{4})}}$ & $\frac{2 \pi x_{1}^{1.5}}{(x_{2} (x_{3} + x_{4}))^{0.5}}$ & 1.0000 & 1 \\ \hline
		II.4.23 & $\frac{x_{1}}{4 \pi x_{2} x_{3}}$ & $\frac{x_{1}}{4 \pi x_{2} x_{3}}$ & 1.0000 & 1 \\ \hline
		test\_8 & $\frac{x_{1}}{\frac{x_{1} (1 - \cos{(x_{4} )})}{x_{2} x_{3}^{2}} + 1}$ & $0.2 x_{1} \log{(x_{3} )} + 0.725 x_{1} e^{- \frac{x_{4}}{x_{2} x_{3}}} + 0.355$ & 0.9859 & 0 \\ \hline
		II.11.27 & $\frac{x_{1} x_{2} x_{3} x_{4}}{- \frac{x_{1} x_{2}}{3} + 1}$ & $\frac{x_{1} x_{2} x_{3} x_{4}}{- 0.333 x_{1} x_{2} + 1}$ & 1.0000 & 1 \\ \hline
		I.39.1 & $\frac{3 x_{1} x_{2}}{2}$ & $\frac{3 x_{1} x_{2}}{2}$ & 1.0000 & 1 \\ \hline
		I.6.2 & $\frac{\sqrt{2} e^{- \frac{x_{2}^{2}}{2 x_{1}^{2}}}}{2 \sqrt{\pi} x_{1}}$ & $\frac{1.32 e^{- \frac{0.5 x_{2}^{2}}{|{x_{1}}|^{2}}}}{\pi x_{1}}$ & 1.0000 & 0 \\ \hline
		III.10.19 & $x_{1} \sqrt{x_{2}^{2} + x_{3}^{2} + x_{4}^{2}}$ & $x_{1} \sqrt{x_{2}^{2} + x_{3}^{2} + x_{4}^{2}}$ & 1.0000 & 1 \\ \hline
		test\_1 & $\frac{x_{1}^{2} x_{2}^{2} x_{3}^{2} x_{4}^{2} x_{5}^{2}}{16 x_{6}^{2} \sin^{4}{(\frac{x_{7}}{2} )}}$ & $\frac{1.1 x_{1}^{2} x_{2}^{2} x_{3}^{2} x_{4}^{2} x_{5}^{2}}{x_{6}^{2} x_{7}^{3}}$ & 0.9942 & 0 \\ \hline
		I.40.1 & $x_{1} e^{- \frac{x_{2} x_{3} x_{5}}{x_{4} x_{6}}}$ & $x_{1} e^{- \frac{x_{2} x_{3} x_{5}}{x_{4} x_{6}}}$ & 1.0000 & 1 \\ \hline
		I.43.16 & $\frac{x_{1} x_{2} x_{3}}{x_{4}}$ & $\frac{x_{1} x_{2} x_{3}}{x_{4}}$ & 1.0000 & 1 \\ \hline
		I.37.4 & $x_{1} + x_{2} + 2 \sqrt{x_{1} x_{2}} \cos{(x_{3} )}$ & $x_{1} + x_{2} + 2 (x_{1} x_{2})^{0.5} \cos{(x_{3} + 12.566 )}$ & 0.9991 & 0 \\ \hline
		I.12.5 & $x_{1} x_{2}$ & $x_{1} x_{2}$ & 1.0000 & 1 \\ \hline
		I.34.1 & $\frac{x_{3}}{1 - \frac{x_{2}}{x_{1}}}$ & $\frac{x_{2} x_{3}}{x_{1} - x_{2}} + x_{3}$ & 1.0000 & 1 \\ \hline
		I.12.2 & $\frac{x_{1} x_{2}}{4 \pi x_{3} x_{4}^{2}}$ & $\frac{x_{1} x_{2}}{4 \pi x_{3} x_{4}^{2}}$ & 1.0000 & 1 \\ \hline
		test\_20 & $\frac{x_{2}^{2} x_{3}^{2} x_{4}^{2} (\frac{x_{1}}{x_{2}} - \sin^{2}{(x_{7} )} + \frac{x_{2}}{x_{1}})}{4 \pi x_{1}^{2} x_{5}^{2} x_{6}^{2}}$ & $\frac{x_{1} x_{2} x_{7}}{x_{3} x_{4} x_{5} x_{6}^{3}} + 1$ & -0.0020 & 0 \\ \hline
		II.37.1 & $x_{1} x_{2} (x_{3} + 1)$ & $x_{1} x_{2} (x_{3} + 1)$ & 1.0000 & 1 \\ \hline
		test\_6 & $\sqrt{\frac{2 x_{1}^{2} x_{2}^{2} x_{7}}{x_{3} x_{4}^{2} x_{5}^{2} x_{6}^{4}} + 1}$ & $\frac{x_{1} x_{2} x_{7}}{x_{3} x_{4} x_{5} x_{6}^{3}} + 1$ & 0.9049 & 0 \\ \hline
		III.19.51 & $- \frac{x_{1} x_{2}^{4}}{8 x_{3}^{2} x_{4}^{2} x_{5}^{2}}$ & $- \frac{0.125 x_{1} x_{2}^{4}}{x_{3}^{2} x_{4}^{2} x_{5}^{2}}$ & 1.0000 & 1 \\ \hline
		III.4.33 & $\frac{x_{1} x_{2}}{2 \pi (e^{\frac{x_{1} x_{2}}{2 \pi x_{3} x_{4}}} - 1)}$ & $\frac{38 x_{3} (x_{4} - 0.183)}{\pi^{3.167}}$ & 0.9981 & 0 \\ \hline
		I.32.5 & $\frac{x_{1}^{2} x_{2}^{2}}{6 \pi x_{3} x_{4}^{3}}$ & $ \frac{0.053{x_1}^{2}x_2^{2}}{x_3{x_4}^3}$ & 1.0000 & 1 \\ \hline
		I.13.4 & $\frac{x_{1} (x_{2}^{2} + x_{3}^{2} + x_{4}^{2})}{2}$ & $x_{1} (x_{2} x_{4} + \frac{x_{3}^{2}}{2} + \frac{(- x_{2} + x_{4})^{2}}{2})$ & 1.0000 & 1 \\ \hline
		II.13.17 & $\frac{x_{3}}{2 \pi x_{1} x_{2}^{2} x_{4}}$ & $\frac{x_{3}}{2 \pi x_{1} x_{2}^{2} x_{4}}$ & 1.0000 & 1 \\ \hline
		II.10.9 & $\frac{x_{1}}{x_{2} (x_{3} + 1)}$ & $\frac{x_{1}}{x_{2} x_{3} + x_{2}}$ & 1.0000 & 1 \\ \hline
		III.7.38 & $\frac{4 \pi x_{1} x_{2}}{x_{3}}$ & $\frac{12.566 x_{1} x_{2}}{x_{3}}$ & 1.0000 & 1 \\ \hline
		I.32.17 & $\frac{4 \pi x_{1} x_{2} x_{3}^{2} x_{4}^{2} x_{5}^{4}}{3 (x_{5}^{2} - x_{6}^{2})^{2}}$ & $x_1x_2{x_3}^{2}{x_4}^{2}(0.0029e^{\frac{10.217x_5}{x_6}} - 0.022)$ & 0.9998 & 0 \\ \hline
		test\_14 & $x_{1} (- x_{3} + \frac{x_{4}^{3} (x_{5} - 1)}{x_{3}^{2} (x_{5} + 2)}) \cos{(x_{2} )}$ & $x_{1} (- x_{3} + \frac{0.36 x_{4}^{3} \log{(x_{5} )}}{x_{3}^{2}}) \cos{(x_{2} )}$ & 0.9999 & 0 \\ \hline
		test\_13 & $\frac{x_{1}}{4 \pi x_{5} \sqrt{x_{2}^{2} - 2 x_{2} x_{3} \cos{(x_{4} )} + x_{3}^{2}}}$ & $\frac{0.116 x_{1}}{x_{3}^{1.19} x_{5}^{1.01} \left(x_{4} + 0.034\right)^{0.162} \sin^{0.088}{\left(x_{2} \right)}}$ & 0.7880 & 0 \\ \hline
		I.27.6 & $\frac{1}{\frac{x_{3}}{x_{2}} + \frac{1}{x_{1}}}$ & $x_{2} (1.117 e^{- 0.552 x_{3} - \frac{0.57 x_{2}}{x_{1}}} + 0.128)$ & 0.9976 & 0 \\ \hline
		II.3.24 & $\frac{x_{1}}{4 \pi x_{2}^{2}}$ & $\frac{0.08x_1}{{x_2}^2}$ & 1.0000 & 1 \\ \hline
		I.18.14 & $x_{1} x_{2} x_{3} \sin{(x_{4} )}$ & $x_{1} x_{2} x_{3} \sin{(x_{4} )}$ & 1.0000 & 1 \\ \hline
		II.8.7 & $\frac{3 x_{1}^{2}}{20 \pi x_{2} x_{3}}$ & $\frac{0.048 x_{1}^{2}}{x_{2} x_{3}}$ & 1.0000 & 1 \\ \hline
		I.16.6 & $\frac{x_{2} + x_{3}}{1 + \frac{x_{2} x_{3}}{x_{1}^{2}}}$ & $x_{1} + \frac{(- x_{1} + x_{2}) \log{(\frac{x_{1}}{x_{3}} )}}{x_{2}}$ & 0.9803 & 0 \\ \hline
		I.30.5 & $\arcsin{(\frac{x_{1}}{x_{2} x_{3}} )}$ & $\arcsin{(\frac{x_{1}}{x_{2} x_{3}} )}$ & 1.0000 & 0 \\ \hline
		II.35.21 & $x_{1} x_{2} \tanh{(\frac{x_{2} x_{3}}{x_{4} x_{5}} )}$ & $x_{1} (x_{2} + \frac{12.262 x_{4} (-0.85 + \sin{(1 )})}{x_{3}})$ & 0.9660 & 0 \\ \hline
		I.12.1 & $x_{1} x_{2}$ & $x_{1} x_{2}$ & 1.0000 & 1 \\ \hline
		I.15.1 & $\frac{x_{1} x_{2}}{\sqrt{- \frac{x_{2}^{2}}{x_{3}^{2}} + 1}}$ & $x_{1} x_{2} (0.238 e^{\frac{2 x_{2}^{2}}{x_{3}^{2}}} + 0.763)$ & 0.9993 & 0 \\ \hline
		II.6.15b & $\frac{3 x_{2} \sin{(x_{3} )} \cos{(x_{3} )}}{4 \pi x_{1} x_{4}^{3}}$ & $\frac{0.12 x_{2} \sin{(2 x_{3} )}}{x_{1} x_{4}^{3}}$ & 1.0000 & 0 \\ \hline
		test\_15 & $\frac{x_{3} \sqrt{1 - \frac{x_{2}^{2}}{x_{1}^{2}}}}{1 + \frac{x_{2} \cos{(x_{4} )}}{x_{1}}}$ & $\frac{x_{3} (x_{1} e^{- \frac{x_{2} \cos{(x_{4} )}}{x_{1}}} - 0.136 x_{2})}{x_{1}}$ & 0.9986 & 0 \\ \hline
		I.9.18 & $\frac{x_{1} x_{2} x_{3}}{(- x_{4} + x_{5})^{2} + (- x_{6} + x_{7})^{2} + (- x_{8} + x_{9})^{2}}$ & $\frac{x_{1} \left(x_{2} + \frac{x_{5} x_{8}}{x_{6} x_{7}^{2}}\right)}{x_{3} x_{4}^{1.767}}$ & 0.4030 & 0 \\ \hline
		I.34.14 & $\frac{x_{3} (1 + \frac{x_{2}}{x_{1}})}{\sqrt{1 - \frac{x_{2}^{2}}{x_{1}^{2}}}}$ & $\frac{x_{3} (x_{1} + 0.178 x_{2})}{x_{1} - 0.733 x_{2}}$ & 1.0000 & 0 \\ \hline
		I.14.4 & $\frac{x_{1} x_{2}^{2}}{2}$ & $\frac{x_{1} x_{2}^{2}}{2}$ & 1.0000 & 1 \\ \hline
		test\_16 & $x_{4} x_{6} + \sqrt{x_{1}^{2} x_{2}^{4} + x_{2}^{2} (x_{3} - x_{4} x_{5})^{2}}$ & $x_{1} (x_{2}^{2} - 1.639) - x_{3} + x_{4} (1.754 x_{5} + x_{6})$ & 0.9784 & 0 \\ \hline
		II.13.34 & $\frac{x_{1} x_{2}}{\sqrt{- \frac{x_{2}^{2}}{x_{3}^{2}} + 1}}$ & $x_{1} x_{2} (0.237 e^{\frac{2 x_{2}^{2}}{x_{3}^{2}}} + 0.764)$ & 0.9996 & 0 \\ \hline
		test\_19 & $\frac{- \frac{x_{2} x_{6}^{4}}{x_{3}^{2}} - x_{4}^{2} x_{6}^{2} (1 - 2 x_{5})}{8 \pi x_{1}}$ & $\frac{3 x_{3}^{0.162} x_{6} (x_{4} + \cos{(1.234 x_{4} )}) \log{(x_{5} )} - 8.644}{x_{1}}$ & 0.8917 & 0 \\ \hline
		test\_9 & $- \frac{32 x_{1}^{4} x_{3}^{2} x_{4}^{2} (x_{3} + x_{4})}{5 x_{2}^{5} x_{5}^{5}}$ & $- \frac{6.2 x_{1}^{4} x_{3}^{3} x_{4}^{2.589}}{x_{2}^{5.3} x_{5}^{5.5}}$ & 0.9766 & 0 \\ \hline
		I.26.2 & $\arcsin{(x_{1} \sin{(x_{2} )} )}$ & $\arcsin{(x_{1} \sin{(x_{2} )} )}$ & 1.0000 & 1 \\ \hline
		I.8.14 & $\sqrt{(- x_{1} + x_{2})^{2} + (- x_{3} + x_{4})^{2}}$ & $\sqrt{(- x_{1} + x_{2})^{2} + (- x_{3} + x_{4})^{2}}$ & 1.0000 & 1 \\ \hline
		II.21.32 & $\frac{x_{1}}{4 \pi x_{2} x_{3} (- \frac{x_{4}}{x_{5}} + 1)}$ & $\frac{0.2 x_{1} x_{4}^{0.414}}{x_{2} x_{3} x_{5}^{0.43}}$ & 0.9921 & 0 \\ \hline
		I.43.43 & $\frac{x_{2} x_{4}}{x_{3} (x_{1} - 1)}$ & $\frac{x_{2} x_{4}}{x_{3} (x_{1} - 1)}$ & 1.0000 & 1 \\ \hline
		I.39.22 & $\frac{x_{1} x_{2} x_{4}}{x_{3}}$ & $\frac{x_{1} x_{2} x_{4}}{x_{3}}$ & 1.0000 & 1 \\ \hline
		III.12.43 & $\frac{x_{1} x_{2}}{2 \pi}$ & $0.16 x_{1} x_{2}$ & 1.0000 & 1 \\ \hline
		test\_4 & $\sqrt{2} \sqrt{\frac{x_{2} - x_{3} - \frac{x_{4}^{2}}{2 x_{1} x_{5}^{2}}}{x_{1}}}$ & $1.461 \sqrt{\frac{x_{2} - x_{3}}{x_{1}}} + 0.421 \cos{(\frac{x_{4}}{x_{1} x_{5}} )} - 0.5$ & 0.9988 & 0 \\ \hline
		I.39.11 & $\frac{x_{2} x_{3}}{x_{1} - 1}$ & $\frac{x_{2} x_{3}}{x_{1} - 1}$ & 1.0000 & 1 \\ \hline
		I.47.23 & $\sqrt{\frac{x_{1} x_{2}}{x_{3}}}$ & $\sqrt{\frac{x_{1} x_{2}}{x_{3}}}$ & 1.0000 & 1 \\ \hline
		I.6.2a & $\frac{\sqrt{2} e^{- \frac{x_{1}^{2}}{2}}}{2 \sqrt{\pi}}$ & $0.4 e^{- \frac{x_{1}^{2}}{2}}$ & 1.0000 & 1 \\ \hline
		I.30.3 & $\frac{x_{1} \sin^{2}{(\frac{x_{2} x_{3}}{2} )}}{\sin^{2}{(\frac{x_{2}}{2} )}}$ & $\frac{x_{1} (1 - \cos{(x_{2} x_{3} )})}{1 - \cos{(x_{2} )}}$ & 1.0000 & 1 \\ \hline
		II.34.2a & $\frac{x_{1} x_{2}}{2 \pi x_{3}}$ & $\frac{5.7 x_{2} |{x_{1}}|}{\pi^{3.127} x_{3}}$ & 1.0000 & 0 \\ \hline
		II.34.29b & $\frac{2 \pi x_{1} x_{3} x_{4} x_{5}}{x_{2}}$ & $\frac{2 \pi x_{1} x_{3} x_{4} x_{5}}{x_{2}}$ & 1.0000 & 1 \\ \hline
		I.12.11 & $x_{1} (x_{2} + x_{3} x_{4} \sin{(x_{5} )})$ & $x_{1} (x_{2} + x_{3} x_{4} \sin{(x_{5} )})$ & 1.0000 & 1 \\ \hline
		III.15.12 & $2 x_{1} (1 - \cos{(x_{2} x_{3} )})$ & $2 x_{1} (1 - \cos{(x_{2} x_{3} )})$ & 1.0000 & 1 \\ \hline
		II.38.14 & $\frac{x_{1}}{2 x_{2} + 2}$ & $\frac{x_{1}}{2 x_{2} + 2}$ & 1.0000 & 1 \\ \hline
		test\_17 & $\frac{x_{1}^{2} x_{2}^{2} x_{5}^{2} (1 + \frac{x_{5} x_{6}}{x_{4}}) + x_{3}^{2}}{2 x_{1}}$ & $0.5 x_{1} x_{2}^{2} x_{5} (x_{5} + \frac{x_{5}^{2} x_{6}}{x_{4}}) + x_{3}$ & 0.9992 & 0 \\ \hline
		II.8.31 & $\frac{x_{1} x_{2}^{2}}{2}$ & $\frac{x_{1} x_{2}^{2}}{2}$ & 1.0000 & 1 \\ \hline
		I.18.12 & $x_{1} x_{2} \sin{(x_{3} )}$ & $x_{1} x_{2} \sin{(x_{3} )}$ & 1.0000 & 1 \\ \hline
		test\_10 & $\arccos{(\frac{\cos{(x_{3} )} - \frac{x_{2}}{x_{1}}}{1 - \frac{x_{2} \cos{(x_{3} )}}{x_{1}}} )}$ & $\frac{x_{2} \sin{(x_{3} )}}{x_{1} + x_{3} - 1.821} + x_{3}$ & 0.9995 & 0 \\ \hline
		I.18.4 & $\frac{x_{1} x_{3} + x_{2} x_{4}}{x_{1} + x_{2}}$ & $\frac{x_{1} x_{3} + x_{2} x_{4}}{x_{1} + x_{2}}$ & 1.0000 & 1 \\ \hline
		III.17.37 & $x_{1} (x_{2} \cos{(x_{3} )} + 1)$ & $x_{1} (x_{2} \cos{(x_{3} )} + 1)$ & 1.0000 & 1 \\ \hline
		II.35.18 & $\frac{x_{1}}{e^{\frac{x_{4} x_{5}}{x_{2} x_{3}}} + e^{- \frac{x_{4} x_{5}}{x_{2} x_{3}}}}$ & $0.47 x_{1} \sin{(2 e^{- \frac{x_{4} x_{5}}{x_{2} x_{3}}} )}$ & 0.9995 & 0 \\ \hline
		II.15.4 & $x_{1} x_{2} \cos{(x_{3} )}$ & $x_{1} x_{2} \cos(x_3)$ & 1.0000 & 1 \\ \hline
	\end{longtable}
}

\section{Supplementary results on ODE-Strogatz benchmark}
\label{Supplementary results on ODE-Strogatz benchmark}

{
	\renewcommand{\arraystretch}{2.0}
	\begin{longtable}{| >{\centering\arraybackslash}m{0.1\linewidth} | >{\centering\arraybackslash}m{0.3\linewidth} | >{\centering\arraybackslash}m{0.3\linewidth} | >{\centering\arraybackslash}m{0.09\linewidth} | >{\centering\arraybackslash}m{0.06\linewidth} |}
		\caption{Details results on ODE-Strogatz benchmark} \label{tab:details_ode_inside} \\

		\hline
		\textbf{Name} & \textbf{True} & \textbf{Predicted} & \textbf{$R^2$} & \textbf{SSR} \\ \hline
		\endfirsthead
		
		\multicolumn{5}{c}%
		{{\bfseries \tablename\ \thetable{} -- continued}} \\
		\hline
		\textbf{Name} & \textbf{True} & \textbf{Predicted} & \textbf{$R^2$} & \textbf{SSR} \\ \hline
		\endhead
		
		\hline \multicolumn{5}{|r|}{{Continued}} \\ \hline
		\endfoot
		
		\hline
		\endlastfoot
		d\_bacres1 & $- \frac{x_{1} x_{2}}{0.5 x_{1}^{2} + 1} - x_{1} + 20$ & $- \frac{1.195 x_{1}^{1.077} e^{\frac{0.395 \log{\left(x_{2} \right)}^{2.958}}{x_{1}^{1.455}}}}{\pi^{0.517}} + 17.296$ & 0.9977 & 0 \\ \hline
		d\_bacres2 & $- \frac{x_{1} x_{2}}{0.5 x_{1}^{2} + 1} + 10$ & $- \frac{1.4 x_{2}}{x_{1}^{0.7}} + 10.567$ & 0.9999 & 0 \\ \hline
		d\_barmag1 & $-sin(x_1)+0.5sin(x_1-x_2)$ & $\frac{(\cos{(x_{2} - 24.16 )} + 1.842) \cos{(0.264 x_{1} )}}{(\sin{(x_{2} + 2.5 )} + 1)^{0.723}}$ & 0.9563 & 0 \\ \hline
		d\_barmag2 & $-sin(x_1)-0.5sin(x_1-x_2)$ & $-2 \sin{\left(x_{1}^{0.601} \right)} \cos{\left(x_{2} - 1.989 \right)}$ & 0.9788 & 0 \\ \hline
		d\_glider1 & $- 0.05 x_{1}^{2} - \sin{(x_{2} )}$ & $- 0.05 x_{1}^{2} - \sin{(x_{2} )}$ & 1.0000 & 1\\ \hline
		d\_glider2 & $x_{1} - \frac{\cos{(x_{2} )}}{x_{1}}$ & $x_{1} - \frac{\cos{(x_{2} )}}{x_{1}}$ & 1.0000 & 1 \\ \hline
		d\_lv1 & $- x_{1}^{2} - 2 x_{1} x_{2} + 3 x_{1}$ & \tiny{$-0.06 \cdot x_1 \left( 16.782 \cdot x_1 + \frac{33.564 \cdot x_2}{(0.05 \cdot x_1 + 1)^{6.626 \times 10^{-6}}} - 50.345 \right)$} & 0.9999 & 0 \\ \hline
		d\_lv2 & $- x_{1} x_{2} - x_{2}^{2} + 2 x_{2}$ & $x_2 \left( -1 \cdot \frac{x_1}{(x_1 + 0.827)^0} - 1 \cdot x_2 + 2 \right)$ & 1.0000 & 1 \\ \hline
		d\_predprey1 & $x_{1} (- x_{1} - \frac{x_{2}}{x_{1} + 1} + 4)$ & $x_{1} (- x_{1} - \frac{x_{2}}{x_{1} + 1} + 4)$ & 1.0000 & 1 \\ \hline
		d\_predprey2 & $x_{2} (\frac{x_{1}}{x_{1} + 1} - 0.075 x_{2})$ &$x_{2} \left(\frac{x_{1}}{x_{1} + 1} - 0.075 x_{2}\right)$ & 1.0000 & 1 \\ \hline
		d\_shearflow1 & $\cos{(x_{1} )} \cot{(x_{2} )}$ & $\frac{\cos{(x_{1} )} \cos{(x_{2} )}}{\sin{(x_{2} )}}$ & 1.0000 & 1 \\ \hline
		d\_shearflow2 & $(0.1 \sin^{2}{(x_{2} )} + \cos^{2}{(x_{2} )}) \sin{(x_{1} )}$ & $(0.1 \sin^{2}{(x_{2} )} + \cos^{2}{(x_{2} )}) \sin{(x_{1} )}$ & 1.0000 & 1 \\ \hline
		d\_vdp1 & $- \frac{10 x_{1}^{3}}{3} + \frac{10 x_{1}}{3} + 10 x_{2}$ & $0.367 (x_{2} + \sin{(x_{2} - 4.4 )})^{17.862} \sin{(x_{1} - 4.62 )}$ & 0.9293 & 0 \\ \hline
		d\_vdp2 & $-x_1/10$ & $-0.1 x_1$ & 1.0000 & 1 \\ \hline
	\end{longtable}
}

\section{Supplementary results on low-dimensional benchmarks }
\label{Supplementary results on the low-dimensional benchmarks}

{
	\renewcommand{\arraystretch}{2.0}
	
	\begin{longtable}{| >{\centering\arraybackslash}m{0.1\linewidth} | >{\centering\arraybackslash}m{0.23\linewidth} | >{\centering\arraybackslash}m{0.36\linewidth} | >{\centering\arraybackslash}m{0.05\linewidth} | >{\centering\arraybackslash}m{0.03\linewidth} | >{\centering\arraybackslash}m{0.1\linewidth} |}
		\caption{Details results on low-dimensional benchmarks} \label{tab:Details on low-dimensional benchmarks} \\
		\hline
		\textbf{Name} & \textbf{True} & \textbf{Predicted} & \textbf{$R^2$} & \textbf{SSR} & \textbf{Range  $\mathcal{U}(\cdot,\cdot,200)$} \\ \hline
		\endfirsthead
		
		\multicolumn{6}{c}%
		{{\bfseries \tablename\ \thetable{} -- continued}} \\
		\hline
		\textbf{Name} & \textbf{True} & \textbf{Predicted} & \textbf{$R^2$} & \textbf{SSR} & \textbf{Range} \\ \hline
		\endhead
		
		\hline \multicolumn{6}{|r|}{{Continued}} \\ \hline
		\endfoot
		
		\hline
		\endlastfoot
		Keijzer-1 & $0.3 x_{1} \sin{(2 \pi x_{1} )}$ & $0.3 x_{1} \sin{(2 \pi x_{1} )}$ & 1.0000 & 1 & [-1, 1] \\ \hline
		Keijzer-2 & $0.3 x_{1} \sin{(2 \pi x_{1} )}$ & $0.3 x_{1} \sin{(2 \pi x_{1} )}$ & 1.0000 & 1 & [-2, 2] \\ \hline
		Keijzer-3 & $0.3 x_{1} \sin{(2 \pi x_{1} )}$ & $0.3 x_{1} \sin{(2 \pi x_{1} )}$ & 1.0000 & 1 & [-3, 3] \\ \hline
		Keijzer-4 & $x_{1}^{3} (\sin^{2}{(x_{1} )} \cos{(x_{1} )} - 1) e^{- x_{1}} \sin{(x_{1} )} \cos{(x_{1} )}$ & $(7.207 - 11.538 \cos{(1.862 x_{1} - 7.5)}) \sin{(0.041 x_{1}^{3} )}$ & 0.9996 & 0 & [-1, 1] \\ \hline
		Keijzer-6 & $\frac{30 x_{1} x_{2}^{2} x_{3}}{x_{1} - 10}$ & $- \frac{3.003 x_{1} x_{2}^{2} x_{3}}{(1 - 0.062 x_{1})^{1.627}}$ & 0.9999 & 0 & [-1, 1] \\ \hline
		Keijzer-7 & $\log{(x_{1} )}$ & $\log{(x_{1} )}$ & 1.0000 & 1 & [1, 100] \\ \hline
		Keijzer-8 & $\sqrt{x_{1}}$ & $x_{1}^{0.5}$ & 1.0000 & 1 & [0, 100] \\ \hline
		Keijzer-9 & $\log{(x_{1} + \sqrt{x_{1}^{2} + 1} )}$ & $\log{(x_{1} + \sqrt{x_{1}^{2} + 1} )}$ & 1.0000 & 1 & [0, 100] \\ \hline
		Keijzer-10 & $x_{1}^{x_{2}}$ & $x_{1}^{x_{2}}$ & 1.0000 & 1 & [-1, 1] \\ \hline
		Keijzer-11 & $x_{1} x_{2} + \sin{((x_{1} - 1) (x_{2} - 1) )}$ & $- 0.593 x_{2} - 0.412 \sin{\left(x_{1} - 7.699 \right)} - 0.585 \sin{\left(x_{1} - 1.902 x_{2} + 17.855 \right)}$ & 0.9065 & 0 & [-1, 1] \\ \hline
		Keijzer-12 & $x_{1}^{4} - x_{1}^{3} + \frac{x_{2}^{2}}{2} - x_{2}$ & $\frac{(0.224 x_{2} - 0.604) (1.576 x_{2} (0.003 x_{1} + 1)^{1141.628} + 0.203 \sin{(x_{1} )})}{(0.003 x_{1} + 1)^{1141.628}}$ & 0.9836 & 0 & [-1, 1] \\ \hline
		Keijzer-13 & $6 \sin{(x_{1} )} \cos{(x_{2} )}$ & $6 \sin{(x_{1} )} \cos{(x_{2} )}$ & 1.0000 & 1 & [-1, 1] \\ \hline
		Keijzer-14 & $x_{2}^{2} + \frac{8}{x_{1}^{2} + 2}$ & $x_{2}^{2} + 10.422 \sin{(x_{1} + 4.712 )} + 7.927 \sin{(1.339 x_{1} - 4.713 )} + 6.49$ & 0.9999 & 0 & [-1, 1] \\ \hline
		Keijzer-15 & $\frac{x_{1}^{3}}{5} - x_{1} + \frac{x_{2}^{3}}{2} - x_{2}$ & $- 0.883 \sin{(1.137 x_{1} )} - 0.54 \sin{(1.9 x_{2} )}$ & 0.9999 & 0 & [-1, 1] \\ \hline
		Livermore-1 & $x_{1} + \sin{(x_{1}^{2} )} + 0.333$ & $x_{1} + \sin{(x_{1}^{2} )} + 0.333$ & 1.0000 & 1 & [-3, 3] \\ \hline
		Livermore-2 & $\sin{(x_{1}^{2} )} \cos{(x_{1} )} - 2$ & $\sin{(x_{1}^{2} )} \cos{(x_{1} )} - 2$ & 1.0000 & 1 & [-3, 3] \\ \hline
		Livermore-3 & $\sin{(x_{1}^{3} )} \cos{(x_{1}^{2} )} - 1$ & $\sin{(x_{1}^{3} )} \cos{(x_{1}^{2} )} - 1$ & 1.0000 & 1 & [-3, 3] \\ \hline
		Livermore-4 & $\log{(x_{1} )} + \log{(x_{1} + 1 )} + \log{(x_{1}^{2} + 1 )}$ & $\log{(x_{1} )} + \log{(x_{1} + 1 )} + \log{(x_{1}^{2} + 1 )}$ & 1.0000 & 1 & [-3, 3] \\ \hline
		Livermore-5 & $x_{1}^{4} - x_{1}^{3} + x_{2}^{2} - x_{2}$ & $x_{1}^{4} - x_{1}^{3} + x_{2}^{2} - x_{2}$ & 1.0000 & 1 & [-3, 3] \\ \hline
		Livermore-6 & $4 x_{1}^{4} + 3 x_{1}^{3} + 2 x_{1}^{2} + x_{1}$ & $3.16 x_{1}^{3} + 5.444 |{x_{1}}|^{3.706} + 0.008 |{x_{1}}|^{7.261}$ & 0.9999 & 0 & [-3, 3] \\ \hline
		Livermore-7 & $\frac{e^{x_{1}}}{2} - \frac{e^{- x_{1}}}{2}$ & $ (0.173 x_1^3 + x_1)$ & 0.9999 & 0 & [-1, 1] \\ \hline
		Livermore-8 & $\frac{e^{x_{1}}}{2} + \frac{e^{- x_{1}}}{2}$ & $\frac{599.784 e^{4.883 |{x_{1}}|^{0.413}}}{\pi^{10.383}} + 1$ & 0.9999 & 0 & [-3, 3] \\ \hline
		Livermore-9 & $x_{1}^{9} + x_{1}^{8} + x_{1}^{7} + x_{1}^{6} + x_{1}^{5} + x_{1}^{4} + x_{1}^{3} + x_{1}^{2} + x_{1}$ & $0.669 x_{1} + 0.051 e^{5.098 x_{1}} - 0.02 e^{x_{1}^{2}}$ & 0.9995 & 0 & [-1, 1] \\ \hline
		Livermore-10 & $6 \sin{(x_{1} )} \cos{(x_{2} )}$ & $6 \sin{(x_{1} )} \cos{(x_{2} )}$ & 1.0000 & 1 & [-3, 3] \\ \hline
		Livermore-11 & $\frac{x_{1}^{2} x_{2}^{2}}{x_{1} + x_{2}}$ & $\frac{x_{1}^{2} x_{2}^{2}}{x_{1} + x_{2}}$ & 1.0000 & 1 & [-3, 3] \\ \hline
		Livermore-12 & $\frac{x_{1}^{5}}{x_{2}^{3}}$ & $\frac{x_{1}^{5}}{x_{2}^{3}}$ & 1.0000 & 1 & [-3, 3] \\ \hline
		Livermore-13 & $\sqrt[3]{x_{1}}$ & ${x_{1}}^{\frac{1}{3}}$ & 1.0000 & 1 & [-3, 3] \\ \hline
		Livermore-14 & $x_{1}^{3} + x_{1}^{2} + x_{1} + \sin{(x_{1} )} + \sin{(x_{2}^{2} )}$ & $x_{1}^{3} + 1.923 x_{1} - 0.024 x_{2}^{2} (x_{1} - 37.848) + 0.998 |{x_{1}}|^{1.923}$ & 0.9996 & 0 & [-1, 1] \\ \hline
		Livermore-15 & $\sqrt[5]{x_{1}}$ & ${x_{1}}^{\frac{1}{5}}$ & 1.0000 & 1 & [-3, 3] \\ \hline
		Livermore-16 & $x_{1}^{\frac{2}{3}}$ & $x_{1}^{\frac{2}{3}}$ & 1.0000 & 1 & [-3, 3] \\ \hline
		Livermore-17 & $4 \sin{(x_{1} )} \cos{(x_{2} )}$ & $4 \sin{(x_{1} )} \cos{(|{x_{2}}| )}$ & 1.0000 & 0 & [-3, 3] \\ \hline
		Livermore-18 & $\sin{(x_{1}^{2} )} \cos{(x_{1} )} - 5$ & $\sin{(x_{1}^{2} )} \cos{(x_{1} )} - 5$ & 1.0000 & 1 & [-3, 3] \\ \hline
		Livermore-19 & $x_{1}^{5} + x_{1}^{4} + x_{1}^{2} + x_{1}$ & $x_{1}^{3} - 55.263 x_{1} + 22.809 e^{1.0 x_{1}} - 24.473 e^{- 0.813 x_{1}}$ & 0.9973 & 0 & [-3, 3] \\ \hline
		Livermore-20 & $e^{- x_{1}^{2}}$ & $e^{-x_{1}^{2}}$ & 1.0000 &1 & [-3, 3] \\ \hline
		Livermore-21 & $x_{1}^{8} + x_{1}^{7} + x_{1}^{6} + x_{1}^{5} + x_{1}^{4} + x_{1}^{3} + x_{1}^{2} + x_{1}$ & $0.87 x_{1} + 0.028 (0.144 x_{1} + 1)^{40.202} + 0.756 |{x_{1}}|^{3.273}$ & 0.9999 & 0 & [-1, 1] \\ \hline
		Livermore-22 & $e^{- 0.5 x_{1}^{2}}$ & $e^{- \frac{x_{1}^{2}}{2}}$ & 1.0000 & 1 & [-3, 3] \\ \hline
		Nguyen-2' & $4 x_{1}^{4} + 3 x_{1}^{3} + 2 x_{1}^{2} + x_{1}$ & $319.795 x_{1}^{2} (\sin{(\pi (0.05 x_{1} + 11.52) )} + 1.004) + \sin{(x_{1} )}$ & 0.9999 & 0 & [-1, 1] \\ \hline
		Nguyen-5' & $\sin{(x_{1}^{2} )} \cos{(x_{1} )} - 2$ & $- 0.283 \sin{\left(\left|{x_{1}}\right|^{0.03} \right)} \cos{\left(3.112 x_{1} \right)} - 1.776$ & 0.9999 & 0 & [-1, 1] \\ \hline
		Nguyen-8' & $\sqrt[3]{x_{1}}$ & $\sqrt[3]{x_{1}}$ & 1.0000 & 1 & [0, 4] \\ \hline
		Nguyena-8'' & $\sqrt[3]{x_{1}^{2}}$ & $x_{1}^{\frac{2}{3}}$ & 1.0000 & 1 & [0, 4] \\ \hline
		Nguyen-1c & $3.39 x_{1}^{3} + 2.12 x_{1}^{2} + 1.78 x_{1}$ 
		& $(155.094 x_{1} + 155.094 \sin{(0.004 x_{1}^{2} )}) (\cos{(0.209 x_{1} - 9.36 )} + 1)$ & 0.9999 & 0 & [-1, 1] \\ \hline
		Nguyen-5c & $\sin{(x_{1}^{2} )} \cos{(x_{1} )} - 0.75$ & $- 0.283 \sin{\left(\left|{x_{1}}\right|^{0.027} \right)} \cos{\left(3.1 x_{1} \right)} - 0.525$ & 0.9999 & 0 & [-1, 1] \\ \hline
		Nguyen-7c & $\log{(x_{1} + 1.4 )} + \log{(x_{1}^{2} + 1.3 )}$ & $\log{(x_{1} + 1.4 )} + \log{(x_{1}^{2} + 1.3 )}$ & 1.0000 & 1 & [0, 2] \\ \hline
		Nguyenc-8c & $1.109 x_{1}$ & $1.109 x_{1}$ & 1.0000 & 1 & [0, 4] \\ \hline
		Nguyen-10c & $\sin{(1.5 x_{1} )} \cos{(0.5 x_{1} )}$ & $\sin{(1.5 x_{1} )} \cos{(0.5 x_{1} + 6.282 )}$ & 1.0000 & 0 & [0, 1] \\ \hline
		Korns-1 & $24.3 x_{1}^{4} + 1.57$ 
		& $-\frac{11.559}{\left|{x_{1}}\right|^{4.287 \cdot 10^{-8}}} + 24.3 \left|{x_{1}}\right|^{4.0} + 13.129$ & 1.0000 & 0 & [-1, 1] \\ \hline
		Korns-4 & $0.13 \sin{(x_{1} )} - 2.3$ & $0.13 \sin{(x_{1} )} - 2.3$ & 1.0000 & 1 & [-1, 1] \\ \hline
		Korns-5 & $2.13 \log{(x_{1} )} + 3$ & $2.13 \log{(x_{1} )} + 3$ & 1.0000 & 1 & [-1, 1] \\ \hline
		Korns-6 & $0.13 \sqrt{x_{1}} + 1.3$ & $0.13 x_{1}^{0.5} + 1.3$ & 1.0000 & 1 & [-1, 1] \\ \hline
		Korns-7 & $2.1 - 2.1 e^{- 0.55 x_{1}}$ & $20.354 x_{1} (\sin{(\pi (0.047 x_{1} - 12.535) )} + 1.002) + \sin{(x_{1} )}$ & 0.9999 & 0 & [-1, 1] \\ \hline
		Korns-11 & $11 \cos{(7.23 x_{1}^{3} )} + 6.87$ & $11 \cos{(7.23 x_{1}^{3} )} + 6.87$ & 1.0000 & 1 & [-1, 1] \\ \hline
		Korns-12 & $- 2.1 \sin{(1.3 x_{2} )} \cos{(9.8 x_{1}^{3} )} + 2$ & $- 3 \cos{(29.4 x_{1} )} \sin{(1.3 x_{2} )} + 2$ & 1.0000 & 0 & [-1, 1] \\ \hline
		Neat-1 & $x_{1}^{4} + x_{1}^{3} + x_{1}^{2} + x_{1}$ & $x_{1}^{3} + x_{1} + x_{1}^{4} + x_{1}^{2}$ & 1.0000 & 1 & [-1, 1] \\ \hline
		Neat-2 & $x_{1}^{5} + x_{1}^{4} + x_{1}^{3} + x_{1}^{2} + x_{1}$ & $- 0.186 x_{1} + 0.409 e^{2.734 x_{1}} - 0.422 e^{x_{1}^{2}}$ & 0.9999 & 0 & [-1, 1] \\ \hline
		Neat-3 & $\sin{(x_{1}^{2} )} \cos{(x_{1} )} - 1$ & $\sin{(x_{1}^{2} )} \cos{(x_{1} )} - 1$ & 1.0000 & 1 & [-1, 1] \\ \hline
		Neat-4 & $\log{(x_{1} + 1 )} + \log{(x_{1}^{2} + 1 )}$ & $\frac{21.597 x_{1} + 21.597 e^{- \frac{1.188}{x_{1}^{1.037}}}}{\pi^{2.629}}$ & 0.9999 & 0 & [0, 2] \\ \hline
		Neat-5 & $2 \sin{(x_{1} )} \cos{(x_{2} )}$ & $2 \sin{(x_{1} )} \cos{(x_{2} )}$ & 1.0000 & 1 & [-1, 1] \\ \hline
		Neat-7 & $- 2.1 \cos{(9.8 x_{1} )} \sin{(1.3 x_{2} )} + 2$ & $2.1 \sin{(1.3 x_{2} )} \cos{(9.8 x_{1} + 3.141 )} + 2$ & 1.0000 & 0 & [-1, 1] \\ \hline
		Neat-8 & $\frac{e^{- x_{1}^{2}}}{6.25 (0.4 x_{2} - 1)^{2} + 1.2}$ & $\frac{e^{-x_1^2}}{(6.25 (0.4 x_2 - 1)^2 + 1.2)}$ & 0.9591 & 0 & [0.3, 4] \\ \hline
		Neat-9 & $\frac{1}{1 + \frac{1}{x_{2}^{4}}} + \frac{1}{1 + \frac{1}{x_{1}^{4}}}$ & $|{x_{1}}|^{3.786} - 0.498 |{x_{1}}|^{5.782} - 0.083 |{x_{2}}| + 0.607 |{x_{2}}|^{2.541}$ & 0.9990 & 0 & [-1, 1] \\ \hline
		Jin-1 & $2.5 x_{1}^{4} - 1.3 x_{1}^{3} + 0.5 x_{2}^{2} - 1.7 x_{2}$ & $2.5 x_{1}^{4} - \frac{0.016 x_{1}^{3} x_{2}}{(1 - 0.001 x_{1})^{916.176}} - 1.31 x_{1}^{3} + |{x_{2}}|$ & 0.9974 & 0 & [-3, 3] \\ \hline
		Jin-2 & $8 x_{1}^{2} + 8 x_{2}^{3} - 15$ & $8 x_{1}^{2} + 8 x_{2}^{3} - 15$ & 1.0000 & 1 & [-3, 3] \\ \hline
		Jin-3 & $0.2 x_{1}^{3} - 0.5 x_{1} + 0.5 x_{2}^{3} - 1.2 x_{2}$ & $x_{2}^{3} - 260.53 x_{2} - 1.963 \sin{(x_{1} )} + 2403.031 \sin{(0.001 x_{1} + 0.108 x_{2} )}$ & 0.9986 & 0 & [-3, 3] \\ \hline
		Jin-4 & $1.5 e^{x_{1}} + 5 \cos{(x_{2} )}$ & $1.5 e^{x_{1}} + 5 \cos{(x_{2} )}$ & 1.0000 & 1 & [-3, 3] \\ \hline
		Jin-5 & $6 \sin{(x_{1} )} \cos{(x_{2} )}$ & ${6\sin(x_1)\cos(x_2)}$ & 1.0000 & 1 & [-3, 3] \\ \hline
		Jin-6 & $1.35 x_{1} x_{2} + 5.5 \sin{((x_{1} - 1) (x_{2} - 1) )}$ & $1.35 x_{1} x_{2} + 5.5 \sin{((x_{1} - 1) (x_{2} - 1) )}$ & 1.0000 & 1 & [-3, 3] \\ \hline
		Nguyen-1 & $x_{1}^{3} + x_{1}^{2} + x_{1}$ & $x_{1}^{3} + x_{1}^{2} + x_{1}$ & 1.0000 & 1 & [-1, 1] \\ \hline
		Nguyen-2 & $x_{1}^{4} + x_{1}^{3} + x_{1}^{2} + x_{1}$ & $x_{1}^{3} +  x_{1} + x_{1}^{2} + x_{1}^{4}$ & 1.0000 & 1 & [-1, 1] \\ \hline
		Nguyen-3 & $x_{1}^{5} + x_{1}^{4} + x_{1}^{3} + x_{1}^{2} + x_{1}$ & $- 0.203 x_{1} + 0.417 e^{2.723 x_{1}} - 0.43 e^{x_{1}^{2}}$ & 0.9999 & 0 & [-1, 1] \\ \hline
		Nguyen-4 & $x_{1}^{6} + x_{1}^{5} + x_{1}^{4} + x_{1}^{3} + x_{1}^{2} + x_{1}$ & $0.885 x_{1} + 0.032 (0.313 x_{1} + 1)^{18.072} + 0.812 |{x_{1}}|^{3.067}$ & 0.9998 & 0 & [-1, 1] \\ \hline
		Nguyen-5 & $\sin{(x_{1}^{2} )} \cos{(x_{1} )} - 1$ & $\sin{(x_{1}^{2} )} \cos{(x_{1} )} - 1$ & 1.0000 & 1 & [-1, 1] \\ \hline
		Nguyen-6 & $\sin{(x_{1} )} + \sin{(x_{1}^{2} + x_{1} )}$ & $\sin{(x_{1} )} + \sin{(x_{1}^{2} + x_{1} )}$ & 1.0000 & 1 & [-1, 1] \\ \hline
		Nguyen-7 & $\log{(x_{1} + 1 )} + \log{(x_{1}^{2} + 1 )}$ & $1.364 x_{1} - 0.007 \sin{(5.999 x_{1} )} + 0.056 \sin{(x_{1} - 4.378 (x_{1} + 0.035)^{0.964} )}$ & 0.9999 & 0 & [0, 2] \\ \hline
		Nguyen-8 & $\sqrt{x_{1}}$ & $\sqrt{x_{1}}$ & 1.0000 & 1 & [0, 4] \\ \hline
		Nguyen-9 & $\sin{(x_{1} )} + \sin{(x_{2}^{2} )}$ & $\sin{(x_{1} )} + \sin{(x_{2}^{2} )}$ & 1.0000 & 1 & [0, 1] \\ \hline
		Nguyen-10 & $2 \sin{(x_{1} )} \cos{(x_{2} )}$ & $2 \sin{(x_{1} )} \cos{(x_{2} )}$ & 1.0000 & 1 & [0, 1] \\ \hline
		Nguyen-11 & $x_{1}^{x_{2}}$ & $x_{1}^{x_{2}}$ & 1.0000 & 1 & [0, 1] \\ \hline
		Nguyen-12 & $x_{1}^{4} - x_{1}^{3} + \frac{x_{2}^{2}}{2} - x_{2}$ & $- 0.998 x_{1}^{3} + 0.998 x_{1}^{4.002} - 1.0 x_{2} + 0.5 x_{2}^{2.0}$ & 1.0000 & 0 & [0, 1] \\ \hline
	\end{longtable}
}

\section{Supplementary results on Phenomenological \& First-principles benchmark}
\label{Supplementary_Phenomenological_First-principles benchmark}
\renewcommand{\arraystretch}{2.0}
\begin{longtable}{| >{\centering\arraybackslash}m{2.5cm} | >{\centering\arraybackslash}m{10.6cm} | >{\centering\arraybackslash}m{4cm} |}
	\caption{The details of the equations generated by ViSymRe for Phenomenological \& First-principles problems.}
	\label{tb:Supplementary_Phenomenological_First-principles benchmark} \\
	\hline
	\textbf{Dataset} & \textbf{equation} & \textbf{$R^2$} \\
	\hline
	\endfirsthead

	\multicolumn{3}{c}%
	{{\bfseries Table \thetable\ continued from previous page}} \\
	\hline
	\textbf{Dataset} & \textbf{equation} & \textbf{$R^2$} \\
	\hline
	\endhead
	\hline
	\multicolumn{3}{r}{{Continued on next page}} \\
	\endfoot

	\hline
	\endlastfoot

	bode & $0.15 e^{0.68 x_{1}} + 0.39$ & 0.997 \\
	hubble & $480 x_{1} + 169 \sin{\left(10 x_{1} + 9.11 \right)} + 25$ & 0.925 \\
	kepler & $362 x_{1}^{1.5}$ & 0.999 \\
	tully\_fisher & $\begin{aligned} & \left(\sin{\left(6.23 \cdot 10^{-8} x_{1}^{3.6} \right)} + 3.57\right) \\ & \cdot \cos{\left(0.1 \sqrt{x_{1}} \right)} - 18.4 \end{aligned}$ & 0.986 \\
	planck & $\frac{- 1.06 \cdot 10^{-12} x_{1} - 47.9}{\left(0.011 x_{2} + 1\right)^{1.3}} - 23$ & 0.996 \\
	ideal\_gas & $\log{\left(\frac{10.5 x_{1} x_{2}}{x_{3}} - 6.93 \right)}$ & 0.998 \\
	leavitt & $- 0.43 x_{1} + 11.8 + 3.14 e^{- 0.32 x_{1}^{3}}$ & 0.980 \\
	newton & $\frac{0.89 x_{3}^{0.016} \log{\left(x_{2} \right)}}{x_{1}^{0.031}} - 17$ & 0.990 \\
	rydberg & $\frac{3.55 x_{1} - 3.73}{x_{2}^{0.69}} - 15.9$ & 0.993 \\
	schechter & $\frac{831}{x_{1}^{0.0015}} - 947 + 114 e^{- 3.75 \cdot 10^{-11} x_{1}}$ & 0.996 \\
	absorption & $- 0.0045 x_{1} + 3.57 - 3.43 e^{- 0.032 x_{1}^{1.41}}$ & 0.995 \\
	supernovae\_zr & $0.0064 \pi^{4.44} e^{- 0.084 \left|{x_{1}}\right|^{0.82}}$ & 0.967 \\
	supernovae\_zg & $0.94 - 0.93 e^{- \frac{89.4}{x_{1}^{2}}}$ & 0.981 \\
\end{longtable}

\section{Supplementary details on ablation analysis}
\label{Supplementary_scale_ablation}
\renewcommand{\arraystretch}{2}

\begin{longtable}{| >{\centering\arraybackslash}m{2.5cm} | >{\centering\arraybackslash}m{10.6cm} | >{\centering\arraybackslash}m{4cm} |}
	
	\caption{True equations used for scale ablation analysis and their sampling Ranges.}
	\label{tab:Supplementary_scale_ablation} \\
	\hline
	\textbf{Name} & \textbf{True equation} & \textbf{Range} \\
	\hline
	\endfirsthead
	\hline
	\textbf{Name} & \textbf{True equation} & \textbf{Range} \\
	\hline
	\endhead
	\hline
	\endfoot
	\endlastfoot
	Random-1 & {\small $1 \times 10^{-5} \log{(x_{1} )}$} & {\small $[1e^{12},$ $1e^{15}]$} \\
	\hline
	Random-2 & {\small $1.000 \times 10^{6} \sqrt{x_{1}}$} & {\small $[1e^{16},$ $1e^{18}]$} \\
	\hline
	Random-3 & {\small $\begin{gathered} 1.000 \times 10^{5} \log{\begin{gathered}(x_{1} + \sqrt{x_{1}^{2} + 1} )\end{gathered}} \end{gathered}$} & {\small $[1e^{14},$  $1e^{16}]$} \\
	\hline
	Random-4 & {\small $\begin{gathered} x_{1}^{4} - x_{1}^{3} + \frac{x_{2}^{2}}{2} - x_{2} \end{gathered}$} & {\small $[1e^{8},$  $1e^{10}]$} \\
	\hline
	Random-5 & {\small $\begin{gathered} x_{2}^{2} + \frac{8}{x_{1}^{2} + 2} \end{gathered}$} & {\small $[-1e^{5},$ $-1e^{3}]$} \\
	\hline
	Random-6 & {\small $1.000 \times 10^{-4} \sin{(10 x_{1}^{2} )}$} & {\small $[-1,$ $1]$} \\
	\hline
	Random-7 & {\small $\begin{gathered} 40000 x_{1}^{4} + 3000000 x_{1}^{3} + 200000 x_{1}^{2} + x_{1} \end{gathered}$} & {\small $[1e^{-4},$ $1e^{-2}]$} \\
	\hline
	Random-8 & {\small $\begin{gathered} x_{1}^{9} + x_{1}^{8} + x_{1}^{7} + x_{1}^{6} + x_{1}^{5} + x_{1}^{4} + x_{1}^{3} + x_{1}^{2} + x_{1} \end{gathered}$} & {\small $[0,$ $0.01]$} \\
	\hline
	Random-9 & {\small $\frac{x_{1}^{2} x_{2}^{2}}{x_{1} + x_{2}}$} & {\small $[1e^{12},$  $1e^{14}]$} \\
	\hline
	Random-10 & {\small $\frac{10000 x_{1}^{5}}{x_{2}^{3}}$} & {\small $[1e^{13},$ $1e^{15}]$} \\
	\hline
	Random-11 & {\small $2 \times 10^{-6} \sqrt[3]{x_{1}}$} & {\small $[-3,$ $3]$} \\
	\hline
\end{longtable}

\end{document}